\pdfoutput=1
\documentclass{article}

\PassOptionsToPackage{numbers, compress}{natbib}
\PassOptionsToPackage{hyphens}{url}

\usepackage[eandd, final]{neurips_2026}

\usepackage[utf8]{inputenc} 
\usepackage[T1]{fontenc}    
\usepackage{hyperref}  
  
\usepackage{url}            
\hypersetup{
  colorlinks=true,
  linkcolor={blue!50!black},
  citecolor={blue!70!black},
  urlcolor={blue!80!black},
  filecolor=blue,
}
\usepackage{booktabs}       
\usepackage{amsfonts}       
\usepackage{amsmath}
\usepackage{cleveref} 
\usepackage{nicefrac}       
\usepackage{microtype}      
\usepackage{xcolor}         
\usepackage[table]{xcolor}

\usepackage{booktabs,multirow,colortbl,xcolor,graphicx}
\usepackage[most]{tcolorbox}
\usepackage{caption}
\usepackage{enumitem}
\usepackage{xspace}
\usepackage{pifont}
\usepackage{algorithm}
\usepackage{algpseudocode}
\usepackage{array}
\usepackage{tabularx}
\usepackage{subcaption}
\usepackage{wrapfig}
\usepackage{placeins}

\definecolor{anchorcolor}{HTML}{1F77B4}
\definecolor{typcolor}{HTML}{2CA02C}
\definecolor{atypcolor}{HTML}{D62728}

\definecolor{oursrow}{HTML}{DCEEFA}

\definecolor{findingbg}{HTML}{EAF5EA}
\definecolor{findingborder}{HTML}{4CAF50}
\newtcolorbox{finding}{
  colback=findingbg, colframe=findingborder,
  boxrule=0.5pt, left=4pt, right=4pt, top=3pt, bottom=3pt,
  arc=2pt, fontupper=\small,
  breakable,
  before skip=6pt, after skip=6pt
}

\providecommand{\eg}{\textit{e.g.}\@\xspace}
\providecommand{\ie}{\textit{i.e.}\@\xspace}

\newcommand{\cmark}{\textcolor{green!60!black}{\ding{51}}}
\newcommand{\xmark}{\textcolor{red!70!black}{\ding{55}}}

\newcommand{\codino}{Co-DINO\xspace}
\newcommand{\gdino}{Grounding~DINO\xspace}
\newcommand{\dinov}{DINOv3\xspace}
\newcommand{\openclip}{OpenCLIP\xspace}
\newcommand{\llava}{LLaVA-OneVision\xspace}

\title{Stranger Things: When Objects Appear Without Their Typical Neighbors}

\author{%
  Siddhartha Gairola$^{1}$\thanks{Corresponding author: \texttt{sgairola@mpi-inf.mpg.de}}\hspace{0.5em} Jiahao Xie$^{1}$\hspace{0.5em} Anna Kukleva$^{1}$\hspace{0.5em} Francesco Locatello$^{2}$\hspace{0.5em} Bernt Schiele$^{1}$ \\
  $^{1}$Max Planck Institute for Informatics, Saarland Informatics Campus, Germany \\
  $^{2}$Institute of Science and Technology Austria \\
}

\begin{document}

\maketitle
\vspace{-11pt}
\begin{abstract}
Visual context is a powerful cue for object recognition: scenes inform what, where, and at what scale objects are likely to appear. Yet over-reliance on context becomes a failure mode when models predict objects based on their typical neighbors rather than the objects' own visual features, a phenomenon we refer to as \textit{\textbf{typicality bias}}. While this issue has been extensively studied in the pre-Transformer era, it has often been assumed that architectural and scale improvements would naturally solve it. Despite recent attempts in isolated settings, a systematic diagnostic across the modern vision stack remains absent.
In this work, we provide such a diagnostic framework. Specifically, we measure typicality bias via normalized pointwise mutual information (NPMI) over object pairs, and introduce \textbf{STRANGE-Bench} (\textbf{S}ynthetic \textbf{T}ypicality-\textbf{RAN}ked \textbf{G}rounded \textbf{E}valuation \textbf{Bench}mark): a photorealistic, layout-controlled benchmark spanning 17 sub-tiers from typical to never-co-occurring object pairs. Across \textit{five model families} (specialist detectors, open-vocabulary detectors, self-supervised, vision-language, and multimodal large language models) and \textit{three recognition tasks} (supervised detection, open-vocabulary detection, and classification), we find the bias to be highly prevalent: gaps of 5--11~p.p.\@ persist when objects appear outside their typical company, and they do not consistently shrink with scale.
This persistence motivates a complementary question: how to mitigate it? Data augmentation is a natural lever, since it directly shapes the training distribution that produces the bias; yet existing augmentations are designed for visual diversity, not for breaking co-occurrence shortcuts: they sample from the very distribution that produces typicality bias, preserving rather than perturbing it. We therefore propose Anchor-Atypicality Sampling (AAS), a simple sampling rule for paste-based augmentations that uses the NPMI signal to deliberately tilt the training-time co-occurrence distribution toward atypical partners. AAS reduces the typicality gap by up to 1.6 p.p.\@ on natural-image COCO splits and 23.0 p.p.\@ on synthetic STRANGE-Bench across three representative model families (specialist detection, open-vocabulary detection, self-supervised classification), without sacrificing overall performance. Dataset: \href{https://huggingface.co/datasets/sidgairo18/strange-bench}{\nolinkurl{huggingface.co/datasets/sidgairo18/strange-bench}}

\end{abstract}
\vspace{-10pt}

\section{Introduction}

Modern recognition models are trained on large-scale datasets of natural images. Such images capture objects in their typical surroundings, \eg, a cow on grass, a boat in water, or a stove in a kitchen. These models can exploit co-occurrence cues that go unnoticed on standard benchmarks, since training and test sets draw from the same patterns. Failures surface only when objects appear in atypical contexts, \eg, a cow on a beach~\cite{beery2018recognition}. While context is itself informative for recognition~\cite{torralba2003contextual,oliva2007role}, models can come to use it as a shortcut around the object's visual features. This \emph{over-reliance} on co-occurring context has been studied through context-based interventions~\cite{shetty2019not,singh2020don,chakraborty2024visual,sarridis2025mavias}, controlled benchmarks~\cite{liu2015deep,sagawa2020distributionally,li2023whac}, and dedicated out-of-context detection~\cite{acharya2022detecting,bomatter2021pigs}. All these studies converge on the same conclusion: classifiers exploit co-occurrence shortcuts and fail when they are absent.

\looseness=-1It is unclear whether these findings still hold across the modern vision stack. Recent architectures and training paradigms such as end-to-end detection transformers~\cite{carion2020end,zong2023detrs,zhang2022dino}, open-vocabulary detectors that ground predictions in language~\cite{liu2024grounding,minderer2022simple,minderer2023scaling}, self-supervised vision foundation models~\cite{caron2021emerging,oquab2024dinov,simeoni2025dinov3}, vision-language models pretrained on billions of image-text pairs~\cite{radford2021learning,cherti2023reproducible,zhai2023sigmoid,tschannen2025siglip}, and large multimodal models that fuse vision and language at scale~\cite{li2024llava} are expected to alleviate context reliance through enhancements such as deformable attention, language grounding, as well as web-scale data diversity~\cite{xia2022vision,liu2024grounding,radford2021learning}. Recent work has examined whether such shortcuts persist for specific subsets of the stack: segmentation and detection through explainable artificial intelligence (XAI) methods~\cite{dreyer2023revealing}, and nuisance-based robustness benchmarks~\cite{zhao2022ood,dunkel2025cns}. However, a systematic diagnostic across diverse model families that evaluates on object pairs never co-occurring in natural data remains underexplored.

\begin{figure}[t]
\centering
\includegraphics[width=\linewidth]{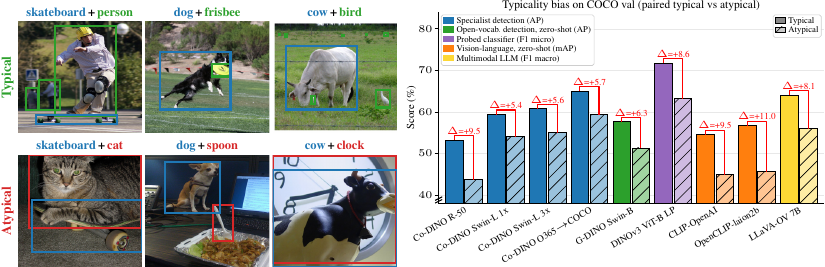}

\vspace{-3pt}
\caption{\textbf{Typicality bias across the modern vision stack.}
\emph{Left:} Three \emph{anchor classes} (target object categories we examine): \textit{\textcolor{anchorcolor}{skateboard}}, \textit{\textcolor{anchorcolor}{dog}}, and \textit{\textcolor{anchorcolor}{cow}}, shown in their {\color{typcolor}\textbf{typical}} company (top row: with a \textit{\textcolor{typcolor}{person}}, with a \textit{\textcolor{typcolor}{frisbee}}, with a \textit{\textcolor{typcolor}{bird}}) and in {\color{atypcolor}\textbf{atypical}} surroundings (bottom row: with a \textit{\textcolor{atypcolor}{cat}}, with a \textit{\textcolor{atypcolor}{spoon}}, with a \textit{\textcolor{atypcolor}{clock}}). \emph{Right:} Each model's score on the anchor classes in COCO \texttt{val} images where they appear with typical vs.\ atypical partners (NPMI quintile splits, \S\ref{sec:diagnostic}). The performance gap $\textcolor{atypcolor}{\Delta} =$ (\textcolor{black}{Typical} -- \textcolor{black}{Atypical}) persists across all five model families and three tasks we examine, ranging from $\textcolor{atypcolor}{\Delta = 5.4}$~p.p.\@ (Co-DINO Swin-L) to $\textcolor{atypcolor}{\Delta = 11.0}$~p.p.\@ (OpenCLIP).}
\label{fig:teaser}
\vspace{-12pt}
\end{figure}

In this work, we revisit this phenomenon systematically and at scale across modern vision models. We refer to it as \textit{\textbf{typicality bias}}: the tendency of recognition models to base predictions on familiar co-occurring context rather than the target object's own visual evidence. As shown in Fig.~\ref{fig:teaser}, a model exhibiting typicality bias performs well when objects appear in typical contexts and degrades when they appear alone or in atypical contexts.

To diagnose typicality bias systematically, we introduce a framework based on normalized pointwise mutual information (NPMI) over object pairs. For each anchor class, we use NPMI to construct paired evaluation splits in which the anchor (object category under examination) appears alongside either typical or atypical partners. To extend this natural-data evaluation with co-occurrence regimes that natural data do not cover (see Fig.~\ref{fig:npmi_matrix}), we further release \textbf{STRANGE-Bench} (\textbf{S}ynthetic \textbf{T}ypicality-\textbf{RAN}ked \textbf{G}rounded \textbf{E}valuation \textbf{Bench}mark), a photorealistic, layout-controlled synthetic benchmark spanning the full spectrum from typical to never-co-occurring object pairs.

~\looseness=-1A complementary question is how to mitigate typicality bias. Data augmentation is a natural lever for this, since it directly shapes the training distribution that produces the bias. In practice, existing augmentations are designed to add visual diversity, not to break co-occurrence shortcuts: both non-generative~\cite{bochkovskiy2020yolov4,zhang2017mixup,ghiasi2021simple} and generative~\cite{zhao2023x,kupyn2024dataset,xie2025mosaicfusion} methods sample from the natural training distribution, preserving rather than perturbing it. We present the first systematic study of how popular data augmentations interact with typicality bias across representative model families, and propose a simple \emph{context-aware} sampling rule that reduces the bias while preserving overall accuracy.

Our findings reveal three patterns. \textit{First,} typicality bias persists across diverse model families and tasks, with gaps of $\mathbf{5}$--$\mathbf{11}$~p.p.\@ between typical and atypical splits. \textit{Second}, the bias does not consistently shrink with scale: more pretraining data does not narrow the gap, and larger models narrow it without closing it, indicating that scale alone is not a solution. \textit{Third}, most non-paste augmentations leave the gap largely unchanged or narrow it by lowering both sides, while paste-based ones narrow it on STRANGE-Bench. AAS lifts atypical scenes further without hurting typical ones, and reduces the gap of standard training by up to $\mathbf{1.6}$~p.p.\@ on natural-image COCO~\cite{mscoco14} splits and $\mathbf{23.0}$~p.p.\@ on synthetic STRANGE-Bench, without sacrificing overall performance.

Our main \textbf{contributions} are summarized as follows:

\begin{enumerate}[leftmargin=*,nosep]
\item \textbf{Typicality bias persists across the modern vision stack.} We provide a unified diagnostic of typicality bias spanning five model families (specialist detection, open-vocabulary detection, self-supervised representations, vision-language models, and multimodal LLMs) and three recognition tasks. Our key finding is that typicality bias persists across diverse tasks and scales.

\item \textbf{A photorealistic synthetic benchmark for tiered typicality.} We present \textbf{STRANGE-Bench}, a layout-controlled synthetic benchmark spanning the full spectrum from typical to never-co-occurring object pairs, enabling controlled evaluation in low-NPMI regimes that natural data cannot cover; blinded user studies confirm that its gaps are not rendering artifacts.

\item \textbf{Anchor-Atypicality Sampling for mitigation.} We present the first systematic study of how popular data augmentations interact with typicality bias across representative model families, and propose Anchor-Atypicality Sampling (AAS), a simple sampling rule for paste-based augmentation that reduces typicality bias while preserving overall performance.
\end{enumerate}

\section{Related Work}

\noindent\textbf{Contextual and co-occurrence bias in recognition models.} Reliance on co-occurring context as a predictive shortcut is well-documented for older architectures, but less so for the modern stack. \emph{Causal interventions} demonstrate the dependence by manipulating features or inputs: via causal inpainting on convolutional neural network (CNN) features~\cite{shetty2019not}, and via feature-space splitting on COCO-Stuff~\cite{caesar2018coco,singh2020don}. 
\emph{Out-of-context benchmarks} measure model behavior when objects appear in atypical scenes~\cite{acharya2022detecting,bomatter2021pigs}. \emph{Concept-graph approaches} discover and mitigate co-occurrence biases via concept-graph clique sampling~\cite{chakraborty2024visual} or open-set foundation-model tagging~\cite{sarridis2025mavias}.
More recent work has examined context bias in modern models, but only in narrower settings: XAI for segmentation and detection~\cite{dreyer2023revealing}, nuisance-based robustness benchmarks~\cite{zhao2022ood}, referring expression generation~\cite{merlo2026cooco}, and compositional action recognition~\cite{ahn2026can}.
In contrast to these narrower studies, we apply one diagnostic across five model families, complemented by \textbf{STRANGE-Bench}, which covers co-occurrence regimes absent from natural data, and we also test mitigations.

\begin{table}[t!]
\centering
\caption{\textbf{Taxonomy of popular data augmentations for visual recognition.} None are designed to break co-occurrence shortcuts. Columns: Std (standard: resize, crop, flip), MixUp~\cite{zhang2017mixup}, Mosaic~\cite{bochkovskiy2020yolov4}, YOLOX~\cite{ge2021yolox} (MixUp + Mosaic), SCP (Simple Copy-Paste~\cite{ghiasi2021simple}), XPaste~\cite{zhao2023x}, InstAug (InstanceAug~\cite{kupyn2024dataset}), MFusion (MosaicFusion~\cite{xie2025mosaicfusion}). Static augmentations operate within a single image; dynamic ones introduce new instances at training time.}
\label{tab:augmentations}
\small
\begin{tabular}{@{}lcccccccc@{}}
\toprule
\textbf{Property} & \textbf{Std} & \textbf{MixUp} & \textbf{Mosaic} & \textbf{YOLOX} & \textbf{SCP} & \textbf{XPaste} & \textbf{InstAug} & \textbf{MFusion} \\
\midrule
Non-generative         & \cmark       & \cmark         & \cmark         & \cmark          & \cmark       & \xmark          & \xmark           & \xmark           \\
Generative             & \xmark       & \xmark         & \xmark         & \xmark          & \xmark       & \cmark          & \cmark           & \cmark           \\
\midrule
Static                 & \cmark       & \xmark         & \xmark         & \xmark          & \xmark       & \xmark          & \cmark           & \xmark           \\
Dynamic                & \xmark       & \cmark         & \cmark         & \cmark          & \cmark       & \cmark          & \xmark           & \cmark           \\
\midrule
\textbf{Context-aware} & \xmark       & \xmark         & \xmark         & \xmark          & \xmark       & \xmark          & \xmark           & \xmark           \\
\bottomrule
\end{tabular}
\vspace{-10pt}
\end{table}

\noindent\textbf{Data augmentation for visual recognition.} Data augmentation shapes the training distribution that models see, but existing methods were not designed with co-occurrence shortcuts in mind. \emph{Non-generative methods} perturb individual images (flip, crop, color jitter) or stitch existing instances together (MixUp~\cite{zhang2017mixup}, Mosaic~\cite{bochkovskiy2020yolov4}, Simple Copy-Paste~\cite{ghiasi2021simple}). \emph{Generative methods} synthesize new instances using text-to-image diffusion (XPaste~\cite{zhao2023x}, InstanceAug~\cite{kupyn2024dataset}, MosaicFusion~\cite{xie2025mosaicfusion}). Across both families (see Table~\ref{tab:augmentations}), pasted or generated instances are sampled from the natural co-occurrence distribution. Whether this preserves typicality bias, and whether targeted modifications can reduce it, have not been systematically studied. We present such a study on representative model families, and
propose \textbf{Anchor-Atypicality Sampling (AAS)}, a sampling rule that plugs into any paste-based augmentation and reduces typicality bias while preserving overall performance.

\noindent\textbf{Modern visual recognition models.} The vision recognition stack has evolved rapidly along architectural and supervision axes. \emph{Specialist detection transformers} such as DETR~\cite{carion2020end}, Deformable DETR~\cite{zhu2021deformable}, and Co-DETR/Co-DINO~\cite{zong2023detrs} formulate detection end-to-end via (deformable) attention. \emph{Open-vocabulary detectors} such as GLIP~\cite{li2022grounded} and Grounding DINO~\cite{liu2024grounding} extend detection to text-conditioned localization. \emph{Self-supervised foundation models} such as DINOv2~\cite{oquab2024dinov} and DINOv3~\cite{simeoni2025dinov3} learn general-purpose visual representations from hundreds of millions of unlabeled images. \emph{Vision--language models} (CLIP~\cite{radford2021learning}, OpenCLIP~\cite{cherti2023reproducible}) and \emph{large multimodal models} (LLaVA-OneVision~\cite{li2024llava}) use web-scale image--text pretraining for zero-shot recognition and image understanding. Each of these design choices, from deformable attention focusing on object regions to language grounding and web-scale data diversity, is expected to be resistant to context shortcuts. Yet none is trained or evaluated with such shortcuts in mind, leaving their susceptibility to typicality bias underexamined, a gap our diagnostic, benchmark, and mitigation directly address.

\section{Diagnosing and Mitigating Typicality Bias}

\looseness=-1 In this section, we present a framework for diagnosing and mitigating typicality bias, built around normalized pointwise mutual information (NPMI). We first use NPMI to construct paired \emph{typical} and \emph{atypical} evaluation splits on real images (Sec.~\ref{sec:diagnostic}). To probe co-occurrence regimes that real data cannot reach, we then introduce \textbf{STRANGE-Bench}, a layout-grounded synthetic benchmark spanning 17 sub-tiers from natural co-occurrence to never-co-occurring object pairs (Sec.~\ref{sec:strange}). Finally, we present \textbf{Anchor-Atypicality Sampling (AAS)}, a sampling rule for paste-based augmentations that perturbs training-time co-occurrence using the same NPMI signal (Sec.~\ref{sec:aas}). STRANGE-Bench is an evaluation instrument and AAS a training-time intervention.

\subsection{NPMI Diagnostic on Natural Data}
\label{sec:diagnostic}

To quantify typicality bias on real data, we measure how a model's behavior changes between paired scenes where an \emph{anchor class} (the target object category we examine) appears alongside its typical co-occurrence partners versus its atypical ones. This requires a typicality measure that is comparable across object categories of widely different frequencies, since we aggregate gaps over all anchor classes (\eg, \emph{person} co-occurs with almost everything, while \emph{toaster} with almost nothing). We use normalized pointwise mutual information (NPMI), which is bounded in $[-1, 1]$ and corrects for marginal class frequency, making typicality directly comparable across rare and common categories.

\begin{figure}[t!]
\centering
\includegraphics[width=\linewidth]
{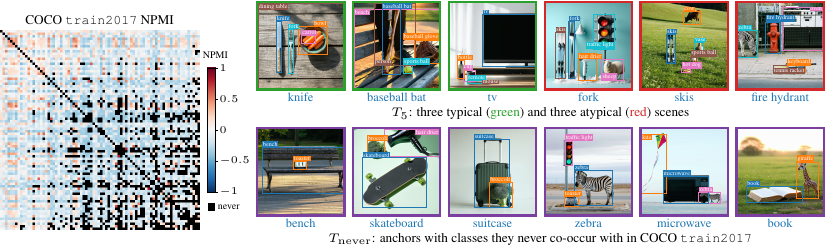}
\vspace{-15pt}
\caption{\textbf{NPMI signal on COCO \texttt{train2017} and example
STRANGE-Bench scenes.} \emph{Left:} The $80{\times}80$ NPMI matrix over object categories computed from COCO \texttt{train2017} (Sec.~\ref{sec:diagnostic}). Hot (red) cells correspond to high-NPMI (typical) pairs, cool (blue) cells to low-NPMI (atypical) pairs, and black cells to category pairs that never co-occur in
training. \emph{Right, top row ($T_5$):} three typical STRANGE-Bench (Sec.~\ref{sec:strange}) scenes
(\textcolor{typcolor}{green border}: anchors \emph{knife, baseball bat, tv}) and three atypical scenes (\textcolor{atypcolor}{red border}:
anchors \emph{fork, skis, fire hydrant}); the anchor class is shown in
\textcolor{anchorcolor}{blue} below each panel. Anchors differ across
columns and are matched only by NPMI tier, not by identity.
\emph{Right, bottom row ($T_{\text{never}}$):} STRANGE-Bench scenes for
anchor classes \emph{bench, skateboard, suitcase, zebra, microwave}, and \emph{book}, each paired with partners that never co-occur with
the anchor in COCO \texttt{train2017}.
}
\label{fig:npmi_matrix}
\vspace{-5pt}
\end{figure}

\noindent\textbf{NPMI definition.}
For categories $c_i, c_j$, we compute pointwise mutual information from image-level co-occurrence (\ie, whether both classes appear in an image, ignoring count and location) and normalize it:
\begin{equation}
\text{NPMI}(c_i, c_j) \;=\; \frac{\log \big[ p(c_i, c_j) / p(c_i)\,p(c_j) \big]}{- \log p(c_i, c_j)} \;\in\; [-1, 1].
\end{equation}
The numerator is positive when $c_i$ and $c_j$ co-occur more often than chance, negative when they co-occur less, and zero when they are independent; normalizing by $-\log p(c_i, c_j)$ bounds the score to $[-1, 1]$, with the bound saturated when $c_i$ and $c_j$ always or never co-occur. Typical pairs (high NPMI) include \emph{toaster--microwave}; atypical pairs (low NPMI) include \emph{toaster--car}, and some pairs, such as \emph{toaster--giraffe}, never co-occur (black cells in Fig.~\ref{fig:npmi_matrix}, left). We compute the matrix on COCO \texttt{train2017} ($N = {}$$117{,}266$ images with at least one annotation, $80$ categories), which provides clean instance annotations across diverse everyday objects~\cite{mscoco14}. The same matrix drives split construction here and the AAS sampler in Sec.~\ref{sec:aas}.

\noindent\textbf{Paired typical/atypical splits.}
For each anchor class $c$, we build paired evaluation splits. Both splits contain images where $c$ is present alongside at least one partner, such that \textit{typical} partners appear only in one split and \textit{atypical} partners only in the other. Concretely, for anchor $c$ and a quantile parameter $\alpha \in (0, 0.5)$, we define typical and atypical partner sets via per-anchor NPMI quantiles:
\begin{equation}
\mathcal{N}^{\text{typ}}(c) = \{\, j : \text{NPMI}(c, j) \geq q_{1-\alpha}^{(c)} \,\}, \qquad
\mathcal{N}^{\text{aty}}(c) = \{\, j : \text{NPMI}(c, j) \leq q_{\alpha}^{(c)} \,\},
\end{equation}
where $q_{p}^{(c)}$ is the $p$-quantile of NPMI in row $c$, restricted to partners with sufficient train-set co-occurrence with $c$.\footnote{Specifically, typical candidates require $N_{ij} \geq 5$; atypical candidates require $N_{ij} \geq 1$ and at least one val image containing both $c$ and $j$, and the atypical quantile is taken over these candidates. The count floors keep wide quantile cuts on noisy long-tail rows from drifting into the neutral region.} Per-anchor quantiles, rather than a fixed top-$K$, calibrate each anchor's partner set to its own NPMI distribution: a fixed $K$ would over-include partners for common anchors (which have many high-NPMI pairs) and under-include for rare ones. The choice of $\alpha$ trades partner-set sharpness against split sample size. A val image $I$ joins the Typical split for $c$ iff $c \in I$, $I \cap \mathcal{N}^{\text{typ}}(c) \neq \emptyset$, and $I \cap \mathcal{N}^{\text{aty}}(c) = \emptyset$ (and symmetrically for the Atypical split). This scene-purity requirement prevents typical and atypical partners from leaking across splits when an image contains both. The choice of $\alpha$, the val-coverage floor for long-tail anchors, and additional implementation details are provided in Appendix~\ref{app:splits-impl}. Figure~\ref{fig:npmi_matrix} shows the NPMI matrix used here, alongside example scenes from STRANGE-Bench at representative NPMI tiers; we describe its construction next.

\subsection{STRANGE-Bench: Tiered Synthetic Evaluation}
\label{sec:strange}
\begin{figure*}[t!]
    \centering
    \includegraphics[width=\linewidth]{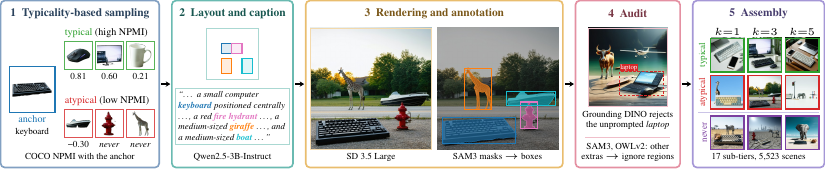}
    \caption{\textbf{Overview of the STRANGE-Bench generation pipeline}, illustrated with outputs of the actual build. (1)~\emph{Typicality-based sampling}: partners are drawn by their COCO NPMI with the anchor, high for the typical arm and low or never co-occurring for the atypical arm. (2)~\emph{Layout and caption}: a planner places the categories in a layout and writes a caption from it; each category keeps its color in panels 2 and 3. (3)~\emph{Rendering and annotation}: SD 3.5 Large renders the caption up to three times until SAM3 finds every category, and the boxes of the SAM3 masks are the annotations. (4)~\emph{Audit}: Grounding DINO rejects scenes with unintended objects (here, a laptop), and SAM3 and OWLv2 turn the remaining ones into ignore regions or drop the scene. (5)~\emph{Assembly}: typical and atypical sub-tiers at each contextual load $k$, plus $T_{\text{never}}$, form STRANGE-Bench.}
    \label{fig:pipeline_overview}
\end{figure*}

\vspace{-5.0pt}
The diagnostic in Sec.~\ref{sec:diagnostic} covers only the portion of the NPMI spectrum that exists in natural data, \ie, pairs from typical (high NPMI) to atypical (low NPMI). It cannot reach pairs that never co-occur in training, where NPMI is undefined and no natural images exist. Nor can it vary contextual load while controlling for typicality, since natural images do not allow for adding or removing partners. To examine these regimes, we build \textbf{STRANGE-Bench} (\textbf{S}ynthetic \textbf{T}ypicality-\textbf{RAN}ked \textbf{G}rounded \textbf{E}valuation \textbf{Bench}mark), a layout-grounded photorealistic benchmark of $5{,}523$ scenes with $17{,}332$ annotations across 17 sub-tiers (see Fig.~\ref{fig:pipeline_overview}). Each scene contains an \emph{anchor} category and $k \in \{0, \dots, 5\}$ partners, all uniformly typical or atypical with respect to the anchor. We call $k$ the \emph{contextual load} and a ($k$, typicality) cell a \emph{tier}: $T_1$ to $T_6$ cover $k = 0$ to $5$ ($T_1$ varies only the background), and App.~\ref{app:strange:tiers} lists every sub-tier. $k \le 4$ covers $91\%$ of COCO \texttt{train2017} images; $k$ is set by design rather than matched to natural statistics. A separate $T_{\text{never}}$ tier pairs the anchor with categories that never co-occur with it in COCO~\texttt{train2017}; atypical tiers also contain such pairs, but $T_{\text{never}}$ requires every pair in the scene, partners included, to have zero co-occurrences. This sweeps bias as a function of both typicality and contextual load.

\noindent\textbf{Generation pipeline.}
Each scene is generated by a five-stage pipeline (Fig.~\ref{fig:pipeline_overview}) that turns a set of categories into an annotated photorealistic image. (1)~\emph{Typicality-based sampling.} For a target sub-tier, we sample an anchor and $k$ partners from the COCO label set, preferring partners with high NPMI (typical arm) or low NPMI (atypical arm) with the anchor and with each other (App.~\ref{app:strange:tiers}). (2)~\emph{Layout and caption.} A \textit{Qwen2.5-3B-Instruct}~\cite{qwen2024qwen25} planner (or, for $21\%$ of the scenes, a rule-based solver) places the categories in a bounding-box layout and writes from it one wide-shot caption that names every object with its position in the image; the renderer sees the layout only through this caption. The caption asks for a minimalist scene with every object fully visible, which targets the two failure modes of our pilot renders: cropped objects and unprompted clutter. (3)~\emph{Rendering and annotation.} \textit{Stable Diffusion 3.5 Large}~\cite{stabilityai2024sd35,esser2024scaling} renders the caption, and \textit{SAM3}~\cite{carion2026sam3} segments each intended category from a text prompt (\eg, ``a giraffe''); the boxes of its instance masks become the annotations. We render up to three times and keep the first render in which SAM3 finds every intended category ($89.5\%$ of the categories are found on the first render, $95.9\%$ within three). (4)~\emph{Audit.} An object that was not asked for would turn its detections into false positives and could bring typical context into an atypical scene. A \textit{Grounding DINO}~\cite{liu2024grounding} auditor therefore rejects scenes with unintended or duplicate objects, and a second audit with SAM3 and OWLv2~\cite{minderer2023scaling} turns each remaining one into an ignore region, where predictions are not counted, or drops the scene if it is atypical and the object co-occurs with an intended category more often than chance (NPMI $> 0$). (5)~\emph{Assembly.} Scenes that miss an intended category are dropped, and the typical and atypical sub-tiers at each $k$ are cut to the same size; $44.6\%$ of the rendered scenes enter STRANGE-Bench. Prompt templates, audit thresholds, the per-tier composition, and the yield of each stage are provided in Appendix~\ref{app:strange-pipeline}.

\subsection{Anchor-Atypicality Sampling (AAS)}
\label{sec:aas}

Standard Copy-Paste~\cite{ghiasi2021simple} and X-Paste~\cite{zhao2023x} sample paste classes uniformly (or by inverse frequency), so the trained co-occurrence distribution mirrors the source distribution and reinforces typicality bias. AAS replaces uniform sampling with an anchor-conditioned, atypicality-tilted distribution that shifts the training-time joint toward low-NPMI pairs while preserving overall class coverage.

For each training image with present classes $\mathcal{P}$, we take as anchor $a$ the class of the largest ground-truth object. We then sample each paste class $j$ from an instance pool $\mathcal{C}_{\text{pool}}$ (pre-computed real and synthetic individual instances over all categories as in~\cite{zhao2023x}) with probability
\begin{equation}
P(j \mid a) \;\propto\; \pi(j) \, \exp\big(\beta \, A(a, j)\big),
\end{equation}
\looseness=-1 where $A(a, j) = (1 - \text{NPMI}(a, j))/2 \in [0, 1]$ is the atypicality of the pair, set to $1$ for pairs that never co-occur in COCO \texttt{train}, $\beta \geq 0$ controls the sharpness of the tilt ($\beta = 0$ recovers the baseline), and $\pi(j)$ is uniform. We use $\beta = 3.5$ throughout and ablate it in Appendix~\ref{app:beta_sweep}. Pasted instances are scaled per-class via $S_r \sim \mathcal{N}(\mu_c, \sigma_c^2)$ statistics estimated from the pool, randomly placed, and composited with occlusion-aware mask updates following X-Paste~\cite{zhao2023x}.
We apply AAS as a late stage: a paste-trained model (SCP or XPaste) resumes its last epochs on COCO plus a small number of atypical scenes generated by the pipeline of Sec.~\ref{sec:strange}, and the resulting weights are averaged 1:1 with those of the base model (App.~\ref{app:exp-recipes}). These scenes come from the generator of STRANGE-Bench but are separate renders, never benchmark images; App.~\ref{app:circularity} tests this link and the shared NPMI statistics.

\looseness=-1 AAS explicitly perturbs the training-time NPMI structure: the model sees anchors paired with their bottom-NPMI partners far more often than the natural distribution would allow. This directly attacks the shortcut behind typicality bias (``predict $j$ because $a$ is present'') by making that shortcut unreliable during training. Since the tilt is anchor-conditional rather than global, typical co-occurrences in the original data are largely preserved, so the intervention is asymmetric: gap-closing without overall-performance regression.

\section{Experiments}
\label{sec:experiments}
\vspace{-5.0pt}
In this section, we evaluate typicality bias across the modern vision stack and study its mitigation. We first provide our experimental setup in Sec.~\ref{sec:experimental_setup}, including datasets, evaluation metrics, models, and augmentation suites. We then survey five model families on the COCO NPMI splits and the synthetic STRANGE-Bench in Sec.~\ref{sec:results_diagnostic}, establishing that the bias persists across the stack and does not consistently shrink with scale. We finally run the systematic mitigation study on three representative families in Sec.~\ref{sec:results_mitigation}, showing that AAS reduces the typicality gap while preserving overall performance. Sec.~\ref{sec:results_human} checks with human raters that the gaps come from the models and not from the rendered scenes.

\subsection{Experimental Setup}
\label{sec:experimental_setup}

\noindent\textbf{Datasets and metrics.}
We evaluate on COCO \texttt{val}, Objects365v2 \texttt{val}~\cite{shao2019objects365} (mapped to COCO's 80-class label space for cross-domain transfer), and the synthetic STRANGE-Bench of Sec.~\ref{sec:strange}. On COCO and O365, we report overall performance as a sanity check (typicality-bias mitigation should not regress overall accuracy) and the typicality gap $\Delta^{\text{T-A}} = M^{\text{T}} - M^{\text{A}}$ on the per-anchor NPMI \emph{quintile} ($\alpha = 0.2$) splits of Sec.~\ref{sec:diagnostic}, where $M$ is AP for detection and micro-averaged F1 at a $0.5$ threshold (F1$_{\text{micro}}$) for DINOv3 classification. We compare rows within a family only, since the metrics (AP, F1) and their operating ranges differ. On STRANGE-Bench, we report the per-tier average $T_{1\text{-}5}^{\text{T}}, T_{1\text{-}5}^{\text{A}}$ across tiers $T_1$ to $T_5$ ($k = 0 \dots 4$) and the single-point $T_{\text{never}}$ score, the most extreme atypical case by construction (see Appendix~\ref{app:additional_results} for more details). The main paper uses $11$ of the $17$ sub-tiers ($T_1$ to $T_5$ in both arms and $T_{\text{never}}$); Tab.~\ref{tab:codino_extra_tiers} and Fig.~\ref{fig:strange_diagnostic_t6} report the others.

\begin{wrapfigure}{r}{0.50\linewidth}
    \vspace{-10pt}
    \centering
    \includegraphics[width=\linewidth]{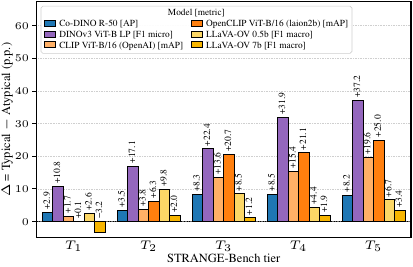}
    \captionsetup{skip=3pt}
    \caption{\textbf{Typicality bias is universal across modern vision models.} Per-tier typicality gap $\Delta = M^{\text{T}} - M^{\text{A}}$ on STRANGE-Bench for six models. The gap grows at every tier for the DINOv3 ViT-B linear probe (up to $\Delta = 37.2$ at T5), CLIP and OpenCLIP. Co-DINO R-50 rises to about $8.0$ by T3 and stays there. The LLaVA-OV gaps (F1 macro) are smaller.}
    \label{fig:strange_diagnostic}
    \vspace{-10pt}
\end{wrapfigure}

\noindent\textbf{Models.} We evaluate typicality bias on a representative model from each of five families spanning the modern vision stack: \emph{specialist detection} (Co-DINO~\cite{zong2023detrs} R-50, Swin-L), \emph{open-vocabulary detection} (Grounding DINO~\cite{liu2024grounding} Swin-B), \emph{self-supervised classification} (DINOv3~\cite{simeoni2025dinov3} ViT-B, both linear-probe and fine-tuned), \emph{vision--language models} (CLIP~\cite{radford2021learning} ViT-B/16, OpenCLIP~\cite{cherti2023reproducible} ViT-B/16 LAION-2B), and \emph{multimodal LLMs} (LLaVA-OneVision-7B~\cite{li2024llava}). 
For the diagnostic survey, we use the public pretrained checkpoints officially released by the authors. These families span the modern vision stack along two axes: \emph{supervision regime} (fully supervised, language-grounded, self-supervised, contrastive image--text, and instruction-tuned) and \emph{pretraining data scale} (from $\sim 100$K images for COCO-trained specialists to $\sim 2$ billion image--text pairs for OpenCLIP and LLaVA-OneVision). Any persistence of bias across this set is therefore unlikely to be an artifact of a single paradigm or scale.

For the systematic mitigation study, we select one representative from each of the three task settings, spanning supervision regime, architecture, and task: \textbf{Co-DINO R-50} (specialist closed-vocabulary detection), \textbf{Grounding DINO Swin-B} (open-vocabulary, text-conditioned), and \textbf{DINOv3 ViT-B/16 fine-tuned} (SSL foundational backbone fine-tuned for classification). All representative models are trained on COCO~\texttt{train2017} (initialized from the public pretrained checkpoints) under the augmentation variants below. Full training recipes, checkpoint sources, hyperparameters, and compute budgets are provided in Appendix~\ref{app:experimental-details}. We leave vision--language models and multimodal LLMs out of the mitigation study, since their co-occurrence prior is set during web-scale pretraining.

\noindent\textbf{Augmentation suite.}
We compare AAS against a representative suite of seven augmentations
spanning two axes, \emph{static}/\emph{dynamic} and \emph{non-generative}/\emph{generative}: \textbf{Color jitter}, \textbf{MixUp}~\cite{zhang2017mixup}, \textbf{Mosaic}~\cite{bochkovskiy2020yolov4}, \textbf{YOLOX}~\cite{ge2021yolox} (MixUp + Mosaic), \textbf{SCP}~\cite{ghiasi2021simple}, \textbf{XPaste}~\cite{zhao2023x}, and \textbf{InstAug}~\cite{kupyn2024dataset}. We additionally report a \textbf{Std.} baseline using only standard augmentation (flip, crop, resize). None of the existing augmentations are designed to break co-occurrence shortcuts; whether they incidentally affect typicality bias is an empirical question we answer below. AAS hyperparameters are selected once on a $25\%$ COCO subset and held fixed across architectures and tasks; per-augmentation hyperparameters are detailed in Appendix~\ref{app:exp-recipes}.

\subsection{Typicality Bias Persists Across the Modern Vision Stack}
\label{sec:results_diagnostic}
\vspace{-3.0pt}
To isolate the bias contained in the \emph{pretrained representation}
from any introduced by downstream task adaptation, we evaluate each family with the pretrained representation \emph{frozen}: zero-shot for Grounding DINO, CLIP, and OpenCLIP; linear probing for DINOv3; and
the released checkpoint for Co-DINO. We
survey the five model families on the COCO quintile split
(Fig.~\ref{fig:teaser}, right) and on the STRANGE-Bench per-tier breakdown (Fig.~\ref{fig:strange_diagnostic}).
The COCO quintile gap $\Delta^{\text{T-A}}$ is non-zero for every family we examine, ranging from \textbf{5.4}~p.p.\@ (Co-DINO Swin-L $1\times$) to \textbf{11.0}~p.p.\@ (OpenCLIP ViT-B/16); concretely, specialist detection shows $\Delta^{\text{T-A}}_{\text{AP}} = $ \textbf{9.5}~p.p.\@
(Co-DINO R-50), \textbf{5.4}~p.p.\@ (Swin-L $1\times$), and \textbf{5.6}~p.p.\@
(Swin-L $3\times$), open-vocabulary detection
$\Delta^{\text{T-A}}_{\text{AP}} = $ \textbf{6.3}~p.p.\@ (Grounding DINO
Swin-B), self-supervised classification
$\Delta^{\text{T-A}}_{\text{F1}}$ $= $ \textbf{8.6}~p.p.\@ (DINOv3 ViT-B
linear probe), and vision--language models
$\Delta^{\text{T-A}}_{\text{mAP}} = $ \textbf{9.5}~p.p.\@ (CLIP ViT-B/16)
and \textbf{11.0}~p.p.\@ (OpenCLIP ViT-B/16 LAION-2B). The Co-DINO Swin-L checkpoint, pretrained on a multi-dataset blend~\cite{zong2023detrs}, also exhibits $\Delta^{\text{T-A}}_{\text{AP}} = $ \textbf{5.7}~p.p.\@ on COCO val, indicating the bias is a property of the training distribution rather than dataset overfitting. From CLIP (400M image-text pairs) to OpenCLIP (LAION-2B), more pretraining data \emph{widens} rather than shrinks the gap (\textbf{9.5} $\rightarrow$ \textbf{11.0}~p.p.\@); scale alone is not a solution (App.~\ref{app:scale_ladder} separates data from capacity).
STRANGE-Bench amplifies this picture (Fig.~\ref{fig:strange_diagnostic}): the gap grows with contextual load $k$ at every tier for the DINOv3 linear probe, CLIP and OpenCLIP, reaching $\Delta = 37.2$ for the DINOv3 linear probe at the highest tier; Co-DINO R-50 rises to about $8.0$ by T3 and then stays flat, and the LLaVA-OV gaps (F1 macro) are smaller and do not grow with $k$. Per-tier breakdowns are provided in App.~\ref{app:additional_results}.

\begin{finding}
\textbf{Diagnosis.} Typicality bias persists across all five model families and four orders of magnitude of pretraining data ($\sim$100K $\rightarrow$ $\sim$2B images); scale and architectural diversity alone do not solve it.
\end{finding}

\subsection{Anchor-Atypicality Sampling Closes the Typicality Gap}
\label{sec:results_mitigation}

We evaluate the augmentation suite on the three representative families. An augmentation mitigates typicality bias when it lifts the atypical side while the typical side and overall accuracy hold; a smaller gap that comes from lowering both sides, as with Mosaic on Co-DINO (Tab.~\ref{tab:codino_main}), is not desirable. On STRANGE-Bench, the gap of the Std.\ baselines grows with the number of partners because the atypical side falls while the typical side changes little: from T1 to T5, Co-DINO goes from $90.8$ to $85.5$ AP on typical scenes and from $87.9$ to $77.3$ on atypical ones, DINOv3 from $91.3$ to $88.6$ F1 and from $92.0$ to $48.0$ (per-tier tables in App.~\ref{app:codino_full} and~\ref{app:dinov3_full}). Across all three, AAS preserves overall accuracy, closes
the typicality gap on the majority of (model, eval-split) cells, and
improves extreme-atypical performance the most.

\begin{table*}[t]
\centering
\small
\caption{\textbf{Co-DINO ResNet-50: typicality bias under augmentation
on COCO and STRANGE-Bench.} SCP + AAS \emph{(ours, blue)} has the highest COCO AP, and XPaste + AAS \emph{(ours, blue)} achieves
the smallest STRANGE-Bench tier gap, and the
highest never-co-occurring tier score; on COCO, Mosaic and YOLOX
attain the smallest typicality gaps, but do so by uniformly degrading both splits. Full-annotation bbox mAP (\%); $\Delta$ is typical minus atypical, $T_{1\text{-}5}$ the mean over tiers T1 to T5 of STRANGE-Bench. Best per column in \textbf{bold}, second-best \underline{underlined}.}
\vspace{-2pt}
\label{tab:codino_main}
\begin{tabular}{l cccc cccc}
\toprule
& \multicolumn{4}{c}{\textbf{COCO}} & \multicolumn{4}{c}{\textbf{STRANGE-Bench}} \\
\cmidrule(lr){2-5} \cmidrule(lr){6-9}
Augmentation & $\text{AP}^{\text{val}}$ $\uparrow$ & $\text{AP}^{\text{T}}$ $\uparrow$ & $\text{AP}^{\text{A}}$ $\uparrow$ & $\Delta^{\text{T-A}}$ $\downarrow$ & $T_{1\text{-}5}^{\text{T}}$ $\uparrow$ & $T_{1\text{-}5}^{\text{A}}$ $\uparrow$ & $\Delta_{1\text{-}5}$ $\downarrow$ & $T_{\text{never}}$ $\uparrow$ \\
\midrule
No aug & 43.5 & 44.2 & 35.5 & 8.7 & 83.4 & 76.4 & 7.0 & 76.3 \\
\midrule
\multicolumn{9}{l}{\emph{Non-generative}} \\
Std. & 51.8 & \underline{53.2} & 44.0 & 9.2 & 88.2 & 81.9 & 6.3 & 83.3 \\
Color jitter & 47.6 & 49.1 & 40.1 & 9.0 & 86.6 & 79.4 & 7.2 & 77.0 \\
MixUp & 46.2 & 47.2 & 39.1 & 8.1 & 84.0 & 77.0 & 7.0 & 78.0 \\
Mosaic & 45.8 & 46.1 & 39.7 & \textbf{6.4} & 72.2 & 68.0 & 4.2 & 69.3 \\
YOLOX & 46.7 & 47.0 & 40.3 & \underline{6.7} & 83.2 & 78.8 & 4.3 & 79.4 \\
SCP & \underline{52.1} & \underline{53.2} & 44.2 & 9.0 & 90.2 & 86.3 & 3.9 & 87.5 \\
\midrule
\multicolumn{9}{l}{\emph{Generative}} \\
InstAug & 51.7 & 52.6 & \underline{44.3} & 8.3 & 87.8 & 82.6 & 5.2 & 81.9 \\
XPaste & 50.9 & 52.0 & 43.0 & 9.0 & 90.7 & 89.0 & \underline{1.7} & \underline{90.4} \\
\midrule
\rowcolor{oursrow} SCP + AAS \emph{(ours)} & \textbf{52.9} & \textbf{54.0} & \textbf{45.5} & 8.5 & \underline{91.4} & \underline{89.2} & 2.1 & 90.1 \\
\rowcolor{oursrow} \textbf{XPaste + AAS \emph{(ours)}} & 51.4 & 52.1 & 43.3 & 8.8 & \textbf{91.7} & \textbf{90.7} & \textbf{1.0} & \textbf{92.1} \\
\bottomrule
\end{tabular}

\vspace{-8pt}
\end{table*}

\noindent\textbf{Supervised detection.}
On Co-DINO R-50 (Tab.~\ref{tab:codino_main}), SCP + AAS achieves the highest AP on COCO (overall: \textbf{52.9}, typical: \textbf{54.0}, atypical: \textbf{45.5}; $+1.1$~p.p.\@, $+0.8$~p.p.\@ and $+1.5$~p.p.\@ over Std.) and XPaste + AAS on STRANGE-Bench (typical: \textbf{91.7}, atypical: \textbf{90.7}, $T_{\text{never}}$: \textbf{92.1}; $+3.5$~p.p.\@, $+8.8$~p.p.\@ and $+8.8$~p.p.\@ over Std.). AAS lowers the COCO quintile gap of the baseline it starts from on
both sides (SCP \textbf{9.0}~$\rightarrow$ \textbf{8.5}, XPaste \textbf{9.0}~$\rightarrow$ \textbf{8.8}) and raises overall AP
($52.1 \rightarrow 52.9$ and $50.9 \rightarrow 51.4$). On STRANGE-Bench the change is larger with AAS:
$\Delta_{1\text{-}5}$ falls from \textbf{3.9} to \textbf{2.1} ($-1.8$~p.p.\@) for SCP and from \textbf{1.7} to \textbf{1.0} ($-0.7$~p.p.\@) for XPaste, and
$T_{\text{never}}$ rises from \textbf{87.5} to \textbf{90.1} ($+2.6$~p.p.\@) and from \textbf{90.4} to \textbf{92.1} ($+1.7$~p.p.\@).
Mosaic and YOLOX reach the smallest COCO gaps ($6.4$ and $6.7$) by lowering AP on both splits
($-3.7$ to $-7.1$~p.p.\@ vs.\ Std.). AAS closes the
gap by lifting atypical performance above its Std.\ baseline. Per-tier
breakdowns and results on the cross-domain Objects365 split are provided in Appendix~\ref{app:codino_full}.

\noindent\textbf{Open-vocabulary detection.}
Table~\ref{tab:grounding_dino_main} reports Grounding DINO Swin-B
under five configurations. Three observations stand out. First,
the zero-shot baseline already exhibits a substantial COCO quintile
gap ($\Delta^{\text{T-A}} = 6.3$~p.p.\@); fine-tuning on COCO with paste augmentation raises COCO val AP by $1.6$ to $2.0$~p.p.\@ but keeps the gap at $5.6$ to $6.2$. Second, AAS does not lower COCO val AP relative to the row it starts from ($+0.1$~p.p.\@ for SCP, $+0.2$~p.p.\@ for XPaste), with XPaste + AAS the highest (\textbf{58.9}), and keeps the COCO gap within $0.4$~p.p.\@ of that row ($6.2 \rightarrow 6.2$, $5.6 \rightarrow 6.0$). Third,
on STRANGE-Bench, AAS lifts atypical AP over the row it starts from ($+1.3$~p.p.\@ for SCP, $+1.1$~p.p.\@ for XPaste) and $T_{\text{never}}$ ($+1.2$~p.p.\@, $+0.6$~p.p.\@), and XPaste + AAS attains the highest typical (\textbf{95.0}), atypical (\textbf{94.7}) and $T_{\text{never}}$ (\textbf{95.2}) scores ($+1.7$~p.p.\@, $+3.8$~p.p.\@ and $+2.8$~p.p.\@ over zero-shot) and, with XPaste, the smallest mean tier gap ($\Delta_{1\text{-}5} = $ \textbf{0.3}~p.p.\@; zero-shot: $2.4$~p.p.\@). 

\begin{table*}[h]
\centering
\small
\caption{\textbf{Grounding DINO Swin-B: typicality bias under
augmentation on COCO and STRANGE-Bench.} XPaste + AAS \emph{(ours,
blue)} attains the highest COCO val AP and the highest
STRANGE-Bench scores.}
\label{tab:grounding_dino_main}
\vspace{-2pt}
\begin{tabular}{l cccc cccc}
\toprule
& \multicolumn{4}{c}{\textbf{COCO}} & \multicolumn{4}{c}{\textbf{STRANGE-Bench}} \\
\cmidrule(lr){2-5} \cmidrule(lr){6-9}
Augmentation & $\text{AP}^{\text{val}}$ $\uparrow$ & $\text{AP}^{\text{T}}$ $\uparrow$ & $\text{AP}^{\text{A}}$ $\uparrow$ & $\Delta^{\text{T-A}}$ $\downarrow$ & $T_{1\text{-}5}^{\text{T}}$ $\uparrow$ & $T_{1\text{-}5}^{\text{A}}$ $\uparrow$ & $\Delta_{1\text{-}5}$ $\downarrow$ & $T_{\text{never}}$ $\uparrow$ \\
\midrule
Zero Shot & 56.9 & 57.7 & 51.4 & 6.3 & 93.3 & 90.9 & 2.4 & 92.4 \\
\midrule
\multicolumn{9}{l}{\emph{Non-generative}} \\
SCP & 58.5 & 59.6 & 53.4 & 6.2 & 93.3 & 92.4 & 0.9 & 93.3 \\
\midrule
\multicolumn{9}{l}{\emph{Generative}} \\
XPaste & \underline{58.7} & \underline{59.7} & \textbf{54.1} & \textbf{5.6} & 93.9 & 93.6 & \textbf{0.3} & \underline{94.6} \\
\midrule
\rowcolor{oursrow} SCP + AAS \emph{(ours)} & 58.6 & \textbf{59.8} & 53.6 & 6.2 & \underline{94.3} & \underline{93.7} & \underline{0.6} & 94.5 \\
\rowcolor{oursrow} \textbf{XPaste + AAS \emph{(ours)}} & \textbf{58.9} & \underline{59.7} & \underline{53.7} & \underline{6.0} & \textbf{95.0} & \textbf{94.7} & \textbf{0.3} & \textbf{95.2} \\
\bottomrule
\end{tabular}

\vspace{-3.0pt}
\end{table*}

\begin{table*}[t]
\centering
\small
\caption{\textbf{DINOv3 ViT-B/16: typicality bias under augmentation
on COCO and STRANGE-Bench.} XPaste + AAS \emph{(ours, blue)} achieves
the highest scores on both datasets and the smallest STRANGE-Bench typicality gap, with the largest gains over Std.\ on atypical scenes ($+2.5$~p.p.\@ on COCO, $+28.0$~p.p.\@ on STRANGE-Bench and $+33.5$~p.p.\@ on $T_{\text{never}}$); its COCO gap ($3.9$) is second to SCP and YOLOX ($3.7$). F1$_{\text{micro}}$ (\%), columns as in Tab.~\ref{tab:codino_main}.}
\label{tab:dinov3_main}
\vspace{-2pt}
\begin{tabular}{l cccc cccc}
\toprule
& \multicolumn{4}{c}{\textbf{COCO}} & \multicolumn{4}{c}{\textbf{STRANGE-Bench}} \\
\cmidrule(lr){2-5} \cmidrule(lr){6-9}
Augmentation & $\text{F1}^{\text{val}}$ $\uparrow$ & $\text{F1}^{\text{T}}$ $\uparrow$ & $\text{F1}^{\text{A}}$ $\uparrow$ & $\Delta^{\text{T-A}}$ $\downarrow$ & $T_{1\text{-}5}^{\text{T}}$ $\uparrow$ & $T_{1\text{-}5}^{\text{A}}$ $\uparrow$ & $\Delta_{1\text{-}5}$ $\downarrow$ & $T_{\text{never}}$ $\uparrow$ \\
\midrule
Std.\ \emph{(linear probe)} & 74.6 & 71.8 & 63.2 & 8.6 & 63.5 & 39.6 & 23.9 & 38.9 \\
\midrule
No aug & 78.9 & 77.8 & 70.1 & 7.7 & 85.5 & 66.3 & 19.2 & 61.5 \\
\midrule
\multicolumn{9}{l}{\emph{Non-generative}} \\
Std. & 81.9 & 79.8 & 74.3 & 5.5 & 89.7 & 67.5 & 22.2 & 62.6 \\
Color jitter & 81.8 & 79.7 & 75.1 & 4.6 & 89.2 & 66.7 & 22.5 & 62.7 \\
MixUp & \underline{82.2} & 79.6 & 75.4 & 4.2 & 89.9 & 72.4 & 17.5 & 69.9 \\
Mosaic & 69.6 & 71.0 & 67.0 & 4.0 & 74.7 & 68.0 & 6.7 & 70.6 \\
YOLOX & 70.0 & 71.2 & 67.4 & \textbf{3.7} & 75.1 & 69.1 & 6.0 & 70.7 \\
SCP & 81.6 & 78.8 & 75.1 & \textbf{3.7} & 91.4 & 87.1 & 4.3 & 86.5 \\
\midrule
\multicolumn{9}{l}{\emph{Generative}} \\
InstAug & 81.9 & 79.5 & 74.4 & 5.2 & 89.5 & 67.7 & 21.8 & 63.3 \\
XPaste & 82.1 & \underline{80.2} & \underline{76.1} & 4.1 & \underline{93.8} & \underline{94.4} & \underline{$-0.7$} & \underline{95.0} \\
\midrule
\rowcolor{oursrow} SCP + AAS \emph{(ours)} & 81.7 & 79.0 & 75.0 & 4.0 & 92.5 & 91.2 & 1.2 & 89.5 \\
\rowcolor{oursrow} \textbf{XPaste + AAS \emph{(ours)}} & \textbf{82.5} & \textbf{80.6} & \textbf{76.8} & \underline{3.9} & \textbf{94.6} & \textbf{95.5} & \textbf{\boldmath $-0.8$} & \textbf{96.1} \\
\bottomrule
\end{tabular}

\vspace{-10pt}
\end{table*}

\noindent\textbf{SSL classification.}
Table~\ref{tab:dinov3_main} reports the full augmentation suite on
DINOv3 ViT-B/16 fine-tuned, with the linear probe shown in the top
row for reference. AAS yields the largest mitigation effect of the
three families: XPaste+AAS reduces the COCO quintile gap from
$\Delta^{\text{T-A}} = $ \textbf{5.5} (Std.) to \textbf{3.9} ($-1.6$
p.p.\@, a 29\% relative reduction), while improving overall F1
($81.9 \rightarrow 82.5$, $+0.6$ p.p.\@) and lifting the atypical-split
score by $+2.5$ p.p.\@ ($74.3 \rightarrow 76.8$). On STRANGE-Bench, the
gap is essentially eliminated: $\Delta_{1\text{-}5}$ collapses from
\textbf{22.2} to \textbf{$-0.8$} ($-23.0$ p.p.\@), and $T_{\text{never}}$
rises from $62.6$ to \textbf{96.1} ($+33.5$ p.p.\@). The non-paste variants
either leave the gap large (Color jitter, MixUp, InstAug) or reduce
it primarily by uniform degradation (Mosaic, YOLOX collapse overall
F1 by $\sim 12$ p.p.\@), while SCP and XPaste already narrow it (COCO $3.7$ and $4.1$, $\Delta_{1\text{-}5}$ $4.3$ and $-0.7$).
The situation differs for linear probing on DINOv3 ViT-B/16 with the backbone frozen, where augmentations operate only on the head input and their effect is muted; we report and discuss these results in Appendix~\ref{app:dinov3_lp}.

\begin{finding}
\textbf{Mitigation.} AAS closes the typicality gap by up to
$\mathbf{1.6}$ p.p.\@ on COCO splits and $\mathbf{23.0}$ p.p.\@ on
STRANGE-Bench, while preserving overall performance: the
intervention is asymmetric, lifting atypical-context performance
without degrading typical-context performance.
\end{finding}

\subsection{The Typicality Gap Is Not a Rendering Artifact}
\label{sec:results_human}
\vspace{-3.0pt}
\begin{wraptable}{r}{0.33\linewidth}
\vspace{-11pt}
\centering\footnotesize
\setlength{\tabcolsep}{3pt}
\captionsetup{font=small}
\caption{\textbf{Free listing}: anchor recall (\%); models: DINOv3 fine-tuned.}
\label{tab:human_listing}
\begin{tabular}{lccc}
\toprule
Rater & $T_1$ & $T_{\text{never}}$ & $\Delta$ \\
\midrule
Humans & $98.3$ & $97.3$ & $-1.0$ \\
\midrule
Std. & $98.0$ & $67.0$ & $-31.0$ \\
XPaste & $100.0$ & $96.0$ & $-4.0$ \\
\rowcolor{oursrow} XPaste + AAS & $100.0$ & $97.0$ & $-3.0$ \\
\bottomrule
\end{tabular}

\vspace{-10pt}
\end{wraptable}
Three blinded user studies test whether the gaps on STRANGE-Bench come from the renders or from the models (App.~\ref{app:strange:user-study}); they ran on an initial version of the benchmark with one render per scene and a lighter audit. In \emph{free listing}, raters named every category they could identify in $100$ $T_1$ and $100$ $T_{\text{never}}$ scenes shown without boxes. A rendering defect would hide the anchor from humans and models alike, yet human anchor recall drops by $1.0$~p.p.\@ and that of the DINOv3 Std.\ model by $31.0$~p.p.\@ (Tab.~\ref{tab:human_listing}). In a \emph{quality audit} of $330$ scenes, typical and atypical scenes at the same $k$ differ in pass rate by $2.2$~p.p.\@ on average over four criteria, with no systematic direction.

\section{Conclusion}
\vspace{-8pt}
We revisited \emph{typicality bias}, the tendency of recognition models to anchor predictions on familiar co-occurring context rather than the object itself. Across the modern vision stack, the bias persists across diverse model families and pretraining scales, without consistently shrinking with size. To diagnose it on a continuous scale, we introduced an NPMI-based diagnostic and the photorealistic STRANGE-Bench. To mitigate it, we further proposed Anchor-Atypicality Sampling (AAS), a sampling rule for paste-based augmentation that closes the typicality gap by lifting atypical-context performance without degrading typical-context performance.

Our diagnostic relies on annotated co-occurrence statistics, and AAS requires training pipelines that allow for paste-based augmentations, leaving contrastive image-text and instruction-following pretraining, where the bias is most pronounced, as open directions for future work. These findings suggest that contextual robustness should be reported alongside in-distribution performance in the standard evaluation of foundation-scale recognition models.

\clearpage
\begin{ack}
We thank Wolfgang Boettcher for valuable feedback on the manuscript. We would like to thank and acknowledge the Max Planck Computing and Data Facility (MPCDF), whose HPC systems supported a substantial portion of our experiments, including the development of the proposed method, reproduction of baselines, and computation of the final results. Siddhartha Gairola's research was supported in part by ELSA -- European Lighthouse on Secure and Safe AI, funded by the European Union under grant agreement No.~101070617. Views and opinions expressed are, however, those of the authors only and do not necessarily reflect those of the European Union or European Commission. Neither the European Union nor the European Commission can be held responsible for them. The authors do not have any conflicts to disclose for this work.
\end{ack}

{
    \small
    \bibliographystyle{plainnat}
    \bibliography{references}
}

\clearpage
\appendix
\section*{Appendix}
\addcontentsline{toc}{section}{Appendix}
\setlength{\intextsep}{4pt plus 1pt minus 1pt}
\setlength{\textfloatsep}{6pt plus 2pt minus 2pt}
\setlength{\floatsep}{6pt plus 2pt minus 2pt}
\raggedbottom
\makeatletter\setlength{\@fptop}{0pt}\setlength{\@fpsep}{6pt plus 2pt minus 2pt}\setlength{\@fpbot}{0pt plus 1fil}\makeatother
\captionsetup{font=small}

This appendix supplements the main paper with: per-family detailed result breakdowns and the AAS pseudocode (\S\ref{app:additional_results}); a complete description of the experimental setup, including datasets, models, training and evaluation recipes, and compute (\S\ref{app:exp-setup}); the STRANGE-Bench generation pipeline, statistics, and qualitative samples (\S\ref{app:strange-pipeline}); and a discussion of limitations and broader impacts (\S\ref{app:discussion}).

\noindent\textbf{Table of Contents}
\begin{itemize}[leftmargin=*,itemsep=8pt,topsep=6pt,parsep=0pt]
  \item[\textbf{A}] \textbf{\hyperref[app:additional_results]{Additional Quantitative Results}} \dotfill\pageref{app:additional_results}
    \begin{itemize}[leftmargin=2em,itemsep=2pt,topsep=3pt,parsep=0pt,label={}]
      \item[A.1] \hyperref[app:codino_full]{Co-DINO ResNet-50: Full Results} \dotfill\pageref{app:codino_full}
      \item[A.2] \hyperref[app:gdino_full]{Grounding DINO Swin-B: Full Results} \dotfill\pageref{app:gdino_full}
      \item[A.3] \hyperref[app:dinov3_full]{DINOv3 ViT-B/16 Fine-Tuned: Full Results} \dotfill\pageref{app:dinov3_full}
      \item[A.4] \hyperref[app:dinov3_lp]{DINOv3 ViT-B/16 Linear Probe: Full Results} \dotfill\pageref{app:dinov3_lp}
      \item[A.5] \hyperref[app:beta_sweep]{Selecting the AAS Atypicality Temperature $\beta$} \dotfill\pageref{app:beta_sweep}
      \item[A.6] \hyperref[app:circularity]{Circularity Checks} \dotfill\pageref{app:circularity}%
      \item[A.7] \hyperref[app:scale_ladder]{Pretraining Data and Model Capacity} \dotfill\pageref{app:scale_ladder}%
      \item[A.8] \hyperref[app:spatial-arrangement]{Spatial Arrangement as a Typicality Axis} \dotfill\pageref{app:spatial-arrangement}%
      \item[A.9] \hyperref[app:partner_mask]{Partner Masking} \dotfill\pageref{app:partner_mask}%
      \item[A.10] \hyperref[app:aas_draws]{Variability of the AAS Rows over Training Runs} \dotfill\pageref{app:aas_draws}%
    \end{itemize}

  \item[\textbf{B}] \textbf{\hyperref[app:exp-setup]{Experimental Setup and Design}} \dotfill\pageref{app:exp-setup}
    \begin{itemize}[leftmargin=2em,itemsep=2pt,topsep=3pt,parsep=0pt,label={}]
      \item[B.1] \hyperref[app:experimental-details]{Experimental Details and Study} \dotfill\pageref{app:experimental-details}
        \begin{itemize}[leftmargin=2em,itemsep=1pt,topsep=2pt,parsep=0pt,label={}]
          \item[B.1.1] \hyperref[app:exp-datasets]{Evaluation Datasets} \dotfill\pageref{app:exp-datasets}
          \item[B.1.2] \hyperref[app:exp-models]{Models} \dotfill\pageref{app:exp-models}
          \item[B.1.3] \hyperref[app:exp-ckpts]{Checkpoints} \dotfill\pageref{app:exp-ckpts}
          \item[B.1.4] \hyperref[app:exp-recipes]{Training and Evaluation Recipes} \dotfill\pageref{app:exp-recipes}
          \item[B.1.5] \hyperref[app:exp-compute]{Compute} \dotfill\pageref{app:exp-compute}
        \end{itemize}
      \item[B.2] \hyperref[app:splits-impl]{Quantile Choices for the Paired Diagnostic} \dotfill\pageref{app:splits-impl}
    \end{itemize}

  \item[\textbf{C}] \textbf{\hyperref[app:strange-pipeline]{STRANGE-Bench: Pipeline Details and Statistics}} \dotfill\pageref{app:strange-pipeline}
    \begin{itemize}[leftmargin=2em,itemsep=2pt,topsep=3pt,parsep=0pt,label={}]
      \item[C.1] \hyperref[app:strange:release]{Dataset Release} \dotfill\pageref{app:strange:release}
      \item[C.2] \hyperref[app:strange:samples]{Qualitative Samples} \dotfill\pageref{app:strange:samples}
      \item[C.3] \hyperref[app:strange:tiers]{Sub-Tier Composition} \dotfill\pageref{app:strange:tiers}
      \item[C.4] \hyperref[app:strange:diag]{Sanity Diagnostics} \dotfill\pageref{app:strange:diag}
      \item[C.5] \hyperref[app:strange:pipeline]{Generation Pipeline} \dotfill\pageref{app:strange:pipeline}
      \item[C.6] \hyperref[app:strange:impl]{Implementation Details} \dotfill\pageref{app:strange:impl}
      \item[C.7] \hyperref[app:strange:user-study]{User Study} \dotfill\pageref{app:strange:user-study}%
    \end{itemize}

  \item[\textbf{D}] \textbf{\hyperref[app:discussion]{Limitations and Broader Impacts}} \dotfill\pageref{app:discussion}
\end{itemize}

\clearpage

\section{Additional Quantitative Results}
\label{app:additional_results}
\suppressfloats[t]
\noindent In this section, we report the full results behind Sec.~\ref{sec:experiments} and further analyses. We first give the full results of each family in App.~\ref{app:codino_full} to~\ref{app:dinov3_lp}: the Objects365 cross-domain splits and the per-tier STRANGE-Bench breakdown for the rows of Tabs.~\ref{tab:codino_main} to~\ref{tab:dinov3_main}, and the DINOv3 linear probe. We then show the empirical selection of AAS temperature $\beta$ in App.~\ref{app:beta_sweep} and test in App.~\ref{app:circularity} whether the gaps and the AAS gains depend on COCO NPMI or on the generator of the benchmark. We finally separate pretraining data from model capacity in App.~\ref{app:scale_ladder}, test whether models also rely on the usual spatial arrangement of a pair in App.~\ref{app:spatial-arrangement}, evaluate a target-focused baseline that inpaints the partners away in App.~\ref{app:partner_mask}, and report the spread of the AAS rows over training runs in App.~\ref{app:aas_draws}. The evaluation datasets and paired splits are defined in App.~\ref{app:exp-datasets}, and the training and evaluation recipes in App.~\ref{app:exp-recipes}.

\subsection{Co-DINO ResNet-50: Full Results}
\label{app:codino_full}

We complement the main Co-DINO ResNet-50 results (Tab.~\ref{tab:codino_main}) with the cross-domain transfer table (Tab.~\ref{tab:codino_cd}) and the full per-tier STRANGE-Bench breakdown (Tab.~\ref{tab:codino_strange}). All numbers use the same quintile cut ($\alpha = 0.2$) as the main paper. The augmentation suite (No aug, Std., Color jitter, MixUp, Mosaic, YOLOX, SCP, InstAug, XPaste) is identical to that of Sec.~\ref{sec:experiments}, with SCP\,+\,AAS and XPaste\,+\,AAS as the two AAS variants under study. 

\begin{table*}[!htp]%
\centering
\footnotesize
\setlength{\tabcolsep}{4pt}
\caption{\textbf{Co-DINO ResNet-50: in-domain (COCO) and cross-domain
(Objects365) typicality.} SCP\,+\,AAS \emph{(ours)} has the highest typical and atypical AP on all three splits. On Objects365 the gap hardly depends on the augmentation ($2.0$ to $3.0$~p.p.\@ with COCO NPMI) and is $1.6$ to $2.3$ times larger for every row when the pairs are ranked by Objects365's own NPMI ($3.8$ to $5.1$~p.p.\@). Full-annotation bbox mAP (\%) on the quintile paired splits of COCO val and Objects365 val, the latter split by COCO \texttt{train2017} NPMI or by NPMI recomputed on Objects365 train (App.~\ref{app:exp-datasets}); rows as in Tab.~\ref{tab:codino_main}. Best per column in \textbf{bold}, second-best \underline{underlined}.}
\label{tab:codino_cd}
\begin{tabular}{l cccc ccc ccc}
\toprule
& \multicolumn{4}{c}{\textbf{COCO}} & \multicolumn{3}{c}{\textbf{Objects365, COCO NPMI}} & \multicolumn{3}{c}{\textbf{Objects365, O365 NPMI}} \\
\cmidrule(lr){2-5} \cmidrule(lr){6-8} \cmidrule(lr){9-11}
Augmentation & $\text{AP}^{\text{val}}$ $\uparrow$ & $\text{AP}^{\text{T}}$ $\uparrow$ & $\text{AP}^{\text{A}}$ $\uparrow$ & $\Delta^{\text{T-A}}$ $\downarrow$ & $\text{AP}^{\text{T}}$ $\uparrow$ & $\text{AP}^{\text{A}}$ $\uparrow$ & $\Delta^{\text{T-A}}$ $\downarrow$ & $\text{AP}^{\text{T}}$ $\uparrow$ & $\text{AP}^{\text{A}}$ $\uparrow$ & $\Delta^{\text{T-A}}$ $\downarrow$ \\
\midrule
No aug & 43.5 & 44.2 & 35.5 & 8.7 & 27.1 & 25.1 & \textbf{2.0} & 28.4 & 24.6 & \textbf{3.8} \\
\midrule
\multicolumn{11}{l}{\emph{Non-generative}} \\
Std. & 51.8 & \underline{53.2} & 44.0 & 9.2 & \underline{34.3} & \underline{31.9} & 2.4 & \underline{36.0} & \underline{31.4} & 4.6 \\
Color jitter & 47.6 & 49.1 & 40.1 & 9.0 & 30.8 & 27.8 & 3.0 & 32.1 & 27.2 & 4.9 \\
MixUp & 46.2 & 47.2 & 39.1 & 8.1 & 30.1 & 28.1 & \textbf{2.0} & 31.6 & 27.4 & 4.2 \\
Mosaic & 45.8 & 46.1 & 39.7 & \textbf{6.4} & 30.2 & 27.9 & 2.3 & 31.9 & 27.4 & 4.5 \\
YOLOX & 46.7 & 47.0 & 40.3 & \underline{6.7} & 30.8 & 28.5 & 2.3 & 32.3 & 28.2 & \underline{4.1} \\
SCP & \underline{52.1} & \underline{53.2} & 44.2 & 9.0 & 34.2 & 31.7 & 2.5 & 35.5 & 31.1 & 4.4 \\
\midrule
\multicolumn{11}{l}{\emph{Generative}} \\
InstAug & 51.7 & 52.6 & \underline{44.3} & 8.3 & 33.6 & 31.3 & 2.3 & 35.6 & 30.5 & 5.1 \\
XPaste & 50.9 & 52.0 & 43.0 & 9.0 & 32.9 & 30.8 & \underline{2.1} & 34.7 & 29.9 & 4.8 \\
\midrule
\rowcolor{oursrow} SCP + AAS \emph{(ours)} & \textbf{52.9} & \textbf{54.0} & \textbf{45.5} & 8.5 & \textbf{34.7} & \textbf{32.4} & 2.3 & \textbf{36.4} & \textbf{32.1} & 4.3 \\
\rowcolor{oursrow} \textbf{XPaste + AAS \emph{(ours)}} & 51.4 & 52.1 & 43.3 & 8.8 & 33.2 & 31.2 & \textbf{2.0} & 35.1 & 30.5 & 4.6 \\
\bottomrule
\end{tabular}

\end{table*}

\noindent\textbf{In-domain and cross-domain typicality.} On COCO (Tab.~\ref{tab:codino_cd}, left block), Std.\ aug yields the largest typicality gap of all rows ($\Delta^{\text{T-A}} = 9.2$), which is consistent with the $\alpha$-sweep finding in \S\ref{app:splits-impl} that conventional augmentation does not close the gap. Mosaic and YOLOX attain the smallest COCO gaps ($6.4$ and $6.7$) but lower AP on both splits relative to Std.\ (typical AP by $7.1$ and $6.2$~p.p.\@). SCP\,+\,AAS \emph{(ours)} has the highest overall AP ($\text{AP}^{\text{val}} = 52.9$) and the highest typical and atypical AP, with $\Delta^{\text{T-A}} = 8.5$; XPaste\,+\,AAS has $\text{AP}^{\text{val}} = 51.4$ and $\Delta^{\text{T-A}} = 8.8$. On cross-domain Objects365 (right blocks), the gaps of all rows lie between $2.0$ and $3.0$ with COCO NPMI and between $3.8$ and $5.1$ with O365 NPMI, so no configuration stands out; SCP\,+\,AAS again has the highest typical and atypical AP in both blocks ($34.7 / 32.4$ and $36.4 / 32.1$), and each AAS row lowers the gap of the row it starts from by $0.1$ to $0.2$.

\begin{table*}[!htp]%
\centering
\scriptsize
\setlength{\tabcolsep}{2pt}
\caption{\textbf{Co-DINO ResNet-50: STRANGE-Bench per-tier breakdown.}
For No aug, Std., Color jitter and MixUp the gap jumps once a scene holds two partners ($8$ to $11$~p.p.\@ at T3 to T5); the paste rows (SCP, XPaste and both AAS rows) keep it at $5.4$~p.p.\@ or below, and XPaste\,+\,AAS \emph{(ours)} has the smallest gap at T1 to T4 and the highest $T_{\text{never}}$. Full-annotation bbox mAP (\%) of the typical (T) and atypical (A) arm of each tier and their gap $\Delta$; $T_{1\text{-}5}$ is the mean over T1 to T5, rows as in Tab.~\ref{tab:codino_main}. Best per column in \textbf{bold}, second-best \underline{underlined}.}
\label{tab:codino_strange}
\begin{tabular}{l ccc ccc ccc ccc ccc ccc c}
\toprule
& \multicolumn{3}{c}{\textbf{Tier 1}} & \multicolumn{3}{c}{\textbf{Tier 2}} & \multicolumn{3}{c}{\textbf{Tier 3}} & \multicolumn{3}{c}{\textbf{Tier 4}} & \multicolumn{3}{c}{\textbf{Tier 5}} & \multicolumn{3}{c}{\textbf{Mean T1--T5}} & \\
\cmidrule(lr){2-4} \cmidrule(lr){5-7} \cmidrule(lr){8-10} \cmidrule(lr){11-13} \cmidrule(lr){14-16} \cmidrule(lr){17-19}
Aug & T$\uparrow$ & A$\uparrow$ & $\Delta\downarrow$ & T$\uparrow$ & A$\uparrow$ & $\Delta\downarrow$ & T$\uparrow$ & A$\uparrow$ & $\Delta\downarrow$ & T$\uparrow$ & A$\uparrow$ & $\Delta\downarrow$ & T$\uparrow$ & A$\uparrow$ & $\Delta\downarrow$ & T$\uparrow$ & A$\uparrow$ & $\Delta\downarrow$ & $T_{\text{never}}\uparrow$ \\
\midrule
No aug & 87.0 & 86.8 & \underline{0.2} & 85.8 & 81.5 & 4.3 & 83.5 & 72.6 & 10.9 & 81.7 & 71.2 & 10.5 & 79.1 & 69.9 & 9.2 & 83.4 & 76.4 & 7.0 & 76.3 \\
\midrule
\multicolumn{20}{l}{\emph{Non-generative}} \\
Std. & 90.8 & 87.9 & 2.9 & 90.5 & 87.0 & 3.5 & 87.3 & 79.0 & 8.3 & 86.9 & 78.4 & 8.5 & 85.5 & 77.3 & 8.2 & 88.2 & 81.9 & 6.3 & 83.3 \\
Color jitter & 90.1 & 88.4 & 1.7 & 88.1 & 84.7 & 3.4 & 86.3 & 75.9 & 10.4 & 84.9 & 74.3 & 10.6 & 83.8 & 73.8 & 10.0 & 86.6 & 79.4 & 7.2 & 77.0 \\
MixUp & 86.2 & 84.2 & 2.0 & 85.6 & 83.3 & 2.3 & 83.7 & 73.8 & 9.9 & 83.1 & 73.3 & 9.8 & 81.4 & 70.6 & 10.8 & 84.0 & 77.0 & 7.0 & 78.0 \\
Mosaic & 73.9 & 69.9 & 4.0 & 73.7 & 72.3 & 1.4 & 71.4 & 67.4 & 4.0 & 70.9 & 65.5 & 5.4 & 71.2 & 64.8 & 6.4 & 72.2 & 68.0 & 4.2 & 69.3 \\
YOLOX & 84.3 & 82.0 & 2.3 & 84.6 & 82.9 & 1.7 & 82.3 & 77.7 & 4.6 & 82.3 & 75.8 & 6.5 & 82.3 & 75.7 & 6.6 & 83.2 & 78.8 & 4.3 & 79.4 \\
SCP & 92.3 & 89.6 & 2.7 & 91.7 & 88.9 & 2.8 & 89.2 & 85.1 & 4.1 & 89.5 & 84.8 & 4.7 & 88.4 & 83.0 & 5.4 & 90.2 & 86.3 & 3.9 & 87.5 \\
\midrule
\multicolumn{20}{l}{\emph{Generative}} \\
InstAug & 90.3 & 89.3 & 1.0 & 90.0 & 87.4 & 2.6 & 87.2 & 80.2 & 7.0 & 86.3 & 78.5 & 7.8 & 85.2 & 77.7 & 7.5 & 87.8 & 82.6 & 5.2 & 81.9 \\
XPaste & 92.5 & \underline{91.7} & 0.8 & \underline{92.8} & 92.0 & 0.8 & 89.8 & 87.6 & \underline{2.2} & 90.0 & 87.6 & \underline{2.4} & 88.5 & \underline{86.3} & \textbf{2.2} & 90.7 & 89.0 & \underline{1.7} & \underline{90.4} \\
\midrule
\rowcolor{oursrow} SCP + AAS \emph{(ours)} & \textbf{93.7} & \underline{91.7} & 2.0 & 92.7 & \underline{92.3} & \underline{0.4} & \underline{90.5} & \underline{88.2} & 2.3 & \underline{90.6} & \underline{87.8} & 2.8 & \underline{89.4} & 86.2 & 3.2 & \underline{91.4} & \underline{89.2} & 2.1 & 90.1 \\
\rowcolor{oursrow} \textbf{XPaste + AAS \emph{(ours)}} & \underline{93.4} & \textbf{93.4} & \textbf{0.0} & \textbf{93.8} & \textbf{93.5} & \textbf{0.3} & \textbf{91.0} & \textbf{89.6} & \textbf{1.4} & \textbf{90.9} & \textbf{89.9} & \textbf{1.0} & \textbf{89.5} & \textbf{87.2} & \underline{2.3} & \textbf{91.7} & \textbf{90.7} & \textbf{1.0} & \textbf{92.1} \\
\bottomrule
\end{tabular}

\end{table*}

\begin{figure}[!htp]%
    \centering
    \includegraphics[width=\linewidth]{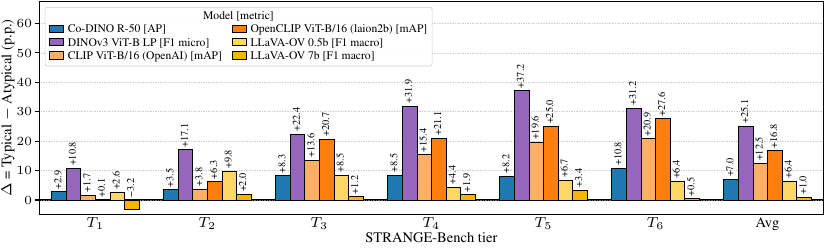}
    \caption{\textbf{Per-tier typicality gap including T6.} Fig.~\ref{fig:strange_diagnostic} with the five-partner tier T6 and an Avg group (each model's mean over T1 to T6) added.
 T6 layouts come from a rule-based layout solver, while those of T1 to T5 mostly come from Qwen, so T6 differs from the other tiers in layout source as well as in $k$.}
    \label{fig:strange_diagnostic_t6}
\end{figure}

\noindent Tab.~\ref{tab:codino_extra_tiers} and Fig.~\ref{fig:strange_diagnostic_t6} report the STRANGE-Bench sub-tiers outside the grid of the main tables (App.~\ref{app:strange:tiers}).
On T6, XPaste\,+\,AAS has the highest typical and atypical AP ($88.3 / 87.5$) and the smallest gap ($0.8$; XPaste $1.1$, SCP $5.8$, Std.\ $10.8$), and it has the highest score in every never-co-occurring column ($92.1 / 88.9 / 90.5$; tied with SCP\,+\,AAS at $k = 3$). On the small-anchor pair every gap stays within $\pm 2.2$, and SCP, XPaste and both AAS rows score the atypical arm higher ($\Delta = -0.2$ to $-1.9$).
\begin{table}[!htp]%
\centering
\small
\caption{\textbf{Co-DINO ResNet-50 on the STRANGE-Bench sub-tiers outside the main grid.} At five partners (T6) the gap is $8$ to $15$~p.p.\@ for the rows without pasting (No aug to YOLOX) and $0.8$~p.p.\@ for XPaste\,+\,AAS \emph{(ours)}, which also scores highest on every never-co-occurring column; on the small-anchor pair no gap exceeds $2.2$~p.p.\@ in absolute value. T6 has five partners; the small-anchor pair has one partner and an anchor box of at most 4\% of the image; the never-co-occurring columns are $T_{\text{never}}$ of Tab.~\ref{tab:codino_main} ($k = 1, 2$) and cliques of four and five classes with no pair co-occurring in COCO \texttt{train2017} ($k = 3, 4$). Full-annotation bbox mAP (\%), rows as in Tab.~\ref{tab:codino_main}; best per column in \textbf{bold}, second-best \underline{underlined}.}
\label{tab:codino_extra_tiers}
\resizebox{\linewidth}{!}{%
\begin{tabular}{l ccc ccc ccc}
\toprule
& \multicolumn{3}{c}{\textbf{T6} ($k{=}5$)} & \multicolumn{3}{c}{\textbf{Small anchor} ($k{=}1$)} & \multicolumn{3}{c}{\textbf{Never co-occurring}} \\
\cmidrule(lr){2-4} \cmidrule(lr){5-7} \cmidrule(lr){8-10}
Augmentation & T $\uparrow$ & A $\uparrow$ & $\Delta$ $\downarrow$ & T $\uparrow$ & A $\uparrow$ & $\Delta$ $\downarrow$ & $k{=}1,2$ $\uparrow$ & $k{=}3$ $\uparrow$ & $k{=}4$ $\uparrow$ \\
\midrule
No aug & 78.1 & 65.9 & 12.2 & 83.0 & 81.1 & 1.9 & 76.3 & 72.1 & 64.3 \\
\midrule
\multicolumn{10}{l}{\emph{Non-generative}} \\
Std. & 83.5 & 72.7 & 10.8 & 87.2 & 85.0 & 2.2 & 83.3 & 78.1 & 69.9 \\
Color jitter & 82.5 & 69.0 & 13.5 & 86.2 & 84.8 & 1.4 & 77.0 & 74.0 & 66.1 \\
MixUp & 81.2 & 65.9 & 15.3 & 82.4 & 82.6 & $-0.2$ & 78.0 & 73.9 & 67.6 \\
Mosaic & 69.7 & 61.4 & 8.3 & 71.9 & 71.2 & 0.7 & 69.3 & 63.5 & 61.9 \\
YOLOX & 80.5 & 71.6 & 8.9 & 84.1 & 84.0 & 0.1 & 79.4 & 77.3 & 72.1 \\
SCP & 85.9 & 80.1 & 5.8 & 88.8 & 89.0 & $-0.2$ & 87.5 & 83.0 & 82.9 \\
\midrule
\multicolumn{10}{l}{\emph{Generative}} \\
InstAug & 83.4 & 73.3 & 10.1 & 87.0 & 85.0 & 2.0 & 81.9 & 77.3 & 71.1 \\
XPaste & \underline{86.9} & \underline{85.8} & \underline{1.1} & \underline{89.8} & 91.0 & \underline{$-1.2$} & \underline{90.4} & \underline{86.1} & \underline{88.1} \\
\midrule
\rowcolor{oursrow} SCP + AAS \emph{(ours)} & 86.7 & 83.6 & 3.1 & 89.3 & \underline{91.2} & \textbf{\boldmath $-1.9$} & 90.1 & \textbf{88.9} & 87.3 \\
\rowcolor{oursrow} \textbf{XPaste + AAS \emph{(ours)}} & \textbf{88.3} & \textbf{87.5} & \textbf{0.8} & \textbf{91.0} & \textbf{92.2} & \underline{$-1.2$} & \textbf{92.1} & \textbf{88.9} & \textbf{90.5} \\
\bottomrule
\end{tabular}

}
\end{table}

\subsection{Grounding DINO Swin-B: Full Results}
\label{app:gdino_full}

As done for Co-DINO, we extend the main Grounding DINO Swin-B results (Tab.~\ref{tab:grounding_dino_main}) with Objects365 transfer (Tab.~\ref{tab:gdino_o365}) and a per-tier STRANGE-Bench breakdown (Tab.~\ref{tab:gdino_strange_full}), on the same quintile splits. The \emph{Zero Shot} row reports the released Grounding DINO Swin-B checkpoint, pretrained on a multi-dataset blend~\cite{liu2024grounding}, evaluated zero-shot on COCO and Objects365; SCP and XPaste fine-tune that checkpoint on COCO \texttt{train2017} for 12 epochs with pasted real, or real and generated, cutouts, and the AAS rows start from them (Sec.~\ref{sec:aas}).

\begin{table*}[!htp]%
\centering
\footnotesize
\setlength{\tabcolsep}{4pt}
\caption{\textbf{Grounding DINO Swin-B: in-domain (COCO) and
cross-domain (Objects365) typicality.}
Fine-tuning on COCO trades Objects365 AP for COCO AP: every fine-tuned row gains $1.6$ to $2.0$~p.p.\@ on COCO val and loses $0.9$ to $2.0$~p.p.\@ on the Objects365 splits, where the Zero Shot checkpoint stays best; AAS \emph{(ours)} wins back $0.1$ to $0.4$~p.p.\@ of that loss over the row it starts from. The gap stays small on all three splits ($5.6$ to $6.3$, $2.1$ to $2.7$ and $3.0$ to $3.5$~p.p.\@). Full-annotation bbox mAP (\%) on the quintile paired splits, Objects365 val split by COCO \texttt{train2017} NPMI or by NPMI recomputed on Objects365 train (App.~\ref{app:exp-datasets}); rows as in Tab.~\ref{tab:grounding_dino_main}. Best per column in \textbf{bold}, second-best \underline{underlined}.}
\label{tab:gdino_o365}
\begin{tabular}{l cccc ccc ccc}
\toprule
& \multicolumn{4}{c}{\textbf{COCO}} & \multicolumn{3}{c}{\textbf{Objects365, COCO NPMI}} & \multicolumn{3}{c}{\textbf{Objects365, O365 NPMI}} \\
\cmidrule(lr){2-5} \cmidrule(lr){6-8} \cmidrule(lr){9-11}
Augmentation & $\text{AP}^{\text{val}}$ $\uparrow$ & $\text{AP}^{\text{T}}$ $\uparrow$ & $\text{AP}^{\text{A}}$ $\uparrow$ & $\Delta^{\text{T-A}}$ $\downarrow$ & $\text{AP}^{\text{T}}$ $\uparrow$ & $\text{AP}^{\text{A}}$ $\uparrow$ & $\Delta^{\text{T-A}}$ $\downarrow$ & $\text{AP}^{\text{T}}$ $\uparrow$ & $\text{AP}^{\text{A}}$ $\uparrow$ & $\Delta^{\text{T-A}}$ $\downarrow$ \\
\midrule
Zero Shot & 56.9 & 57.7 & 51.4 & 6.3 & \textbf{46.5} & \textbf{43.8} & 2.7 & \textbf{47.6} & \textbf{44.1} & 3.5 \\
\midrule
\multicolumn{11}{l}{\emph{Non-generative}} \\
SCP & 58.5 & 59.6 & 53.4 & 6.2 & 44.5 & 42.4 & \textbf{2.1} & 45.9 & 42.9 & \textbf{3.0} \\
\midrule
\multicolumn{11}{l}{\emph{Generative}} \\
XPaste & \underline{58.7} & \underline{59.7} & \textbf{54.1} & \textbf{5.6} & 44.7 & 42.1 & 2.6 & 46.1 & 42.8 & \underline{3.3} \\
\midrule
\rowcolor{oursrow} SCP + AAS \emph{(ours)} & 58.6 & \textbf{59.8} & 53.6 & 6.2 & \underline{44.9} & \underline{42.5} & \underline{2.4} & 46.2 & \underline{43.2} & \textbf{3.0} \\
\rowcolor{oursrow} \textbf{XPaste + AAS \emph{(ours)}} & \textbf{58.9} & \underline{59.7} & \underline{53.7} & \underline{6.0} & 44.8 & 42.4 & \underline{2.4} & \underline{46.3} & 43.0 & \underline{3.3} \\
\bottomrule
\end{tabular}

\end{table*}

\noindent\textbf{In-domain and cross-domain typicality.} On COCO (Tab.~\ref{tab:gdino_o365}, left block), fine-tuning raises $\text{AP}^{\text{val}}$ from $56.9$ (Zero Shot) to between $58.5$ and $58.9$, while the gap stays between $5.6$ and $6.3$ for every row. XPaste has the smallest gap ($5.6$) and the highest atypical AP ($54.1$), XPaste\,+\,AAS the highest $\text{AP}^{\text{val}}$ ($58.9$) and SCP\,+\,AAS the highest typical AP ($59.8$); AAS leaves the SCP gap at $6.2$ and raises the XPaste gap to $6.0$. On Objects365 (right blocks), the Zero Shot checkpoint is best on every typical and atypical column: fine-tuning on COCO costs every row $0.9$ to $2.0$~p.p.\@, and AAS wins back $0.1$ to $0.4$~p.p.\@ over the row it starts from. The Objects365 gaps are small and close together ($2.1$ to $2.7$ with COCO NPMI, $3.0$ to $3.5$ with O365 NPMI); as with Co-DINO, ranking the pairs by Objects365 NPMI widens the gap of every row, here by $0.6$ to $0.9$~p.p.\@

\noindent\textbf{Per-tier STRANGE-Bench breakdown.}
\label{app:gdino_strange_full}
Per tier (Tab.~\ref{tab:gdino_strange_full}), every fine-tuned row lowers the zero-shot gap at T3 to T5, and the mean gap over T1 to T5 drops from $2.4$ (Zero Shot) to between $0.3$ (XPaste and XPaste\,+\,AAS) and $0.9$ (SCP). XPaste\,+\,AAS \emph{(ours)} has the highest typical and atypical AP at every tier and the smallest gap at T4 and T5 ($1.2$, $-0.3$). At T1 both AAS rows have a larger gap than the row they start from ($0.5$ vs.\ $-0.3$ and $0.9$ vs.\ $-0.1$), because typical AP rises more than atypical AP there.

\begin{table*}[!htp]%
\centering
\scriptsize
\setlength{\tabcolsep}{1.8pt}%
\caption{\textbf{Grounding DINO Swin-B: STRANGE-Bench per-tier breakdown.} The zero-shot gap appears once a scene holds two partners ($3.6$ to $4.2$~p.p.\@ at T3 to T5, against $-0.1$ and $0.6$ at T1 and T2); after paste fine-tuning it stays below $3$~p.p.\@ at every tier, and XPaste\,+\,AAS \emph{(ours)} has the highest typical and atypical AP at every tier and the highest $T_{\text{never}}$. Full-annotation bbox mAP (\%) of the typical (T) and atypical (A) arm of each tier and their gap $\Delta$; $T_{1\text{-}5}$ is the mean over T1 to T5, rows as in Tab.~\ref{tab:grounding_dino_main}. Best per column in \textbf{bold}, second-best~\underline{underlined}.}
\label{tab:gdino_strange_full}
\resizebox{\linewidth}{!}{%
\begin{tabular}{l ccc ccc ccc ccc ccc ccc c}
\toprule
& \multicolumn{3}{c}{\textbf{Tier 1}} & \multicolumn{3}{c}{\textbf{Tier 2}} & \multicolumn{3}{c}{\textbf{Tier 3}} & \multicolumn{3}{c}{\textbf{Tier 4}} & \multicolumn{3}{c}{\textbf{Tier 5}} & \multicolumn{3}{c}{\textbf{Mean T1--T5}} & \\
\cmidrule(lr){2-4} \cmidrule(lr){5-7} \cmidrule(lr){8-10} \cmidrule(lr){11-13} \cmidrule(lr){14-16} \cmidrule(lr){17-19}
Aug & T$\uparrow$ & A$\uparrow$ & $\Delta\downarrow$ & T$\uparrow$ & A$\uparrow$ & $\Delta\downarrow$ & T$\uparrow$ & A$\uparrow$ & $\Delta\downarrow$ & T$\uparrow$ & A$\uparrow$ & $\Delta\downarrow$ & T$\uparrow$ & A$\uparrow$ & $\Delta\downarrow$ & T$\uparrow$ & A$\uparrow$ & $\Delta\downarrow$ & $T_{\text{never}}\uparrow$ \\
\midrule
Zero Shot & 94.9 & 95.0 & \underline{$-0.1$} & 95.0 & 94.4 & 0.6 & 93.3 & 89.7 & 3.6 & 92.4 & 88.2 & 4.2 & 90.9 & 87.2 & 3.7 & 93.3 & 90.9 & 2.4 & 92.4 \\
\midrule
\multicolumn{20}{l}{\emph{Non-generative}} \\
SCP & 94.7 & 95.0 & \textbf{\boldmath $-0.3$} & 95.0 & 95.3 & $-0.3$ & 92.6 & 92.2 & 0.4 & 92.8 & 89.9 & 2.9 & 91.5 & 89.8 & 1.7 & 93.3 & 92.4 & 0.9 & 93.3 \\
\midrule
\multicolumn{20}{l}{\emph{Generative}} \\
XPaste & 95.3 & 95.4 & \underline{$-0.1$} & 95.5 & \underline{96.3} & \textbf{\boldmath $-0.8$} & 93.4 & 93.0 & 0.4 & 93.5 & \underline{91.9} & \underline{1.6} & 92.0 & \underline{91.6} & \underline{0.4} & 93.9 & 93.6 & \textbf{0.3} & \underline{94.6} \\
\midrule
\rowcolor{oursrow} SCP + AAS \emph{(ours)} & \underline{96.2} & \underline{95.7} & 0.5 & \underline{96.0} & \underline{96.3} & $-0.3$ & \underline{93.5} & \underline{93.7} & \textbf{\boldmath $-0.2$} & \underline{93.6} & 91.4 & 2.2 & \underline{92.4} & 91.5 & 0.9 & \underline{94.3} & \underline{93.7} & \underline{0.6} & 94.5 \\
\rowcolor{oursrow} \textbf{XPaste + AAS \emph{(ours)}} & \textbf{97.0} & \textbf{96.1} & 0.9 & \textbf{96.4} & \textbf{97.0} & \underline{$-0.6$} & \textbf{94.2} & \textbf{93.9} & \underline{0.3} & \textbf{94.3} & \textbf{93.1} & \textbf{1.2} & \textbf{92.9} & \textbf{93.2} & \textbf{\boldmath $-0.3$} & \textbf{95.0} & \textbf{94.7} & \textbf{0.3} & \textbf{95.2} \\
\bottomrule
\end{tabular}

}
\end{table*}

\subsection{DINOv3 ViT-B/16 Fine-Tuned: Full Results}
\label{app:dinov3_full}

Like the two detectors above, the fine-tuned DINOv3 ViT-B/16 results of Tab.~\ref{tab:dinov3_main} come with a cross-domain transfer table (Tab.~\ref{tab:dinov3_ft_cd}) and a per-tier STRANGE-Bench breakdown (Tab.~\ref{tab:dinov3_ft_strange}). The splits are again the quintile ones, and every number is F1\textsubscript{micro} (the natural metric for the multi-label classification head; see \S\ref{app:exp-recipes}). The augmentation suite mirrors that of the detection families.

\noindent\textbf{In-domain typicality.}
On COCO (Tab.~\ref{tab:dinov3_ft_cd}, left block), Std.\ aug yields a typicality gap of $\Delta^{\text{T-A}} = {}$$5.5$ at $\text{F1}^{\text{val}} = {}$$81.9$. Color jitter and MixUp shave the gap modestly (4.6 and 4.2) without hurting overall F1. Mosaic and YOLOX produce the smallest non-paste gaps ($4.0$ and $3.7$) but \emph{collapse} overall F1 to $69.6$ and $70.0$, the same uniform-degradation trap visible in the detection families (\S\ref{app:codino_full}). XPaste\,+\,AAS \emph{(ours)} attains the highest in-domain F1 ($82.5$) and a gap of $\Delta^{\text{T-A}} = 3.9$ ($1.6$ below Std.; SCP reaches $3.7$).

\noindent\textbf{Cross-domain typicality.}
On Objects365, for the right block (pairs ranked by NPMI recomputed on Objects365 train, \S\ref{app:exp-datasets}, i.e.\ by the co-occurrence statistics of the domain the images come from), the gap is $7.1$ to $8.0$~p.p.\@ for every row except Mosaic and YOLOX, against AP gaps of $3.8$ to $5.1$ for Co-DINO (Tab.~\ref{tab:codino_cd}) and $3.0$ to $3.5$ for GDINO (Tab.~\ref{tab:gdino_o365}) on the same split, so the typicality bias of the fine-tuned classifier carries over to real images of another dataset. Mosaic and YOLOX have smaller gaps ($4.8$ and $4.6$) only because both arms collapse (typical F1 $26.8$ and $27.3$ against $38.1$ to $41.2$ for the other rows), as on COCO. SCP and SCP\,+\,AAS \emph{(ours)} have the highest typical and atypical F1 ($41.2 / 33.7$ and $41.0 / 33.6$); against Std.\ ($40.1 / 32.2$) both raise the atypical arm more than the typical one ($+1.5$ vs.\ $+1.1$ and $+1.4$ vs.\ $+0.9$), so the gap narrows from $7.9$ to $7.5$ and $7.4$, the smallest apart from Mosaic, YOLOX and No aug ($7.1$, at lower F1 on both arms). XPaste and XPaste\,+\,AAS stay at the Std.\ level on both arms ($40.0 / 32.2$ and $40.0 / 32.3$), so the in-domain lead of XPaste\,+\,AAS does not carry into this transfer. With COCO NPMI (middle block), every Objects365 gap is $1.7$ to $1.8$ times larger ($12.6$ to $14.2$~p.p.\@; Mosaic and YOLOX $8.0$ and $8.1$), whereas the Co-DINO and GDINO gaps are smaller there ($2.0$ to $3.0$ and $2.1$ to $2.7$); SCP and SCP\,+\,AAS again have the highest atypical F1 ($31.4$ and $31.1$).

\begin{table*}[!htp]%
\centering
\footnotesize
\setlength{\tabcolsep}{4pt}
\caption{\textbf{DINOv3 ViT-B/16 fine-tuned: in-domain (COCO) and
cross-domain (Objects365) typicality.} On COCO, XPaste\,+\,AAS \emph{(ours)} has the highest F1$^{\text{val}}$, typical and atypical F1. Ranked by Objects365's own NPMI (right block), Objects365 shows a gap of $7.1$ to $8.0$~p.p.\@ for every row except Mosaic and YOLOX, whose arms collapse, and SCP and SCP\,+\,AAS \emph{(ours)} have the highest typical and atypical F1, raising the atypical arm more than the typical one over Std. Ranked by COCO NPMI (middle block), every Objects365 gap is $1.7$ to $1.8$ times larger. F1$_{\text{micro}}$ (\%) on the quintile paired splits of COCO val and Objects365 val, the latter split by COCO \texttt{train2017} NPMI or by NPMI recomputed on Objects365 train (App.~\ref{app:exp-datasets}); rows as in Tab.~\ref{tab:dinov3_main} without the linear probe. Best per column in \textbf{bold}, second-best \underline{underlined}.}
\label{tab:dinov3_ft_cd}
\begin{tabular}{l cccc ccc ccc}
\toprule
& \multicolumn{4}{c}{\textbf{COCO}} & \multicolumn{3}{c}{\textbf{Objects365, COCO NPMI}} & \multicolumn{3}{c}{\textbf{Objects365, O365 NPMI}} \\
\cmidrule(lr){2-5} \cmidrule(lr){6-8} \cmidrule(lr){9-11}
Augmentation & $\text{F1}^{\text{val}}$ $\uparrow$ & $\text{F1}^{\text{T}}$ $\uparrow$ & $\text{F1}^{\text{A}}$ $\uparrow$ & $\Delta^{\text{T-A}}$ $\downarrow$ & $\text{F1}^{\text{T}}$ $\uparrow$ & $\text{F1}^{\text{A}}$ $\uparrow$ & $\Delta^{\text{T-A}}$ $\downarrow$ & $\text{F1}^{\text{T}}$ $\uparrow$ & $\text{F1}^{\text{A}}$ $\uparrow$ & $\Delta^{\text{T-A}}$ $\downarrow$ \\
\midrule
No aug & 78.9 & 77.8 & 70.1 & 7.7 & 42.0 & 29.3 & 12.6 & 38.1 & 30.9 & 7.1 \\
\midrule
\multicolumn{11}{l}{\emph{Non-generative}} \\
Std. & 81.9 & 79.8 & 74.3 & 5.5 & 43.9 & 30.1 & 13.8 & 40.1 & 32.2 & 7.9 \\
Color jitter & 81.8 & 79.7 & 75.1 & 4.6 & 43.8 & 30.1 & 13.7 & 40.2 & 32.2 & 8.0 \\
MixUp & \underline{82.2} & 79.6 & 75.4 & 4.2 & 43.9 & 30.2 & 13.7 & 40.1 & 32.6 & 7.5 \\
Mosaic & 69.6 & 71.0 & 67.0 & 4.0 & 28.6 & 20.5 & \textbf{8.0} & 26.8 & 22.0 & \underline{4.8} \\
YOLOX & 70.0 & 71.2 & 67.4 & \textbf{3.7} & 28.9 & 20.8 & \underline{8.1} & 27.3 & 22.7 & \textbf{4.6} \\
SCP & 81.6 & 78.8 & 75.1 & \textbf{3.7} & \textbf{44.2} & \textbf{31.4} & 12.8 & \textbf{41.2} & \textbf{33.7} & 7.5 \\
\midrule
\multicolumn{11}{l}{\emph{Generative}} \\
InstAug & 81.9 & 79.5 & 74.4 & 5.2 & \underline{44.0} & 30.1 & 13.9 & 40.3 & 32.5 & 7.9 \\
XPaste & 82.1 & \underline{80.2} & \underline{76.1} & 4.1 & 43.8 & 29.7 & 14.2 & 40.0 & 32.2 & 7.8 \\
\midrule
\rowcolor{oursrow} SCP + AAS \emph{(ours)} & 81.7 & 79.0 & 75.0 & 4.0 & \underline{44.0} & \underline{31.1} & 12.8 & \underline{41.0} & \underline{33.6} & 7.4 \\
\rowcolor{oursrow} \textbf{XPaste + AAS \emph{(ours)}} & \textbf{82.5} & \textbf{80.6} & \textbf{76.8} & \underline{3.9} & 43.6 & 29.7 & 13.9 & 40.0 & 32.3 & 7.7 \\
\bottomrule
\end{tabular}

\end{table*}

\noindent\textbf{Per-tier STRANGE-Bench breakdown.} Tab.~\ref{tab:dinov3_ft_strange} reports the full per-tier STRANGE-Bench breakdown for DINOv3 ViT-B/16 FT. Three observations stand out. First, this family has the largest single-tier gap in the paper: the Std.\ baseline gap grows from $\Delta = -0.7$ at $T_1$ to $\Delta = {}$$40.6$ at $T_5$, i.e.\ a $41$ p.p.\@ swing as contextual load increases, with $T_{\text{never}}$ at only $62.6$. The mean gap $\Delta_{1\text{-}5} = {}$$22.2$ is roughly $3.5\times$ that of Co-DINO Std.\ ($6.3$) and about $9\times$ that of GDINO zero-shot ($2.4$), reflecting the strong typicality prior baked into a multi-label head trained on natural co-occurrences. Second, the paste-based interventions are correspondingly more dramatic: SCP alone reduces the mean gap to $4.3$ and lifts $T_{\text{never}}$ to $86.5$ (a $+24$ p.p.\@ absolute improvement over Std.). XPaste\,+\,AAS \emph{(ours)} pushes both numbers further to $\Delta_{1\text{-}5} = {}$$-0.8$ and $T_{\text{never}} = {}$$96.1$ (XPaste alone: $-0.7$ and $95.0$), the largest mitigation effect we observe in this paper. Third, the gap-closure profile is itself revealing: under XPaste\,+\,AAS the per-tier gap stays at or below $0.7$~p.p.\@ for every tier $T_1$ to $T_5$ (negative at $T_1$ to $T_3$, where atypical F1 is higher), including $T_5$ (gap $= {}$$0.7$, vs $40.6$ for Std.\ and $33.8$ for No-aug). This means the model's reliance on typical co-occurrence in the highest-load regime is essentially eliminated, while typical-tier F1 stays well above Std. ($\text{T}5^{\text{T}} = {}$$93.3$ vs Std.'s $88.6$). Mosaic and YOLOX, despite small mean gaps ($6.7$ and $6.0$), again pay for it with collapsed typical scores ($T^{\text{T}}_{1\text{-}5} = 74.7$ and $75.1$ vs.\ $89.7$ for Std.).

\begin{table*}[!htp]%
\centering
\scriptsize
\setlength{\tabcolsep}{2pt}
\caption{\textbf{DINOv3 ViT-B/16 fine-tuned: STRANGE-Bench per-tier breakdown.} The Std.\ gap grows from $-0.7$ at T1 to $40.6$ at T5 (mean $22.2$), the largest single-tier gap in the paper; SCP brings the mean down to $4.3$ and XPaste\,+\,AAS \emph{(ours)} to $-0.8$, with gaps between $-2.4$ and $0.7$ at every tier and the highest $T_{\text{never}}$ ($96.1$). F1$_{\text{micro}}$ (\%) of the typical (T) and atypical (A) arm of each tier and their gap $\Delta$; $T_{1\text{-}5}$ is the mean over T1 to T5, rows as in Tab.~\ref{tab:dinov3_main} without the linear probe. Best per column in \textbf{bold}, second-best \underline{underlined}.}
\label{tab:dinov3_ft_strange}
\resizebox{\linewidth}{!}{%
\begin{tabular}{l ccc ccc ccc ccc ccc ccc c}
\toprule
& \multicolumn{3}{c}{\textbf{Tier 1}} & \multicolumn{3}{c}{\textbf{Tier 2}} & \multicolumn{3}{c}{\textbf{Tier 3}} & \multicolumn{3}{c}{\textbf{Tier 4}} & \multicolumn{3}{c}{\textbf{Tier 5}} & \multicolumn{3}{c}{\textbf{Mean T1--T5}} & \\
\cmidrule(lr){2-4} \cmidrule(lr){5-7} \cmidrule(lr){8-10} \cmidrule(lr){11-13} \cmidrule(lr){14-16} \cmidrule(lr){17-19}
Aug & T$\uparrow$ & A$\uparrow$ & $\Delta\downarrow$ & T$\uparrow$ & A$\uparrow$ & $\Delta\downarrow$ & T$\uparrow$ & A$\uparrow$ & $\Delta\downarrow$ & T$\uparrow$ & A$\uparrow$ & $\Delta\downarrow$ & T$\uparrow$ & A$\uparrow$ & $\Delta\downarrow$ & T$\uparrow$ & A$\uparrow$ & $\Delta\downarrow$ & $T_{\text{never}}\uparrow$ \\
\midrule
No aug & 87.5 & 89.2 & $-1.7$ & 87.2 & 75.6 & 11.6 & 85.5 & 63.2 & 22.3 & 84.4 & 54.4 & 30.1 & 82.8 & 49.1 & 33.8 & 85.5 & 66.3 & 19.2 & 61.5 \\
\midrule
\multicolumn{20}{l}{\emph{Non-generative}} \\
Std. & 91.3 & 92.0 & $-0.7$ & 91.0 & 79.2 & 11.7 & 89.2 & 64.2 & 25.0 & 88.3 & 54.0 & 34.3 & 88.6 & 48.0 & 40.6 & 89.7 & 67.5 & 22.2 & 62.6 \\
Color jitter & 91.6 & 91.5 & 0.1 & 90.6 & 79.6 & 11.0 & 89.2 & 64.3 & 24.8 & 88.3 & 52.4 & 35.9 & 86.5 & 46.0 & 40.6 & 89.2 & 66.7 & 22.5 & 62.7 \\
MixUp & 91.4 & 93.5 & $-2.2$ & 91.0 & 85.6 & 5.4 & 90.2 & 70.8 & 19.4 & 89.1 & 60.5 & 28.5 & 88.0 & 51.6 & 36.4 & 89.9 & 72.4 & 17.5 & 69.9 \\
Mosaic & 57.8 & 51.5 & 6.3 & 71.7 & 70.2 & 1.5 & 76.9 & 73.3 & 3.6 & 83.1 & 73.0 & 10.1 & 84.2 & 72.1 & 12.1 & 74.7 & 68.0 & 6.7 & 70.6 \\
YOLOX & 59.2 & 55.0 & 4.2 & 71.9 & 71.7 & 0.1 & 78.2 & 73.4 & 4.9 & 82.4 & 73.3 & 9.1 & 84.0 & 72.1 & 11.9 & 75.1 & 69.1 & 6.0 & 70.7 \\
SCP & 92.3 & 94.0 & $-1.7$ & 92.4 & 92.5 & $-0.1$ & 91.5 & 89.0 & 2.5 & 91.6 & 83.4 & 8.1 & 89.4 & 76.7 & 12.7 & 91.4 & 87.1 & 4.3 & 86.5 \\
\midrule
\multicolumn{20}{l}{\emph{Generative}} \\
InstAug & 91.4 & 93.6 & $-2.2$ & 91.3 & 79.1 & 12.2 & 89.1 & 64.2 & 24.9 & 88.6 & 53.1 & 35.5 & 87.2 & 48.4 & 38.7 & 89.5 & 67.7 & 21.8 & 63.3 \\
XPaste & 93.5 & \underline{96.7} & \textbf{\boldmath $-3.1$} & \underline{93.8} & \underline{96.4} & \textbf{\boldmath $-2.6$} & \underline{93.9} & \underline{94.0} & \underline{$-0.1$} & \underline{94.1} & \underline{93.6} & \underline{0.5} & \textbf{93.4} & \underline{91.4} & \underline{1.9} & \underline{93.8} & \underline{94.4} & \underline{$-0.7$} & \underline{95.0} \\
\midrule
\rowcolor{oursrow} SCP + AAS \emph{(ours)} & \underline{94.2} & 96.3 & $-2.1$ & 92.9 & 94.7 & \underline{$-1.8$} & 92.2 & 91.6 & 0.6 & 92.3 & 88.8 & 3.5 & 90.8 & 84.8 & 6.0 & 92.5 & 91.2 & 1.2 & 89.5 \\
\rowcolor{oursrow} \textbf{XPaste + AAS \emph{(ours)}} & \textbf{94.9} & \textbf{97.3} & \underline{$-2.4$} & \textbf{95.4} & \textbf{97.2} & \underline{$-1.8$} & \textbf{94.7} & \textbf{95.5} & \textbf{\boldmath $-0.9$} & \textbf{94.9} & \textbf{94.6} & \textbf{0.3} & \underline{93.3} & \textbf{92.6} & \textbf{0.7} & \textbf{94.6} & \textbf{95.5} & \textbf{\boldmath $-0.8$} & \textbf{96.1} \\
\bottomrule
\end{tabular}
}
\end{table*}

\subsection{DINOv3 ViT-B/16 Linear Probe: Full Results}
\label{app:dinov3_lp}

We additionally evaluate typicality bias under the linear-probe (LP) protocol on DINOv3 ViT-B/16, where only a linear classification head is trained on top of the frozen pretrained backbone. This isolates the contribution of the backbone's representations from end-to-end fine-tuning: any gap that survives the LP setup is encoded in the backbone itself and must be read out of features the backbone already provides, while any gap reduction obtained by augmentation must operate purely through the head's training signal. We restrict the LP study to Std.\ and three paste augmentations of the FT sweep (\S\ref{app:dinov3_full}), SCP, XPaste and XPaste\,+\,AAS, since the goal of this section is to characterize the \emph{ceiling} on what augmentation can achieve in the frozen-backbone regime, not to repeat a full sweep. Unlike in the fine-tuned tables, the AAS row here (XPaste\,+\,AAS$^{\dagger}$) applies AAS sampling over the whole LP schedule rather than as the late stage of Sec.~\ref{sec:aas}, and every row is a single training draw. 

Tab.~\ref{tab:dinov3_lp_main} reports the in-domain results together with mean STRANGE-Bench statistics;  Tab.~\ref{tab:dinov3_lp_cd} reports the cross-domain transfer to Objects365; and Tab.~\ref{tab:dinov3_lp_strange} reports the full per-tier STRANGE-Bench breakdown. Three observations stand out. \emph{First}, the standard augmentation recipe (\emph{Std.}) attains the highest in-domain F1 on COCO val and on the typical split, but also exhibits the largest typicality gap ($\Delta^{\text{T-A}} = 8.6$); paste-based methods trade a small amount of in-domain F1 ($1.5$ to $2.3$~p.p.\@) for substantially smaller gaps ($5.7$ to $5.9$). \emph{Second}, on STRANGE-Bench, \emph{XPaste\,+\,AAS$^{\dagger}$} is the strongest mitigator under the LP protocol: it achieves the lowest mean tier gap ($\Delta_{1\text{-}5} = 6.0$), the highest atypical mean ($68.1$) and $T_{\text{never}}$ score ($70.1$), and the smallest gap at T2 to T5; XPaste is second on these three metrics ($8.1$, $67.3$, $67.9$) and SCP third ($9.2$, $57.5$, $55.7$). \emph{Third}, the STRANGE-Bench ordering is the same as under end-to-end fine-tuning (XPaste\,+\,AAS, then XPaste, then SCP; Tab.~\ref{tab:dinov3_main} and \S\ref{app:dinov3_full}): the AAS tilt lowers $\Delta_{1\text{-}5}$ from $8.1$ to $6.0$ and raises $T_{\text{never}}$ from $67.9$ to $70.1$ over XPaste even when only the head is trained, but the residual gap stays larger than after fine-tuning ($-0.8$).

\begin{table*}[!htp]%
\centering
\footnotesize
\setlength{\tabcolsep}{4pt}
\caption{\textbf{DINOv3 ViT-B/16 linear probe: typicality bias under
augmentation.} With the backbone frozen, XPaste\,+\,AAS$^{\dagger}$ \emph{(ours)} has the smallest STRANGE-Bench gap ($6.0$ against $23.9$ for Std.), the highest atypical mean and the highest $T_{\text{never}}$ ($70.1$), while Std.\ keeps the highest COCO F1$^{\text{val}}$ and SCP has the smallest COCO gap ($5.7$). F1$_{\text{micro}}$ (\%) on COCO val, the COCO quintile paired split and STRANGE-Bench; one training draw per row. $^{\dagger}$AAS sampling ($\beta = 3.5$) over the whole linear-probe schedule, without the late stage on generated scenes and the weight averaging of Sec.~\ref{sec:aas}. Best per column in \textbf{bold}, second-best \underline{underlined}.}
\label{tab:dinov3_lp_main}
\begin{tabular}{l cccc cccc}
\toprule
& \multicolumn{4}{c}{\textbf{COCO}} & \multicolumn{4}{c}{\textbf{STRANGE-Bench}} \\
\cmidrule(lr){2-5} \cmidrule(lr){6-9}
Augmentation & $\text{F1}^{\text{val}}$ $\uparrow$ & $\text{F1}^{\text{T}}$ $\uparrow$ & $\text{F1}^{\text{A}}$ $\uparrow$ & $\Delta^{\text{T-A}}$ $\downarrow$ & $T_{1\text{-}5}^{\text{T}}$ $\uparrow$ & $T_{1\text{-}5}^{\text{A}}$ $\uparrow$ & $\Delta_{1\text{-}5}$ $\downarrow$ & $T_{\text{never}}$ $\uparrow$ \\
\midrule
Std. & \textbf{74.6} & \textbf{71.8} & 63.2 & 8.6 & 63.5 & 39.6 & 23.9 & 38.9 \\
SCP & 72.3 & 67.9 & 62.2 & \textbf{5.7} & 66.7 & 57.5 & 9.2 & 55.7 \\
\midrule
XPaste & \underline{73.1} & \underline{70.3} & \textbf{64.4} & \underline{5.9} & \textbf{75.4} & \underline{67.3} & \underline{8.1} & \underline{67.9} \\
\midrule
\rowcolor{oursrow} \textbf{XPaste + AAS$^{\dagger}$ \emph{(ours)}} & 73.0 & 70.1 & \underline{64.2} & \underline{5.9} & \underline{74.1} & \textbf{68.1} & \textbf{6.0} & \textbf{70.1} \\
\bottomrule
\end{tabular}

\end{table*}

\noindent\textbf{Why augmentation hits a ceiling under LP.}
The residual gap that no LP augmentation closes ($\Delta_{1\text{-}5} \geq 6.0$, $T_{\text{never}} \leq 70.1$) points to the frozen features: DINOv3 is pretrained without labels on about $1.7$ billion images (LVD-1689M~\cite{simeoni2025dinov3}), and the gap of the Std.\ probe grows with the pretraining data (DINOv1 $19.3$, DINOv2 $21.5$, DINOv3 $23.9$; App.~\ref{app:scale_ladder}), which is consistent with co-occurrence structure that is already present in the features. Once the backbone is frozen, the head-only training signal cannot edit those internal representations; it can only re-route how they are linearly combined into class logits. Paste-based methods re-balance the head's input distribution and lower the gap from $23.9$ to between $6.0$ and $9.2$; the remainder may reflect co-occurrence structure that a linear head cannot separate from the pretrained representation. End-to-end FT (\S\ref{app:dinov3_full}) lets the gradient reach the backbone, so the pasted atypical samples can also change the features; this is consistent with FT's $\Delta_{1\text{-}5} = -0.8$ for XPaste\,+\,AAS versus LP's $6.0$ for XPaste\,+\,AAS$^{\dagger}$.

\noindent\textbf{In-domain typicality.}
Tab.~\ref{tab:dinov3_lp_main} consolidates the LP results on COCO val, the COCO paired typicality splits, and the STRANGE-Bench means over T1 to T5 plus $T_{\text{never}}$. Two effects are visible. On the COCO columns, Std.\ achieves the highest absolute F1 ($\text{F1}^{\text{val}} = 74.6$, $\text{F1}^{\text{T}} = 71.8$; XPaste has the highest $\text{F1}^{\text{A}}$, $64.4$) but with an $8.6$~p.p.\@ typicality gap; all three paste methods (SCP, XPaste, XPaste\,+\,AAS) give up $1.5$ to $2.3$~p.p.\@ of $\text{F1}^{\text{val}}$ in exchange for gaps of $5.7$ to $5.9$. On the STRANGE-Bench columns the contrast is much sharper: Std.\ has $\Delta_{1\text{-}5} = 23.9$, more than $2\times$ the in-domain gap, and the lowest $T_{\text{never}}$ ($38.9$). All three paste methods cut the gap to between a quarter and two fifths ($6.0$ to $9.2$) and raise $T_{\text{never}}$ to $55.7$ to $70.1$. XPaste\,+\,AAS$^{\dagger}$ \emph{(ours)} leads on $\Delta_{1\text{-}5}$ ($6.0$) and on $T_{\text{never}}$ ($70.1$), ahead of XPaste ($8.1$, $67.9$) and SCP ($9.2$, $55.7$). Note that the LP setting keeps the FT ordering (\S\ref{app:dinov3_full}, $\Delta_{1\text{-}5} = -0.8$ for XPaste\,+\,AAS) but not its size: under the frozen backbone, the best mean tier gap is $6.0$.

\begin{table*}[!htp]%
\centering
\footnotesize
\setlength{\tabcolsep}{4pt}
\caption{\textbf{DINOv3 ViT-B/16 linear probe: cross-domain
(Objects365) typicality.} XPaste\,+\,AAS$^{\dagger}$ \emph{(ours)} has the smallest gap ($15.0$ against $18.0$ for Std.) and SCP the highest typical and atypical F1. F1$_{\text{micro}}$ (\%) on the Objects365 val quintile paired split ranked by COCO \texttt{train2017} NPMI (the Objects365-NPMI split of Tab.~\ref{tab:dinov3_ft_cd} was not evaluated for the probe); COCO columns as in Tab.~\ref{tab:dinov3_lp_main}. Best per column in \textbf{bold}, second-best~\underline{underlined}.}
\label{tab:dinov3_lp_cd}
\begin{tabular}{l ccc}
\toprule
& \multicolumn{3}{c}{\textbf{Objects365 (cross-domain)}} \\
\cmidrule(lr){2-4}
Augmentation & $\text{F1}^{\text{T}}$ $\uparrow$ & $\text{F1}^{\text{A}}$ $\uparrow$ & $\Delta^{\text{T-A}}$ $\downarrow$ \\
\midrule
Std. & \underline{49.6} & 31.7 & 18.0 \\
SCP & \textbf{50.1} & \textbf{33.6} & 16.6 \\
\midrule
XPaste & 47.0 & 31.7 & \underline{15.3} \\
\midrule
\rowcolor{oursrow} \textbf{XPaste + AAS$^{\dagger}$ \emph{(ours)}} & 47.2 & \underline{32.2} & \textbf{15.0} \\
\bottomrule
\end{tabular}

\end{table*}

\noindent\textbf{Cross-domain typicality.}
Tab.~\ref{tab:dinov3_lp_cd} reports the cross-domain transfer to
Objects365 (paired O365 splits, mapped onto the COCO-80 vocabulary).
As in DINOv3 FT, absolute F1 is much lower than in-domain because
of the domain shift, and Std.\ again has the largest gap ($18.0$); SCP has the highest typical and atypical scores ($50.1 / 33.6$), and XPaste\,+\,AAS$^{\dagger}$ \emph{(ours)} the smallest gap ($15.0$; XPaste $15.3$, SCP $16.6$). The LP gaps are larger than the FT gaps on the same split ($12.6$ to $14.2$~p.p.\@, \S\ref{app:dinov3_full}), and paste lowers them more: by $1.4$ to $3.0$~p.p.\@ against Std.\ under LP and by at most $1.0$ under FT (XPaste: $-2.7$ under LP, $+0.4$ under FT).

\begin{table*}[!htp]%
\centering
\scriptsize
\setlength{\tabcolsep}{2pt}
\caption{\textbf{DINOv3 ViT-B/16 linear probe: per-tier
STRANGE-Bench breakdown.} The Std.\ gap grows at every tier, from $10.8$ at T1 to $37.2$ at T5; every paste row keeps it at $12.2$ or below, and XPaste\,+\,AAS$^{\dagger}$ \emph{(ours)} has the smallest gap at T2 to T5 ($3.4$ to $7.4$). F1$_{\text{micro}}$ (\%) of the typical (T) and atypical (A) arm of each tier and their gap $\Delta$; Mean T1--T5 is the mean over the five tiers ($T_{1\text{-}5}$ in Tab.~\ref{tab:dinov3_lp_main}); rows as in Tab.~\ref{tab:dinov3_lp_main}. Best per column in \textbf{bold}, second-best \underline{underlined}.}
\label{tab:dinov3_lp_strange}
\begin{tabular}{l ccc ccc ccc ccc ccc ccc c}
\toprule
& \multicolumn{3}{c}{\textbf{Tier 1}} & \multicolumn{3}{c}{\textbf{Tier 2}} & \multicolumn{3}{c}{\textbf{Tier 3}} & \multicolumn{3}{c}{\textbf{Tier 4}} & \multicolumn{3}{c}{\textbf{Tier 5}} & \multicolumn{3}{c}{\textbf{Mean T1--T5}} & \\
\cmidrule(lr){2-4} \cmidrule(lr){5-7} \cmidrule(lr){8-10} \cmidrule(lr){11-13} \cmidrule(lr){14-16} \cmidrule(lr){17-19}
Aug & T$\uparrow$ & A$\uparrow$ & $\Delta\downarrow$ & T$\uparrow$ & A$\uparrow$ & $\Delta\downarrow$ & T$\uparrow$ & A$\uparrow$ & $\Delta\downarrow$ & T$\uparrow$ & A$\uparrow$ & $\Delta\downarrow$ & T$\uparrow$ & A$\uparrow$ & $\Delta\downarrow$ & T$\uparrow$ & A$\uparrow$ & $\Delta\downarrow$ & $T_{\text{never}}\uparrow$ \\
\midrule
Std. & 68.3 & 57.5 & 10.8 & 73.2 & 56.2 & 17.1 & 62.3 & 39.9 & 22.4 & 58.4 & 26.6 & 31.9 & 55.2 & 18.0 & 37.2 & 63.5 & 39.6 & 23.9 & 38.9 \\
SCP & 76.2 & 66.1 & 10.0 & 75.2 & 67.8 & 7.4 & 65.2 & 58.9 & \underline{6.3} & 59.6 & 49.5 & 10.1 & 57.3 & 45.1 & 12.2 & 66.7 & 57.5 & 9.2 & 55.7 \\
\midrule
XPaste & \underline{81.3} & \textbf{74.8} & \textbf{6.5} & \textbf{82.0} & \underline{76.6} & \underline{5.4} & \textbf{75.1} & \underline{67.4} & 7.8 & \textbf{70.5} & \underline{60.8} & \underline{9.7} & \textbf{68.1} & \underline{57.0} & \underline{11.1} & \textbf{75.4} & \underline{67.3} & \underline{8.1} & \underline{67.9} \\
\midrule
\rowcolor{oursrow} \textbf{XPaste + AAS$^{\dagger}$ \emph{(ours)}} & \textbf{81.8} & \underline{74.4} & \underline{7.5} & \underline{81.0} & \textbf{77.7} & \textbf{3.4} & \underline{73.2} & \textbf{68.5} & \textbf{4.7} & \underline{69.2} & \textbf{61.8} & \textbf{7.4} & \underline{65.3} & \textbf{58.2} & \textbf{7.1} & \underline{74.1} & \textbf{68.1} & \textbf{6.0} & \textbf{70.1} \\
\bottomrule
\end{tabular}

\end{table*}

\noindent\textbf{Per-tier STRANGE-Bench breakdown.}
Tab.~\ref{tab:dinov3_lp_strange} reports the full per-tier
STRANGE-Bench breakdown for DINOv3 LP. Two patterns are visible.
First, the Std.\ baseline gap grows at every tier, from $\Delta = 10.8$ at T1 to $\Delta = 37.2$ at T5, which confirms that contextual load amplifies the bias
under LP just as it does under FT. Second, all paste-based
augmentations cut the gap at every tier, to between $3.4$ and $12.2$ (T5: $7.1$ to $12.2$ vs.\ $37.2$ for Std.); XPaste\,+\,AAS$^{\dagger}$ has the smallest gap at T2 to T5 and XPaste at T1.
The $T_{\text{never}}$ column tells the same story: XPaste\,+\,AAS$^{\dagger}$ at $70.1$ above XPaste at $67.9$ and SCP at $55.7$, all well above Std.'s $38.9$.

\subsection{Selecting the AAS Atypicality Temperature \texorpdfstring{$\beta$}{β}}
\label{app:beta_sweep}

We select the AAS atypicality temperature $\beta$ once via a small ablation and hold the selected value fixed across all model families and full-COCO runs. Tab.~\ref{tab:beta_sweep} reports the sweep on Co-DINO \mbox{R-50} (AP), trained on the $25\%$ subset of COCO \texttt{train2017} (mini-COCO, $29{,}571$ images), and on DINOv3 ViT-B/16 fine-tuned on full COCO (mAP). 
$\beta = 3.5$ achieves the best or second-best score on every metric for both models, simultaneously balancing overall accuracy, in-domain typicality gap, and STRANGE-Bench coverage, and we adopt it for all main-paper experiments. 

\noindent\textbf{Tradeoff intuition.}
The sampling distribution $P(j \mid a) \propto \pi(j) \exp\big(\beta A(a, j)\big)$ of Sec.~\ref{sec:aas} has two endpoints. At $\beta = 0$, AAS samples paste classes uniformly and independently of the anchor, i.e.\ the XPaste baseline, and provides no debiasing signal; as $\beta \to \infty$, AAS concentrates almost all paste mass on the most atypical partners, which over-distorts the training-time joint and starts to hurt typical-pair coverage and overall accuracy. The sweep traces this tradeoff: $\beta = 2$ leaves residual gap ($\Delta^{\text{T-A}} = 8.1$ for Co-DINO, $9.5$ for DINOv3 FT, both worse than $\beta = 3.5$); $\beta = 5$ matches or beats $\beta = 3.5$ on a few metrics (Co-DINO typical AP, DINOv3 $\Delta_{1\text{-}5}$ on STRANGE-Bench) but drops on others (Co-DINO val AP, DINOv3 val mAP and gap), reflecting the over-tilting regime. $\beta = 3.5$ sits near the sweet spot for both families.

\begin{table*}[!htp]%
\centering
\footnotesize
\setlength{\tabcolsep}{4pt}
\caption{\textbf{AAS atypicality temperature $\beta$ sweep.} Co-DINO R-50 trained on mini-COCO, reported in AP; DINOv3 ViT-B/16 fine-tuned on full COCO, reported in mAP. $\beta = 3.5$ \emph{(ours,
blue)} attains the best or second-best score on every metric for
both models, and is the only setting that ranks highly on overall
accuracy, in-domain typicality gap, and STRANGE-Bench coverage
simultaneously.}
\label{tab:beta_sweep}
\begin{tabular}{l c cccc cc}
\toprule
& & \multicolumn{4}{c}{\textbf{COCO (quintile)}} & \multicolumn{2}{c}{\textbf{STRANGE-Bench}} \\
\cmidrule(lr){3-6} \cmidrule(lr){7-8}
Model & $\beta$ & val $\uparrow$ & T $\uparrow$ & A $\uparrow$ & $\Delta^{\text{T-A}}$ $\downarrow$ & $T_{\text{never}}$ $\uparrow$ & $\Delta_{1\text{-}5}$ $\downarrow$ \\
\midrule
\multicolumn{8}{l}{\emph{Co-DINO ResNet-50 (AP)}} \\
Co-DINO R-50         & $2$   & \textbf{43.8}    & \underline{44.3} & \underline{36.2} & \underline{8.1} & \underline{82.5} & \underline{1.6} \\
\rowcolor{oursrow} \textbf{Co-DINO R-50} & $\boldsymbol{3.5}$ & \textbf{43.8}    & \underline{44.3} & \textbf{36.4}    & \textbf{7.9}    & \textbf{83.1}    & \textbf{1.2} \\
Co-DINO R-50         & $5$   & \underline{43.7} & \textbf{44.8}    & 35.9             & 8.9             & 82.0             & 1.9 \\
\midrule
\multicolumn{8}{l}{\emph{DINOv3 ViT-B/16 fine-tuned (mAP)}} \\
DINOv3 ViT-B FT      & $2$   & \underline{85.8} & 84.2             & \underline{74.7} & \underline{9.5} & \underline{95.5} & \underline{0.6} \\
\rowcolor{oursrow} \textbf{DINOv3 ViT-B FT} & $\boldsymbol{3.5}$ & \textbf{86.4}    & \textbf{85.0}    & \textbf{75.8}    & \textbf{9.2}    & \textbf{96.1}    & \underline{0.5} \\
DINOv3 ViT-B FT      & $5$   & 85.7             & \underline{84.5} & 74.3             & 10.2            & 95.3             & \textbf{0.3} \\
\bottomrule
\end{tabular}
\end{table*}

\subsection{Circularity Checks}
\label{app:circularity}

\noindent COCO \texttt{train2017} NPMI defines the paired splits, the STRANGE-Bench tiers and the AAS tilt, and the AAS late stage trains on scenes from the generator of the benchmark. We test each of these links against a source the benchmark was not built from.

\noindent\textbf{Tiers under a second co-occurrence source.} Tab.~\ref{tab:app_circ_tiers} scores every anchor-partner pair of STRANGE-Bench with the NPMI of Objects365 \texttt{train} ($1.70$M images mapped to the COCO classes), which played no part in building the benchmark. Objects365 ranks the typical above the atypical pairs at every level (AUC $0.97$ to $0.99$, against $0.99$ to $1.00$ for COCO). If classes co-occurred independently, $85\%$ of the pairs that never co-occur in COCO \texttt{train2017} would be expected there at least $10$ times, where observing none has probability below $5{\times}10^{-5}$, so their absence is not a sampling accident. The pairs of $T_{\text{never}}$ are rare in Objects365 as well: $36.4\%$ of them never co-occur there, and their median Objects365 NPMI is $-0.22$.

\begin{table*}[!htp]%
\centering
\footnotesize
\setlength{\tabcolsep}{4pt}
\renewcommand{\arraystretch}{0.9}
\caption{\textbf{STRANGE-Bench tiers under Objects365 NPMI.} Objects365, which played no part in building the benchmark, ranks the typical above the atypical pairs at every level (AUC $0.97$ to $0.99$), and $36.4\%$ of the $T_{\text{never}}$ pairs never co-occur there either. Anchor-partner pairs of every scene scored with COCO \texttt{train2017} NPMI, which defines the tiers, and with Objects365 \texttt{train} NPMI; a pair that never co-occurs in a source is scored $-1$. NPMI: median over the typical / atypical arm of a level; $N_{ij}{=}0$: share of atypical pairs that never co-occur in the source; AUC: probability that a typical pair outranks an atypical pair of the same level.}
\label{tab:app_circ_tiers}
\begin{tabular}{l ccc ccc}
\toprule
& \multicolumn{3}{c}{COCO \texttt{train2017} (defines the tiers)} & \multicolumn{3}{c}{Objects365 \texttt{train}} \\
\cmidrule(lr){2-4} \cmidrule(lr){5-7}
Level & NPMI typ.\ / atyp. & $N_{ij}{=}0$ atyp.\ (\%) & AUC & NPMI typ.\ / atyp. & $N_{ij}{=}0$ atyp.\ (\%) & AUC \\
\midrule
$k=1$ & $0.20$ / $-1.00$ & $57.5$ & $1.00$ & $0.18$ / $-0.24$ & $32.2$ & $0.99$ \\
$k=2$ & $0.18$ / $-1.00$ & $53.1$ & $1.00$ & $0.15$ / $-0.21$ & $29.6$ & $0.99$ \\
$k=3$ & $0.18$ / $-0.38$ & $45.9$ & $1.00$ & $0.14$ / $-0.21$ & $28.2$ & $0.98$ \\
$k=4$ & $0.17$ / $-0.36$ & $43.1$ & $0.99$ & $0.14$ / $-0.20$ & $27.2$ & $0.97$ \\
\midrule
$T_{\text{never}}$ & -- / $-1.00$ & $82.4$ & -- & -- / $-0.22$ & $36.4$ & -- \\
\bottomrule
\end{tabular}

\end{table*}

\noindent\textbf{Paired splits defined by Objects365 NPMI.} Re-slicing COCO \texttt{val} with Objects365 NPMI (same construction and image pool, different images) leaves a typicality gap in every row we re-sliced, the baseline, paste and AAS rows of Tabs.~\ref{tab:codino_main} to~\ref{tab:dinov3_main} (Tab.~\ref{tab:app_circ_reslice}). The gap is smaller than under COCO NPMI, by $0.5$ to $3.9$~p.p.\@, so part of the gap of the COCO splits reflects the signal that defines them, but in no row does it disappear.

\begin{table*}[!htp]%
\centering
\footnotesize
\setlength{\tabcolsep}{4pt}
\renewcommand{\arraystretch}{0.9}
\caption{\textbf{COCO \texttt{val} paired splits defined by COCO or by Objects365 NPMI.} Every row keeps a gap on the Objects365-NPMI split, $0.5$ to $3.9$~p.p.\@ smaller than on the COCO-NPMI split. Baseline, paste and AAS rows of Tabs.~\ref{tab:codino_main} to~\ref{tab:dinov3_main} (same checkpoints). Both splits use the same construction and image pool (App.~\ref{app:splits-impl}) but select different images. Detectors: full-annotation AP; DINOv3: F1$_{\text{micro}}$.}
\label{tab:app_circ_reslice}
\begin{tabular}{l ccc ccc}
\toprule
& \multicolumn{3}{c}{Split by COCO NPMI} & \multicolumn{3}{c}{Split by Objects365 NPMI} \\
\cmidrule(lr){2-4} \cmidrule(lr){5-7}
Augmentation & Typ.\,$\uparrow$ & Atyp.\,$\uparrow$ & $\Delta^{\text{T-A}}\downarrow$ & Typ.\,$\uparrow$ & Atyp.\,$\uparrow$ & $\Delta^{\text{T-A}}\downarrow$ \\
\midrule
\multicolumn{7}{l}{\emph{Co-DINO R-50 (AP)}} \\
No aug & 44.2 & 35.5 & \underline{8.7} & 44.3 & 36.2 & 8.2 \\
Std. & \underline{53.2} & 44.0 & 9.2 & \textbf{52.3} & 44.2 & 8.1 \\
SCP & \underline{53.2} & \underline{44.2} & 9.0 & 51.5 & \underline{44.4} & \underline{7.1} \\
XPaste & 52.0 & 43.0 & 9.0 & 50.7 & 43.1 & 7.6 \\
\rowcolor{oursrow} SCP + AAS \emph{(ours)} & \textbf{54.0} & \textbf{45.5} & \textbf{8.5} & \underline{52.2} & \textbf{45.9} & \textbf{6.3} \\
\rowcolor{oursrow} \textbf{XPaste + AAS \emph{(ours)}} & 52.1 & 43.3 & 8.8 & 50.9 & 43.8 & \underline{7.1} \\
\midrule
\multicolumn{7}{l}{\emph{Grounding DINO Swin-B (AP)}} \\
Zero Shot & 57.7 & 51.4 & 6.3 & 56.3 & 50.9 & 5.4 \\
SCP & 59.6 & 53.4 & 6.1 & 58.3 & 53.2 & 5.1 \\
XPaste & \underline{59.7} & \textbf{54.1} & \textbf{5.6} & \textbf{58.6} & \textbf{53.6} & \underline{4.9} \\
\rowcolor{oursrow} SCP + AAS \emph{(ours)} & \textbf{59.8} & 53.6 & 6.2 & \underline{58.5} & \underline{53.3} & 5.1 \\
\rowcolor{oursrow} \textbf{XPaste + AAS \emph{(ours)}} & \underline{59.7} & \underline{53.7} & \underline{6.0} & 58.4 & \textbf{53.6} & \textbf{4.8} \\
\midrule
\multicolumn{7}{l}{\emph{DINOv3 ViT-B/16 (F1$_{\text{micro}}$)}} \\
No aug & 77.8 & 70.1 & 7.7 & 76.7 & 72.9 & 3.8 \\
Std. & 79.8 & 74.3 & 5.5 & 79.8 & 76.1 & 3.7 \\
SCP & 78.8 & 75.1 & \textbf{3.7} & 79.3 & 76.5 & 2.8 \\
XPaste & \underline{80.2} & \underline{76.1} & 4.1 & \underline{80.0} & \underline{77.8} & \textbf{2.2} \\
\rowcolor{oursrow} SCP + AAS \emph{(ours)} & 79.0 & 75.0 & 4.0 & 79.1 & 76.8 & \underline{2.3} \\
\rowcolor{oursrow} \textbf{XPaste + AAS \emph{(ours)}} & \textbf{80.6} & \textbf{76.8} & \underline{3.9} & \textbf{80.6} & \textbf{78.0} & 2.5 \\
\bottomrule
\end{tabular}

\end{table*}

\noindent\textbf{AAS driven by Objects365 NPMI.} We re-ran the XPaste + AAS recipe with Objects365 NPMI wherever NPMI enters training: the generated training scenes, their typical-partner filter and the AAS tilt; the benchmark stays on COCO NPMI. The two sources admit largely the same training pairs ($542$ shared of $603$ COCO and $562$ Objects365 pairs), so this check asks whether the result survives the change of source, not whether it transfers to a disjoint set of pairs. In both families the two AAS rows differ by at most $0.3$~p.p.\@ on any column, about the spread over training draws, and both stay ahead of XPaste on STRANGE-Bench and on its held-out classes (Tab.~\ref{tab:app_circ_o365_aas}).

\begin{table*}[!htp]%
\centering
\footnotesize
\setlength{\tabcolsep}{4pt}
\renewcommand{\arraystretch}{0.9}
\caption{\textbf{XPaste + AAS driven by COCO or by Objects365 NPMI.} The two AAS rows differ by at most $0.3$~p.p.\@ on any column, and both stay ahead of XPaste on STRANGE-Bench and on its held-out classes. Objects365 NPMI replaces COCO NPMI in the generated training scenes, their filter and the AAS tilt; the benchmark stays on COCO NPMI. Means over $n$ training draws; Co-DINO rows are late-stage models trained on mini-COCO (App.~\ref{app:beta_sweep}). Held-out classes: the subset of STRANGE-Bench whose classes never appear in the generated training scenes.}
\label{tab:app_circ_o365_aas}
\begin{tabular}{l c cc cc cc}
\toprule
& & \multicolumn{2}{c}{COCO} & \multicolumn{2}{c}{STRANGE-Bench} & \multicolumn{2}{c}{Held-out classes} \\
\cmidrule(lr){3-4} \cmidrule(lr){5-6} \cmidrule(lr){7-8}
Augmentation & $n$ & val\,$\uparrow$ & $\Delta^{\text{T-A}}\downarrow$ & $\Delta_{1\text{-}5}\downarrow$ & $T_{\text{never}}\uparrow$ & $\Delta_{1\text{-}5}\downarrow$ & $T_{\text{never}}\uparrow$ \\
\midrule
\multicolumn{8}{l}{\emph{Co-DINO R-50, mini-COCO late-stage protocol (AP)}} \\
XPaste & 6 & \textbf{43.3} & \underline{8.0} & 1.0 & \underline{87.3} & \underline{1.4} & 87.7 \\
\rowcolor{oursrow} \textbf{XPaste + AAS, COCO NPMI \emph{(ours)}} & 4 & \underline{43.2} & \textbf{7.8} & \underline{0.2} & \textbf{89.3} & \textbf{0.5} & \underline{89.6} \\
\rowcolor{oursrow} \textbf{XPaste + AAS, Objects365 NPMI \emph{(ours)}} & 2 & \underline{43.2} & 8.1 & \textbf{0.1} & \textbf{89.3} & \textbf{0.5} & \textbf{89.9} \\
\midrule
\multicolumn{8}{l}{\emph{DINOv3 ViT-B/16 (F1$_{\text{micro}}$)}} \\
XPaste & 4 & 82.3 & \underline{4.2} & \underline{$-0.6$} & 95.1 & $-0.9$ & 95.2 \\
\rowcolor{oursrow} \textbf{XPaste + AAS, COCO NPMI \emph{(ours)}} & 4 & \textbf{82.5} & \textbf{4.1} & \textbf{\boldmath $-0.8$} & \textbf{96.1} & \textbf{\boldmath $-1.2$} & \textbf{96.0} \\
\rowcolor{oursrow} \textbf{XPaste + AAS, Objects365 NPMI \emph{(ours)}} & 4 & \underline{82.4} & \underline{4.2} & \textbf{\boldmath $-0.8$} & \underline{95.9} & \underline{$-1.1$} & \underline{95.7} \\
\bottomrule
\end{tabular}

\end{table*}

\noindent\textbf{Held-out classes.} The generated training scenes of the AAS late stage use only part of the classes; the scenes of STRANGE-Bench whose classes all come from the other part form a held-out subset ($3{,}234$ images). On it (Tab.~\ref{tab:app_circ_heldout}), XPaste\,+\,AAS keeps the highest atypical score and $T_{\text{never}}$ in every family, and each AAS row scores higher on both than the row it starts from, with a gap no larger (DINOv3 XPaste and XPaste\,+\,AAS tie at $-1.1$). The gains of the AAS rows therefore do not come from classes seen in generated training scenes.

\begin{table*}[!htp]%
\centering
\footnotesize
\setlength{\tabcolsep}{4pt}
\renewcommand{\arraystretch}{0.9}
\caption{\textbf{STRANGE-Bench and its held-out-class subset.} On the held-out classes, XPaste\,+\,AAS keeps the highest atypical score and $T_{\text{never}}$ in every family, and each AAS row improves both over the row it starts from. Baseline, paste and AAS rows of Tabs.~\ref{tab:codino_main} to~\ref{tab:dinov3_main}; the held-out subset contains only classes that never appear in the generated training scenes of the AAS rows. Means over $T_1$ to $T_5$; detectors: AP; DINOv3: F1$_{\text{micro}}$.}
\label{tab:app_circ_heldout}
\begin{tabular}{l ccc ccc}
\toprule
& \multicolumn{3}{c}{STRANGE-Bench} & \multicolumn{3}{c}{Held-out classes} \\
\cmidrule(lr){2-4} \cmidrule(lr){5-7}
Augmentation & $T_{1\text{-}5}^{\text{A}}\uparrow$ & $\Delta_{1\text{-}5}\downarrow$ & $T_{\text{never}}\uparrow$ & $T_{1\text{-}5}^{\text{A}}\uparrow$ & $\Delta_{1\text{-}5}\downarrow$ & $T_{\text{never}}\uparrow$ \\
\midrule
\multicolumn{7}{l}{\emph{Co-DINO R-50 (AP)}} \\
No aug & 76.4 & 7.0 & 76.3 & 76.6 & 7.3 & 76.7 \\
Std. & 81.9 & 6.3 & 83.3 & 81.6 & 7.3 & 84.1 \\
SCP & 86.3 & 3.9 & 87.5 & 86.1 & 4.9 & 88.1 \\
XPaste & 89.0 & \underline{1.7} & \underline{90.4} & \underline{89.1} & \underline{2.2} & 89.7 \\
\rowcolor{oursrow} SCP + AAS \emph{(ours)} & \underline{89.2} & 2.1 & 90.1 & \underline{89.1} & 2.8 & \underline{90.5} \\
\rowcolor{oursrow} \textbf{XPaste + AAS \emph{(ours)}} & \textbf{90.7} & \textbf{1.0} & \textbf{92.1} & \textbf{90.8} & \textbf{0.9} & \textbf{92.7} \\
\midrule
\multicolumn{7}{l}{\emph{Grounding DINO Swin-B (AP)}} \\
Zero Shot & 90.9 & 2.4 & 92.4 & 91.2 & 2.8 & 93.0 \\
SCP & 92.4 & 0.9 & 93.3 & 92.7 & 1.4 & 93.6 \\
XPaste & 93.6 & \textbf{0.3} & \underline{94.6} & 93.3 & 1.5 & 95.0 \\
\rowcolor{oursrow} SCP + AAS \emph{(ours)} & \underline{93.7} & \underline{0.6} & 94.5 & \underline{93.6} & \underline{1.3} & \underline{95.1} \\
\rowcolor{oursrow} \textbf{XPaste + AAS \emph{(ours)}} & \textbf{94.7} & \textbf{0.3} & \textbf{95.2} & \textbf{94.4} & \textbf{1.1} & \textbf{95.7} \\
\midrule
\multicolumn{7}{l}{\emph{DINOv3 ViT-B/16 (F1$_{\text{micro}}$)}} \\
No aug & 66.3 & 19.2 & 61.5 & 67.4 & 18.9 & 62.7 \\
Std. & 67.5 & 22.2 & 62.6 & 69.0 & 20.9 & 62.8 \\
SCP & 87.1 & 4.3 & 86.5 & 86.8 & 4.5 & 86.3 \\
XPaste & \underline{94.4} & \underline{$-0.7$} & \underline{95.0} & \underline{94.6} & \textbf{\boldmath $-1.1$} & \underline{95.3} \\
\rowcolor{oursrow} SCP + AAS \emph{(ours)} & 91.2 & 1.2 & 89.5 & 90.2 & \underline{2.1} & 88.3 \\
\rowcolor{oursrow} \textbf{XPaste + AAS \emph{(ours)}} & \textbf{95.5} & \textbf{\boldmath $-0.8$} & \textbf{96.1} & \textbf{95.5} & \textbf{\boldmath $-1.1$} & \textbf{95.9} \\
\bottomrule
\end{tabular}

\end{table*}

\noindent\textbf{A second generator.} We re-rendered $999$ held-out-class scenes with SD 3.5 Medium from the captions of the benchmark and passed them through the same audit and filter; enough complete scenes survive for $T_4$ ($67$ per arm) and $T_{\text{never}}$ ($136$). The atypical arm of $T_4$ scores lower on these renders for every model, but every AAS row keeps a smaller $T_4$ gap and a higher $T_{\text{never}}$ than its control (Tab.~\ref{tab:app_circ_xgen}), so the gain of AAS is not tied to the generator of its training scenes. These paste rows are late-stage models of our development runs, trained on mini-COCO (App.~\ref{app:beta_sweep}; recipes in App.~\ref{app:exp-recipes}).

\begin{table*}[!htp]%
\centering
\footnotesize
\setlength{\tabcolsep}{4pt}
\renewcommand{\arraystretch}{0.9}
\caption{\textbf{Held-out-class scenes rendered by the generator of the benchmark (SD 3.5 Large) and by SD 3.5 Medium.} On the SD 3.5 Medium renders the atypical arm of $T_4$ scores lower for every model, yet every AAS row keeps a smaller $T_4$ gap and a higher $T_{\text{never}}$ than its control. Both generators render from the same captions, at $k = 3$ ($T_4$) and for $T_{\text{never}}$. Paste rows: mini-COCO late-stage models, means over training draws.}
\label{tab:app_circ_xgen}
\begin{tabular}{l cccc cccc}
\toprule
& \multicolumn{4}{c}{SD 3.5 Large (benchmark generator)} & \multicolumn{4}{c}{SD 3.5 Medium} \\
\cmidrule(lr){2-5} \cmidrule(lr){6-9}
Augmentation & $T_4^{\text{T}}\uparrow$ & $T_4^{\text{A}}\uparrow$ & $\Delta_4\downarrow$ & $T_{\text{never}}\uparrow$ & $T_4^{\text{T}}\uparrow$ & $T_4^{\text{A}}\uparrow$ & $\Delta_4\downarrow$ & $T_{\text{never}}\uparrow$ \\
\midrule
\multicolumn{9}{l}{\emph{Co-DINO R-50 (AP; paste rows: mini-COCO late-stage protocol)}} \\
Std. & \underline{87.3} & 79.7 & 7.6 & 84.1 & \textbf{84.9} & 74.3 & 10.6 & 80.7 \\
XPaste & 86.2 & \underline{86.9} & \underline{$-0.7$} & 87.6 & 83.1 & 80.4 & \underline{2.8} & \underline{88.0} \\
\rowcolor{oursrow} \textbf{XPaste + AAS \emph{(ours)}} & \textbf{87.4} & \textbf{89.1} & \textbf{\boldmath $-1.8$} & \textbf{89.4} & 84.1 & \textbf{82.3} & \textbf{1.8} & \textbf{89.2} \\
SCP & 85.5 & 82.2 & 3.3 & 84.9 & 83.6 & 79.6 & 4.1 & 83.6 \\
\rowcolor{oursrow} SCP + AAS \emph{(ours)} & 86.9 & 84.9 & 2.0 & \underline{87.9} & \underline{84.4} & \underline{81.1} & 3.3 & 86.9 \\
\midrule
\multicolumn{9}{l}{\emph{Grounding DINO Swin-B (AP; paste rows: mini-COCO late-stage protocol)}} \\
Zero Shot & 94.1 & 90.5 & 3.6 & 93.0 & 90.3 & 88.5 & \underline{1.8} & 91.8 \\
XPaste & \underline{94.8} & \underline{92.6} & \underline{2.3} & 94.8 & \textbf{92.1} & 89.9 & 2.2 & \underline{93.8} \\
\rowcolor{oursrow} \textbf{XPaste + AAS \emph{(ours)}} & \textbf{95.3} & \textbf{93.8} & \textbf{1.5} & \textbf{95.8} & \underline{92.0} & \textbf{91.1} & \textbf{0.9} & \textbf{95.0} \\
SCP & 93.8 & 90.7 & 3.1 & 93.8 & 90.5 & 87.6 & 2.9 & 91.7 \\
\rowcolor{oursrow} SCP + AAS \emph{(ours)} & 94.6 & 92.2 & 2.4 & \underline{95.0} & \textbf{92.1} & \underline{90.3} & \underline{1.8} & 93.4 \\
\bottomrule
\end{tabular}

\end{table*}

\subsection{Pretraining Data and Model Capacity}
\label{app:scale_ladder}

\begin{figure}[!htp]%
\centering
\includegraphics[width=\linewidth]{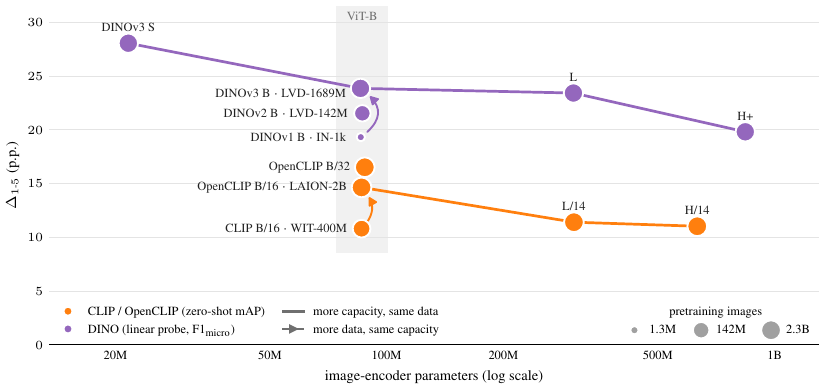}
\caption{\textbf{Pretraining data and model capacity on STRANGE-Bench.} Adding capacity on the same pretraining data (solid lines) lowers the gap in both families; adding pretraining data at ViT-B capacity (curved arrows, gray band) raises it. $\Delta_{1\text{-}5}$: mean typical minus atypical over $T_1$ to $T_5$, against the image-encoder parameters; dot area grows with the logarithm of the number of pretraining images. CLIP and OpenCLIP: zero-shot mAP; DINO: linear probe on COCO, F1$_{\text{micro}}$.}
\label{fig:app_scale_ladder}
\end{figure}

\noindent The families of Sec.~\ref{sec:results_diagnostic} differ in pretraining data, capacity and objective at once. Fig.~\ref{fig:app_scale_ladder} separates the first two on STRANGE-Bench, changing one while holding the other fixed (metrics as in Fig.~\ref{fig:strange_diagnostic}: zero-shot mAP for CLIP and OpenCLIP, linear-probe F1$_{\text{micro}}$ for DINO). More pretraining data at fixed capacity does not narrow the gap. It grows from CLIP~\cite{radford2021learning} (WIT-400M) to OpenCLIP~\cite{cherti2023reproducible} (LAION-2B), $10.8 \rightarrow 14.7$, and from DINOv1~\cite{caron2021emerging} (ImageNet-1k) over DINOv2~\cite{oquab2024dinov} (LVD-142M) to DINOv3~\cite{simeoni2025dinov3} (LVD-1689M), $19.3 \rightarrow 21.5 \rightarrow 23.9$. More capacity at fixed data narrows the gap without closing it: from $16.5$ to $11.0$ for OpenCLIP ViT-B/32 to ViT-H/14, with little change after ViT-L/14 ($11.4$), and from $28.1$ to $19.8$ for DINOv3 ViT-S to ViT-H+ ($T_{\text{never}}$: $19.0 \rightarrow 52.8$).

\subsection{Spatial Arrangement as a Typicality Axis}\label{app:spatial-arrangement}
NPMI scores whether two classes co-occur, not how they are arranged. To test whether models also rely on the usual arrangement of a pair, we measure how stereotyped it is: for each ordered COCO \texttt{train2017} pair with at least $10$ co-occurring images, we take the direction from the largest anchor box to the largest partner box in every such image, bin it into eight $45^\circ$ sectors, and compute the normalized entropy $H \in [0, 1]$ of the histogram. Low $H$ means one dominant arrangement, such as a rider above a horse. STRANGE-Bench layouts are generated without regard to these arrangements, so a model that relies on them should recognize the anchor less often in stereotyped pairs. We use the single-partner tier $T_2$ (both arms) and the \dinov{} ViT-B linear probe, count an anchor as recognized when its score is at least $0.5$, and keep pairs with at least two scenes.

\begin{table}[!htp]%
\centering
\small
\setlength{\tabcolsep}{4pt}%
\caption{\textbf{No measurable effect of a stereotyped arrangement on anchor recognition.} Neither interval of the correlation with recognition excludes zero, and its sign differs between the two versions. $H$: normalized entropy of the anchor-to-partner direction in COCO \texttt{train2017} (low $H$: one dominant arrangement); recognition: \dinov{} ViT-B linear probe on $T_2$, anchor score $\geq 0.5$, per tertile of $H$. Brackets: 95\% pair-bootstrap intervals.}
\label{tab:app_spatial_entropy}
\begin{tabular}{l c ccc cc}
\toprule
& & \multicolumn{3}{c}{Recognition (\%)} & \multicolumn{2}{c}{Pearson $r$ of $H$ with} \\
\cmidrule(lr){3-5} \cmidrule(lr){6-7}
Benchmark & Pairs (scenes) & low $H$ & mid $H$ & high $H$ & recognition & NPMI \\
\midrule
STRANGE-Bench & 86 (221) & 81.0 & 80.2 & 71.4 & $-0.09$ [$-0.26$, 0.09] & 0.39 [0.16, 0.57] \\
Initial version & 139 (384) & 59.8 & 66.5 & 69.7 & 0.10 [$-0.06$, 0.27] & 0.37 [0.20, 0.51] \\
\bottomrule
\end{tabular}

\end{table}

Tab.~\ref{tab:app_spatial_entropy} gives the result. On STRANGE-Bench, anchor recognition is highest for the most stereotyped pairs (Pearson $r(H, \text{recognition}) = -0.09$); on an initial version of the benchmark the order reverses ($r = 0.10$). Neither interval excludes zero, and $H$ correlates with the NPMI of the pair, so arrangement and co-occurrence are entangled at this sample size. We therefore find no measurable effect of spatial arrangement beyond co-occurrence. A direct test would render the same pair in its usual and in a violated arrangement, which the layout stage of the pipeline (\S\ref{sec:strange}) supports; we leave this, and object count as a further axis, to future work.

\subsection{Partner Masking}
\label{app:partner_mask}

\noindent Instead of changing which objects appear together, a mitigation can remove the objects around a target. On COCO \texttt{train2017} images with exactly two classes ($35{,}858$ of $117{,}266$), partner masking picks one class as the anchor, removes the other with a precomputed LaMa~\cite{suvorov2022resolution} inpainting and keeps only the anchor label; other images are unchanged. We fine-tune DINOv3 ViT-B/16 with it for 20 epochs, the recipe of the submitted runs, and compare it with the Std.\ run of the same recipe (Tab.~\ref{tab:app_partner_mask}). Partner masking narrows the STRANGE-Bench gap only from $20.0$ to $18.2$, while the COCO gap widens from $4.8$ to $5.2$ and both $T_{\text{never}}$ ($67.2 \rightarrow 65.8$) and COCO val ($80.9 \rightarrow 80.1$) drop. A model learns the partners it is shown with each anchor. Partner masking shrinks that set, since the anchor is now often seen alone, so an anchor among atypical partners at test time remains unlike the training images; paste-based augmentation (Sec.~\ref{sec:results_mitigation}) widens the set instead. The invariance to context comes from varying the context, not from deleting it.

\begin{table}[!htp]%
\centering
\small
\caption{\textbf{Partner masking against the matched Std.\ run.} Removing the partner narrows the STRANGE-Bench gap slightly but widens the COCO gap and lowers $T_{\text{never}}$ and COCO accuracy. DINOv3 ViT-B/16 fine-tuned for 20 epochs (the submission recipe), F1$_{\text{micro}}$. Better value per column in bold.}
\label{tab:app_partner_mask}
\begin{tabular}{l cccc cccc}
\toprule
& \multicolumn{4}{c}{\textbf{COCO}} & \multicolumn{4}{c}{\textbf{STRANGE-Bench}} \\
\cmidrule(lr){2-5} \cmidrule(lr){6-9}
Augmentation & $\text{F1}^{\text{val}}$ & $\text{F1}^{\text{T}}$ & $\text{F1}^{\text{A}}$ & $\Delta^{\text{T-A}}$ & $T_{1\text{-}5}^{\text{T}}$ & $T_{1\text{-}5}^{\text{A}}$ & $\Delta_{1\text{-}5}$ & $T_{\text{never}}$ \\
\midrule
Std. & \textbf{80.9} & \textbf{79.0} & \textbf{74.2} & \textbf{4.8} & \textbf{88.5} & 68.5 & 20.0 & \textbf{67.2} \\
Partner masking & 80.1 & 78.3 & 73.1 & 5.2 & 88.2 & \textbf{70.0} & \textbf{18.2} & 65.8 \\
\bottomrule
\end{tabular}

\end{table}

\subsection{Variability of the AAS Rows over Training Runs}
\label{app:aas_draws}

\noindent Tabs.~\ref{tab:codino_main} to~\ref{tab:dinov3_main} show one training run per AAS row. Grounding DINO has two draws per row: the late stage (epochs 9 to 12) run twice from the same epoch-8 checkpoint without a fixed seed. DINOv3 has four seeds ($42$ to $45$), each a separate run from the pretrained backbone. Tab.~\ref{tab:app_aas_draws} gives the mean and standard deviation over the runs of each row. The control rows of the tables are single runs.

\begin{table*}[!htp]%
\centering
\small
\setlength{\tabcolsep}{3.3pt}%
\caption{\textbf{Spread of the AAS rows over training runs.} The standard deviation is at most $0.5$ p.p.\ in every column. Mean $\pm$ sample standard deviation of the AAS rows of Tabs.~\ref{tab:grounding_dino_main} and~\ref{tab:dinov3_main} on COCO and STRANGE-Bench: Grounding DINO over two late-stage draws, DINOv3 over seeds $42$ to $45$.}
\label{tab:app_aas_draws}
\begin{tabular}{l cccc cccc}
\toprule
& \multicolumn{4}{c}{\textbf{COCO}} & \multicolumn{4}{c}{\textbf{STRANGE-Bench}} \\
\cmidrule(lr){2-5} \cmidrule(lr){6-9}
Method & val & T & A & $\Delta^{\text{T-A}}$ & $T_{1\text{-}5}^{\text{T}}$ & $T_{1\text{-}5}^{\text{A}}$ & $\Delta_{1\text{-}5}$ & $T_{\text{never}}$ \\
\midrule
\multicolumn{9}{l}{\textbf{Grounding DINO Swin-B (AP, Tab.~\ref{tab:grounding_dino_main})}, 2 late-stage draws} \\
SCP + AAS & $58.6{\pm}0.0$ & $59.7{\pm}0.1$ & $53.6{\pm}0.0$ & $6.1{\pm}0.1$ & $94.3{\pm}0.0$ & $93.5{\pm}0.3$ & $0.8{\pm}0.3$ & $94.4{\pm}0.1$ \\
XPaste + AAS & $58.9{\pm}0.1$ & $59.7{\pm}0.1$ & $53.8{\pm}0.1$ & $5.9{\pm}0.1$ & $94.9{\pm}0.0$ & $94.7{\pm}0.0$ & $0.3{\pm}0.0$ & $95.2{\pm}0.1$ \\
\midrule
\multicolumn{9}{l}{\textbf{DINOv3 ViT-B/16 fine-tuned (F1$_{\text{micro}}$, Tab.~\ref{tab:dinov3_main})}, 4 seeds} \\
SCP + AAS & $81.6{\pm}0.1$ & $79.0{\pm}0.2$ & $74.9{\pm}0.2$ & $4.1{\pm}0.2$ & $92.4{\pm}0.2$ & $90.6{\pm}0.5$ & $1.8{\pm}0.4$ & $89.3{\pm}0.2$ \\
XPaste + AAS & $82.5{\pm}0.1$ & $80.5{\pm}0.2$ & $76.4{\pm}0.3$ & $4.1{\pm}0.2$ & $94.8{\pm}0.2$ & $95.6{\pm}0.1$ & $-0.8{\pm}0.2$ & $96.1{\pm}0.2$ \\
\bottomrule
\end{tabular}

\end{table*}

\clearpage

\section{Experimental Setup and Design}
\label{app:exp-setup}

This section first gathers all reproducibility-relevant setup information (evaluation datasets, the eleven reference models, checkpoints, training and evaluation recipes with the AAS sampler of Alg.~\ref{alg:aas}, and compute; \S\ref{app:experimental-details}) and then revisits a design choice that underpins the entire diagnostic: the quantile cut $\alpha$ that defines typical and atypical partner sets (\S\ref{app:splits-impl}). The latter subsection also reports a small $\alpha$-sweep on Co-DINO R-50 and closes with a discussion of why natural-image evaluation alone is structurally insufficient, motivating the synthetic STRANGE-Bench evaluation in~\S\ref{app:strange-pipeline}.%

\subsection{Experimental Details and Study}\label{app:experimental-details}

This subsection collects all information needed to reproduce our diagnostic and ablation results: the evaluation datasets (\S\ref{app:exp-datasets}), the eleven reference models we benchmark (\S\ref{app:exp-models}), the pretrained / fine-tuned checkpoints we load (\S\ref{app:exp-ckpts}), one canonical training and evaluation recipe per model family (\S\ref{app:exp-recipes}), and the compute spent (\S\ref{app:exp-compute}).

\subsubsection{Evaluation Datasets}\label{app:exp-datasets}

The eval suite combines one full-COCO baseline~\cite{mscoco14}, four COCO/Objects365~\cite{shao2019objects365} paired splits, and STRANGE-Bench (Tab.~\ref{tab:app-eval-datasets}). The four paired splits are constructed as in \S\ref{app:splits-impl}: per-anchor ranking by COCO \texttt{train2017} NPMI, applied to val images, quintile cut for the headline ($\alpha = 0.2$) and decile for an additional ablation ($\alpha = 0.1$); the typical side is down-sampled per-anchor to approximately match the atypical-side image count. O365 splits apply the same construction, with the same COCO \texttt{train2017} NPMI matrix, to Objects365~v2 val images whose 365 classes are mapped onto the 80 COCO classes ($69$ direct name matches, $11$ approximate). The cross-domain tables (Tabs.~\ref{tab:codino_cd}, \ref{tab:gdino_o365} and~\ref{tab:dinov3_ft_cd}) add a fifth paired split, a third one on O365: the quintile construction on the same val images with the NPMI matrix recomputed on O365~train ($1{,}699{,}452$ images) and the partner vocabulary restricted to the COCO/O365 overlap, i.e.\ the same $80$ mapped classes ($4{,}349 / 3{,}834$ typical-matched / atypical images, $76$ anchors). Its partner sets overlap those of the COCO-NPMI split with a mean per-anchor Jaccard index of $0.64$ (typical) and $0.44$ (atypical). Detection runs against full annotations on every split. STRANGE-Bench is reported in the balanced variant (typical and atypical sub-tiers of a tier cut to the same size; $5{,}523$ images across 17 sub-tiers, 11 of them in the main tables); per-tier composition, generation pipeline, and audit are in \S\ref{app:strange-pipeline}.

\begin{table}[h]
\centering
\small
\caption{\textbf{Evaluation suite.} Paired splits contain a typicality
contrast (matched typical / atypical sub-splits). Per-tier
STRANGE-Bench composition is reported in \S\ref{app:strange:tiers}.}
\label{tab:app-eval-datasets}
\begin{tabular}{@{}llrl@{}}
\toprule
Dataset & Split & \#imgs & Pairing \\
\midrule
COCO val2017~\cite{mscoco14} & --                                  & $5{,}000$ & none (baseline) \\
COCO paired (quintile)  & typ.-matched / atyp.     & $465 / 389$  & paired \\
COCO paired (decile)    & typ.-matched / atyp.     & $300 / 259$  & paired \\
O365 paired (quintile)  & typ.-matched / atyp.     & $4{,}298 / 4{,}258$ & paired \\
O365 paired (decile)    & typ.-matched / atyp.     & $2{,}495 / 2{,}428$ & paired \\
O365 paired (quintile, O365 NPMI) & typ.-matched / atyp. & $4{,}349 / 3{,}834$ & paired \\
\midrule
STRANGE-Bench & 17 sub-tiers           & $5{,}523$    & tiered (\S\ref{app:strange-pipeline}) \\
\bottomrule
\end{tabular}
\end{table}

\subsubsection{Models}\label{app:exp-models}

We benchmark eleven reference models spanning five task families (specialist detection, open-vocab detection, probed classification, VL zero-shot, instruction-tuned LMM); see Tab.~\ref{tab:app-models}. The five families correspond to the five rows of the diagnostic survey in Sec.~\ref{sec:results_diagnostic}.

\begin{table}[h]
\centering
\small
\caption{\textbf{Reference models.} Reported metric reflects what each model natively emits (mAP for detectors and ranking-based classifiers, F1 for binary-membership outputs). LP = linear probe; FT = fine-tuned; ZS~=~zero-shot.}%
\label{tab:app-models}
\resizebox{\textwidth}{!}{
\begin{tabular}{@{}lllll@{}}
\toprule
Model & Backbone & Schedule / setting & Task family & Reported metric \\
\midrule
\codino{} R-50~\cite{zhang2022dino}            & ResNet-50      & \begin{tabular}[t]{@{}l@{}}12 epochs at the base LR and\\one fine-tuning epoch at $0.1\times$ LR\end{tabular}        & specialist det.   & mAP \\
\codino{} Swin-L~\cite{zhang2022dino}          & Swin-L         & 12-epoch ($1\times$)        & specialist det.   & mAP \\
\codino{} Swin-L~\cite{zhang2022dino}          & Swin-L         & 36-epoch ($3\times$)        & specialist det.   & mAP \\
\codino{} Swin-L~\cite{zhang2022dino}          & Swin-L         & Objects365 $\rightarrow$ COCO & specialist det.   & mAP \\
\gdino{} Swin-B~\cite{liu2024grounding}           & Swin-B         & released ZS                  & open-vocab det.   & mAP \\
\dinov{} ViT-B/16 (LP)~\cite{simeoni2025dinov3}    & ViT-B/16       & linear probe                 & probed cls.       & F1\textsubscript{micro} \\
\dinov{} ViT-B/16 (FT)~\cite{simeoni2025dinov3}    & ViT-B/16       & fine-tuned                   & probed cls.       & F1\textsubscript{micro} \\
CLIP ViT-B/16~\cite{radford2021learning}             & ViT-B/16       & released ZS                  & VL ZS cls.        & mAP \\
\openclip{} ViT-B/16~\cite{cherti2023reproducible}      & ViT-B/16       & LAION-2B~\cite{schuhmann2022laionb}, ZS                 & VL ZS cls.        & mAP \\
\llava{} 0.5b~\cite{li2024llava}             & Qwen2-0.5B~\cite{qwen2024qwen25}     & instruction-tuned            & LMM (gen.)        & F1\textsubscript{macro} \\
\llava{} 7b~\cite{li2024llava}               & Qwen2-7B~\cite{qwen2024qwen25}       & instruction-tuned            & LMM (gen.)        & F1\textsubscript{macro} \\
\bottomrule
\end{tabular}
}
\end{table}

\subsubsection{Checkpoints}\label{app:exp-ckpts}

Detection checkpoints come from the public Co-DETR~\cite{zong2023detrs} 
and Grounding DINO~\cite{liu2024grounding} releases, loaded via MMDetection~\cite{chen2019mmdetection}; classification probes (\dinov{}~\cite{simeoni2025dinov3}, both LP and FT variants) and the fine-tuned Co-DINO R-50 and Grounding DINO models are trained in-house from the corresponding \href{https://huggingface.co/}{HuggingFace (HF)} / public backbones; CLIP~\cite{radford2021learning} and \openclip{}~\cite{cherti2023reproducible} are loaded via \href{https://github.com/mlfoundations/open_clip}{\texttt{open\_clip}}; \llava{}~\cite{li2024llava} is loaded via HF. Table~\ref{tab:app-ckpts} lists the exact files. 

\begin{table}[h]
\centering
\small
\caption{\textbf{Checkpoints used in the diagnostic and ablation studies.}}
\label{tab:app-ckpts}
\begin{tabularx}{\linewidth}{@{}>{\raggedright\arraybackslash}p{0.44\linewidth}>{\raggedright\arraybackslash}X@{}}
\toprule
Model & Source \\
\midrule
\codino{} R-50                                          & self-fine-tuned \\
\codino{} Swin-L $1\times$~\cite{zong2023detrs}         & MMDetection~\cite{chen2019mmdetection} \\
\codino{} Swin-L $3\times$~\cite{zong2023detrs}         & MMDetection~\cite{chen2019mmdetection} \\
\codino{} Swin-L (Objects365 $\rightarrow$ COCO)~\cite{zong2023detrs}   & MMDetection~\cite{chen2019mmdetection} \\
\gdino{} Swin-B~\cite{liu2024grounding}                 & MMDetection~\cite{chen2019mmdetection} (+ self-fine-tuned) \\
\dinov{} ViT-B/16 (LP)~\cite{simeoni2025dinov3}         & \texttt{timm} (\nolinkurl{vit_base_patch16_dinov3.lvd1689m}) \\
\dinov{} ViT-B/16 (FT)~\cite{simeoni2025dinov3}         & \texttt{timm} (\nolinkurl{vit_base_patch16_dinov3.lvd1689m}) \\ & + self-fine-tuned \\
CLIP ViT-B/16~\cite{radford2021learning}                & \texttt{open\_clip} \\
\openclip{} ViT-B/16~\cite{cherti2023reproducible}      & \texttt{open\_clip} (LAION-2B) \\
\llava{} 0.5b~\cite{li2024llava}                        & HF Hub \\
\llava{} 7b~\cite{li2024llava}                          & HF Hub \\
\bottomrule
\end{tabularx}
\end{table}

\subsubsection{Training and Evaluation Recipes}\label{app:exp-recipes}

We give one canonical recipe per model family. 

\noindent\textbf{AAS sampler.}
The AAS sampler of the AAS rows is shared
across all model families and is summarized in
Alg.~\ref{alg:aas}.
It takes the class of the largest ground-truth object of each training image as anchor and then samples paste classes from
an anchor-conditioned, atypicality-tilted distribution
$P(j \mid a) \propto \pi(j)\exp(\beta A(a, j))$ with
$A(a, j) = (1 - \text{NPMI}(a, j))/2$, $A = 1$ for pairs that never co-occur, and uniform $\pi$. Pasted instances are scaled per-class (DINOv3: uniformly at $0.1$ to $0.4$ of the image short side) and composited with occlusion-aware mask updates; the resulting augmented image and updated ground-truth annotations are passed to the recipe-specific training loop below. The hyperparameters $\beta = 3.5$, never-pair weight $1$, and up to $19$ pastes per image (DINOv3: $1$ to $20$) are held fixed across families; \S\ref{app:beta_sweep} ablates $\beta$.

\begin{algorithm}[h]
\caption{Anchor-Atypicality Sampling (AAS)}
\label{alg:aas}
\small
\begin{algorithmic}[1]
\Require Image $I$ with present classes $\mathcal{P}$ and ground-truth annotations $\text{gt}(I)$
\Require Instance pool $\mathcal{C}_{\text{pool}}$, atypicality function $A(a,j) =$ $(1 - \text{NPMI}(a,j))/2$
\Require Max pastes $M$, tilt $\beta$, class prior $\pi$, per-class scale stats $\{(\mu_c, \sigma_c^2)\}$
\State $a \gets$ class of the largest box in $\text{gt}(I)$ \Comment{anchor: largest object}
\State $n \sim \mathcal{U}\{0, \ldots, M-1\}$ \Comment{number of pastes}
\For{$k = 1, \ldots, n$}
  \State $j \sim P(\cdot \mid a) \propto$ $\pi(\cdot) \exp(\beta A(a, \cdot))$ \Comment{anchor-conditioned, atypicality-tilted}
  \State $\text{inst} \sim \mathcal{C}_{\text{pool}}[j]$ \Comment{sample an instance of class $j$}
  \State $s \sim \mathcal{N}(\mu_j, \sigma_j^2);\; (x, y) \sim \mathcal{U}$ \Comment{per-class scale, random position}
  \State $(I, \text{gt}(I)) \gets \textsc{PasteOcclAware}(I, \text{gt}(I), \text{inst}, s, (x, y))$
\EndFor
\State \Return augmented image $I$ and updated annotations $\text{gt}(I)$
\end{algorithmic}
\end{algorithm}

\noindent\textbf{Specialist detector training (\codino{}).} Backbone: ResNet-50, ImageNet-pretrained. Optimizer: AdamW, base LR $2{\times}10^{-4}$, weight decay $1{\times}10^{-4}$. Schedule: 12 epochs at the base LR and one fine-tuning epoch at $0.1\times$ LR. Augmentation: standard MMDetection train pipeline: resize to short side $\in [480, 800]$ (32-step) or a random-crop branch (resize, crop to $384$ to $600$, resize), random horizontal flip, normalization. Loss: standard Co-DETR collaborative hybrid-assignment loss (one-to-one DETR + one-to-many auxiliary heads). Batch size 16 (the fine-tuning epoch of the A0 to A6 rows: 8). Mixed-precision is \emph{not} used.

\noindent\textbf{Augmentation pipelines for the Co-DINO sweep.}
The No aug baseline (A0) and the eight augmentation variants share the optimizer, schedule, loss,
and batch settings of the \codino training,  and differ only in the
train-time data pipeline (Tab.~\ref{tab:aug_pipelines}). All variants emit fixed $1024{\times}1024$ inputs, terminate in \texttt{PackDetInputs}, and use the identical test-time pipeline (Resize $(1333, 800)$, pad, no augmentation). The two AAS rows (SCP\,+\,AAS, XPaste\,+\,AAS) continue A5 and A8 with the late stage below.

\begin{table}[h]
\centering
\footnotesize
\setlength{\tabcolsep}{4pt}
\caption{\textbf{Train-time augmentation pipelines for the Co-DINO R-50 sweep.} All recipes share the optimizer, schedule, loss, and test-time pipeline; only the train-time data-processing pipeline differs. Source images = images blended into one training sample; LSJ = Large-Scale Jitter ($r \sim \mathcal{U}(0.1, 2.0)$ + $1024{\times}1024$ crop +  flip + pad); PMD = \texttt{PhotoMetricDistortion}; HSV = YOLOX HSV ($\pm 5^\circ / \pm 50\% / \pm 50\%$); 4-tile = YOLOX $2{\times}2$ mosaic + \texttt{RandomAffine} crop back to $1024{\times}1024$. IDs are the config-file prefixes of the released code (\texttt{configs/codino\_r50\_augsweep/a<ID>\_*.py}); rows in the order of Tab.~\ref{tab:codino_main}.}
\label{tab:aug_pipelines}
\resizebox{\textwidth}{!}{
\begin{tabular}{@{}llllll@{}}
\toprule
ID & Aug & Geometric & Photometric & Paste source & Ref. \\
\midrule
A0 & No aug & resize to $1024$ + pad (no flip) & -- & -- & -- \\
A1  & Std.\          & DETR 2-branch resize + flip          & --   & --                          & \cite{zong2023detrs} \\
A4  & Color jitter   & A1 geometry, fixed $1024{\times}1024$ & PMD & --                          & -- \\
A3  & MixUp          & single image, $\mathcal{U}(0.8, 1.6)$ blend & HSV & --                    & \cite{zhang2017mixup} \\
A2  & Mosaic         & 4-tile + \texttt{RandomAffine}       & HSV  & --                          & \cite{bochkovskiy2020yolov4} \\
A15 & YOLOX full     & 4-tile + Affine + $5^{\text{th}}$-img MixUp & HSV & --                  & \cite{ge2021yolox} \\
A5  & LSJ + SCP      & LSJ                                  & --   & 2nd-img all masks (real)    & \cite{ghiasi2021simple} \\
A6  & InstAug        & LSJ                                  & --   & curated cutout pool         & \cite{kupyn2024dataset} \\
A8  & XPaste-base    & LSJ                                  & --   & SD-3.5m~\cite{stabilityai2024sd35} + real, class-balanced ($\leq 19$/img) & \cite{zhao2023x} \\
\bottomrule
\end{tabular}
}
\end{table}

\noindent\textbf{Late stage (AAS rows of Tabs.~\ref{tab:codino_main}, \ref{tab:grounding_dino_main}, \ref{tab:dinov3_main}).} Each AAS row continues a paste-trained run (SCP or XPaste) from its checkpoint after epoch 8 (\codino{}, \gdino{}) or epoch 5 (\dinov{}) for the remaining epochs of the same schedule (\codino{}: 9 to 13, \gdino{}: 9 to 12, \dinov{}: 6 to 10). The training set is COCO \texttt{train2017} plus $7{,}805$ scenes rendered by the STRANGE-Bench pipeline (Sec.~\ref{sec:strange}) from atypical anchor-partner specifications over 40 of the 80 COCO classes. AAS samples the pasted classes (detectors: pastes onto the generated scenes; \dinov{}: every image). The reported weights are the 1:1 average of the late-stage weights and the final weights of the run it continues (\dinov{}: the best validation checkpoints of both),
\begin{equation}
\theta_{\text{AAS}} = \tfrac{1}{2}\left(\theta_{\text{late}} + \theta_{\text{base}}\right),
\label{eq:late_avg}
\end{equation}
where $\theta_{\text{base}}$ are the final weights of the SCP or XPaste run and $\theta_{\text{late}}$ the weights after its late stage. Both share every update up to the branch point (epoch 8; \dinov{}: epoch 5), as in weight averaging of models fine-tuned from a common start~\cite{wortsman2022model,wortsman2022robust}.

\noindent\textbf{Grounding DINO fine-tuning.} The SCP, XPaste, and AAS rows of Tab.~\ref{tab:grounding_dino_main} fine-tune the released Swin-B checkpoint on COCO \texttt{train2017}: 12 epochs with an LR drop at epoch 11, AdamW, LR $1{\times}10^{-4}$ ($0.1\times$ for the backbone), weight decay $1{\times}10^{-4}$, gradient clipping $0.1$, batch 32 (8 GPUs $\times$ 4), multi-scale resize and crop plus the paste of the row.

\noindent\textbf{Detector inference (Co-DETR Swin-L variants, \gdino{}
Swin-B).} Loaded from the public checkpoints listed in Table~\ref{tab:app-ckpts} (survey rows). Batch size~1, no test-time augmentation.

\noindent\textbf{Probed classifier training (\dinov{} ViT-B LP).} 
Frozen DINOv3 ViT-B/16 features, single linear classification head over
$80$ COCO classes, multi-label BCE-with-logits loss. Optimizer: AdamW, 
LR $1{\times}10^{-3}$, weight decay $0$. Schedule: 20 epochs, cosine decay. Train data: COCO \texttt{train2017} (image-level multi-hot labels from instance annotations). No backbone fine-tuning, no augmentation beyond resize+normalize (paste rows of App.~\ref{app:dinov3_lp}: plus the paste). F1 uses a fixed threshold of $0.5$ (F1\textsubscript{micro}). 

\noindent\textbf{Fine-tuned classifier training (\dinov{} ViT-B/16 FT).}
Same architecture as the LP recipe (DINOv3 ViT-B/16 backbone + single linear head over $80$ COCO classes, multi-label BCE-with-logits) but with the backbone unfrozen. Optimizer: AdamW, head LR $5{\times}10^{-4}$ with layer-wise decay $0.75$ over the backbone, weight decay $0.05$, batch 128, $224$\,px inputs. Schedule: 10 epochs, cosine decay with one warmup epoch. Train data: COCO \texttt{train2017}. Augmentation matches the recipe under study (Std., Color jitter, MixUp, Mosaic, YOLOX (Mosaic with strong color jitter and no MixUp, which on a four-image Mosaic adds multi-label noise), SCP, XPaste, InstAug, SCP+AAS,
XPaste+AAS); see Sec.~\ref{sec:experiments} for the full suite. F1 uses the fixed threshold $0.5$ (F1\textsubscript{micro}). 

\noindent\textbf{Zero-shot VL inference (CLIP / \openclip{} ViT-B/16).}
Each class name is wrapped in the standard 80-template ensemble from
\citep{radford2021learning} (templates such as ``a photo of a
\{class\}.''). Per-class text embeddings are mean-pooled across
templates and L2-normalized. Per-image embeddings come from the CLIP /
\openclip{} image encoder, also L2-normalized. The score for class
$c$ on image $i$ is the cosine similarity, used directly as the
ranking score for per-class mAP. No threshold sweep is needed for
mAP; for F1 we use a fixed threshold (which is why F1 is uninformative
for the zero-shot models and is omitted from the headline tables).

\noindent\textbf{LMM inference (\llava{} 0.5b / 7b).}
Two-pass ``gated'' prompting. Both prompts are issued as
chat-template \texttt{user} turns with the image attached:
(i) a \emph{main-objects} prompt asks the model to list up to $12$
unique visible object categories as a JSON array;
(ii) a \emph{small-objects} prompt then asks for up to $8$
small handheld items (utensils, containers, etc.).

\begin{quote}\small
\textbf{Pass 1 (main objects).} \emph{List the object categories
that are actually visible in the image. Return ONLY a JSON array
of at most $12$ UNIQUE short nouns ($1$--$2$ words). If you are
unsure, omit the object. No explanations. No extra text.}

\medskip
\textbf{Pass 2 (small objects).} \emph{List ONLY small objects
that are CLEARLY visible (utensils / containers / handheld items).
Return ONLY a JSON array of at most $8$ UNIQUE short nouns ($1$--$2$
words). If you are not $100\%$ sure an object is present, DO NOT
include it. Do NOT guess utensils. No explanations. No extra
text.}
\end{quote}

The text output of each pass is regex-parsed (multiple JSON arrays
/ dicts / quoted strings, plus a comma-split fallback), normalized
(lower-cased, basic plural$\to$singular), and aliased onto the
COCO-80 vocabulary (\eg ``sofa''$\to$``couch'', ``mobile phone''$\to$``cell phone''). The resulting predicted class labels are converted to a binary multi-hot vector over the $80$ COCO classes and compared against per-image GT. Greedy decoding (\texttt{do\_sample=False}), repetition penalty
$1.2$, \texttt{max\_new\_tokens}$=96$ (main pass) / $64$ (small
pass).

\noindent\emph{Note}: this hand-crafted pipeline only estimates which everyday classes an LMM names in a scene. It is applied identically to the typical and atypical sets, so the gap compares the two sets under the same~pipeline. 

\subsubsection{Compute}\label{app:exp-compute}

All experiments run on shared SLURM clusters; we use three partitions
(\texttt{1}: NVIDIA L40 48\,GB; \texttt{2}: H100 80\,GB; \texttt{3}: A40 48\,GB) as well as local A100 and RTX\,8000 GPUs and H200, MI300A, and A100 GPUs at the MPCDF. All numbers below are wall-clock for one launch, parallel across the array. Table~\ref{tab:app-compute} summarizes.

\begin{table}[h]
\centering
\tiny
\caption{\textbf{Approximate compute per experiment.} Training rows give
\#GPUs\,$\times$\,per-GPU batch, wall-clock, and total GPU-hours
($=$~\#GPUs $\times$ wall-clock). Inference rows are 1 GPU per array task,
all tasks running in~parallel. Two wall-clock values: base run and late stage; rows that list several GPU types sum the GPU-hours of all their runs.}%
\label{tab:app-compute}
\newsavebox{\crcomputebox}%
\sbox{\crcomputebox}{%
\begin{tabular}{@{}llllr@{}}
\toprule
Job & Partition & GPUs $\times$ BS & Wall-clock & GPU-h \\
\midrule
\multicolumn{5}{l}{\emph{Training}} \\
\begin{tabular}[t]{@{}l@{}}\codino{} R-50, full COCO (per recipe A$i$; 12 epochs at the base LR\\and one fine-tuning epoch at $0.1\times$ LR)\end{tabular}
    & \texttt{2}: (H100) & 4 $\times$ 4 & $\approx$~42~h & $\approx$~170 \\
\codino{} R-50 aug-sweep (7 recipes A0 to A4, A6, A15)
    & \texttt{2}: (H100) & 4 $\times$ 4 & 7\,$\times$\,42~h & $\approx$~1{,}190 \\
\dinov{} ViT-B linear probe, COCO train (20 epochs)
    & \texttt{1}: (L40)  & 1 $\times$ 256 & $\approx$~1.5~h & $\approx$~1.5 \\
\dinov{} ViT-B fine-tuning, COCO train (20 epochs)
    & \texttt{2}: (H100) & 2 $\times$ 64 & $\approx$~12~h & $\approx$~24 \\
AAS $\beta$ sweep (App.~\ref{app:beta_sweep}), 3 values per model & \texttt{2}: (H100) & 4 $\times$ 4, 2 $\times$ 64 & $\approx$~10~h, $\approx$~12~h & $\approx$~200 \\
\codino{} R-50 full COCO, bases + late stages (Tab.~\ref{tab:codino_main}) & MI300A, H100, H200 & 4 $\times$ 4, 8 $\times$ 2 & $\approx$~44~h; $\approx$~12--21~h & $\approx$~790 \\
\gdino{} Swin-B full COCO, bases + late stages (Tab.~\ref{tab:grounding_dino_main}) & H100, H200 & 8 $\times$ 4 & $\approx$~20~h; $\approx$~6~h & $\approx$~510 \\
\dinov{} ViT-B FT + late stages (Tab.~\ref{tab:dinov3_main}, App.~\ref{app:circularity}, \ref{app:aas_draws}) & L40, A40, A100, RTX\,8000 & 2--3 $\times$ 42--64 & $\approx$~1--5~h; $\approx$~1--3~h & $\approx$~190 \\
\dinov{} ViT-B LP, paste rows (App.~\ref{app:dinov3_lp}) & L40, A100 & 1 $\times$ 256 & $\approx$~7--11~h & $\approx$~30 \\
Mini-COCO rows of App.~\ref{app:circularity} (\codino{}, \gdino{}) & A100, H200, MI300A, L40 & 2--4 $\times$ 3--10 & $\approx$~15--38~h; $\approx$~6--9~h & $\approx$~700 \\
Linear probes (App.~\ref{app:scale_ladder}), partner masking (App.~\ref{app:partner_mask}) & A40, RTX\,8000 & 2 $\times$ 32--128 & $\approx$~1--4~h & $\approx$~20 \\
\midrule
\multicolumn{5}{l}{\emph{Data generation}} \\
STRANGE-Bench generation ($12{,}375$ specs, $20{,}324$ renders) & L40, A40 & 1 per task & -- & $\approx$~220 \\
Training scenes (App.~\ref{app:exp-recipes}, \ref{app:circularity}), SD~3.5M re-render & A100, L40 & 1 per task & -- & $\approx$~180 \\
\midrule
\multicolumn{5}{l}{\emph{Inference: canonical 5-dataset suite, per model}} \\
COCO val2017 baseline                          & \texttt{1}: (L40) & 1 $\times$ $-$ & $\approx$~10~min & $\approx$~0.17 \\
STRANGE-Bench (17 sub-tiers)                   & \texttt{1}: (L40) & 1 $\times$ $-$ & $\approx$~1.5~min & $\approx$~0.4 \\
Paired real (8 splits)                         & \texttt{1}: (L40) & 1 $\times$ $-$ & $\approx$~15~min & $\approx$~2.0 \\
CLIP / \openclip{} ZS, STRANGE-Bench           & \texttt{1}: (L40) & 1 $\times$ $-$ & $\approx$~1~min  & $\approx$~0.18 \\
\llava{} 0.5b / 7b, STRANGE-Bench + paired     & \texttt{1}: (L40) & 1 $\times$ $-$ & $\approx$~10~min & $\approx$~2.5 \\
\midrule
Evaluation of all reported models & L40, RTX\,8000, H200 & 1 per task & $\approx$~1.5~min / task & $\approx$~190 \\
Total (training, data generation, evaluation) & & & & $\approx$~4{,}200 \\
\bottomrule
\end{tabular}}%
\ifdim\wd\crcomputebox>\linewidth\resizebox{\linewidth}{!}{\usebox{\crcomputebox}}\else\usebox{\crcomputebox}\fi
\end{table}

\noindent\textbf{Total compute footprint.}
Adding the training and inference lines:
\textbf{$\approx$~4{,}200 GPU-hours}, of which $\approx$~3{,}650 for training (the seven \codino{} R-50 augmentation baselines $\approx$~1{,}190, the paste and AAS runs of Tabs.~\ref{tab:codino_main} and~\ref{tab:grounding_dino_main} $\approx$~1{,}300, the mini-COCO rows of App.~\ref{app:circularity} $\approx$~700, the $\beta$ sweep $\approx$~200, and the classification runs $\approx$~270), $\approx$~400 to generate STRANGE-Bench and the training scenes, and $\approx$~190 to evaluate all reported models. Stock Co-DETR Swin-L ($1\times$,
$3\times$, Objects365 $\rightarrow$ COCO) checkpoints and the zero-shot \gdino{} Swin-B
are loaded as released and contribute zero training compute.
CLIP / \openclip{} / \llava{} are zero-shot / off-the-shelf and
contribute zero training compute.
Generating STRANGE-Bench takes $\approx$~220 GPU-hours: $\approx$~155 to render and segment the scenes ($1.6$ SD~3.5L renders per spec), $\approx$~50 for the SAM~3 and OWLv2 passes of the strict filter, and $\approx$~15 for layouts, the first audit, and masks.

The \codino{} R-50 training time is the bottleneck: each run is
12 epochs at the base LR and one fine-tuning epoch at $0.1\times$ LR on full COCO ($\approx$~118k images) at effective batch
$4{\times}4{=}16$ on 4$\times$H100, which clocks in at roughly
3.5~h per epoch.
A \gdino{} Swin-B base run costs about as many GPU-hours ($\approx$~160) but finishes in $\approx$~20~h on eight GPUs (BS=32).

\FloatBarrier
\subsection{Quantile Choices for the Paired Diagnostic}
\label{app:splits-impl}

The paired splits in Sec.~\ref{sec:diagnostic} are parameterized by a
single quantile $\alpha \in (0, 0.5)$: for each anchor $c$, typical
partners are the top-$\alpha$ NPMI quantile of $c$'s row and atypical
partners are the bottom-$\alpha$ quantile. The headline numbers
reported in Sec.~\ref{sec:experiments} use the \textbf{quintile} cut
($\alpha = 0.2$). In this subsection we describe the split
construction (partner-eligibility floors, image-balancing protocol),
report split-quality statistics across five candidate cuts, justify
the quintile choice, and present the full $\alpha$-sweep across the
augmentation suite of Sec.~\ref{sec:experiments}.

\noindent\textbf{Trade-off.}
The choice of $\alpha$ trades partner-set sharpness against split
sample size. Smaller $\alpha$ yields a sharper typical/atypical
contrast but smaller per-anchor partner pools and consequently
smaller per-anchor splits; larger $\alpha$ yields larger and more
stable splits at the cost of admitting partners closer to the
neutral region.

\noindent\textbf{Implementation floors.}
All quantile settings share the same partner-eligibility constraints: 
train-support floors ($N_{ij}^{\text{typ}} \geq 5$, $N_{ij}^{\text{aty}} \geq 1$) prevent wide quantile cuts on noisy long-tail anchor rows from drifting into the neutral region; and a val-observability filter keeps only atypical partners that co-occur with the anchor in at least one val image, so anchors with an empty atypical split are dropped ($16$ of $80$ at the quintile cut). No sign constraint is applied: at the quintile cut $54$ of $850$ typical partner slots have NPMI $\leq 0$ and $3$ of $670$ atypical slots have NPMI $\geq 0$, while the pooled means remain $+0.21$ vs.\ $-0.23$.

\noindent\textbf{Image balancing.}
After the partner-eligibility filters above, the typical and atypical sides of each cut are highly imbalanced in the natural distribution: typical pairs are roughly an order of magnitude more common than atypical ones, so the unfiltered typical-side image pool is much larger than the atypical-side pool (e.g., $3{,}501$ vs.\ $389$ at the quintile cut). To prevent this asymmetry from biasing per-anchor AP estimates, we keep the atypical side intact (it is already the bottleneck) and \emph{down-sample the typical side per-anchor} so that, for each anchor class, the number of typical-matched images is approximately equal to the number of atypical images for that anchor. The resulting post-balancing counts are reported in Table~\ref{tab:alpha_split_quality} and feed directly into the eval suite of \S\ref{app:exp-datasets}. 

\noindent\textbf{Split-quality summary.}
Table~\ref{tab:alpha_split_quality} summarizes the per-cut split quality on COCO val: the partner-set NPMI separation ($\mu^{\text{typ}} - \mu^{\text{aty}}$), the post-balancing typical-matched / atypical image counts, and the number of anchor classes that survive both the implementation floors and the balancing step. Three observations support the quintile choice. First, the quintile cut achieves the highest anchor coverage of any cut ($64/80$, vs.\ $56$ for $\alpha=0.5$ and $61$ for the decile), which is the relevant quantity for stable per-anchor AP estimates.
Second, its pooled NPMI means lie on opposite sides of zero ($+0.21$ vs.\ $-0.23$, a gap of $0.44$), while the wider cuts admit partners closer to zero (gap $0.24$ at $\alpha = 0.5$). Third, it keeps about $1.5\times$ the images of the decile ($465 / 389$ vs.\ $300 / 259$); the decile separates the partner sets more sharply (gap $0.55$) but keeps fewer anchors and images. The quintile cut therefore balances anchor coverage, NPMI separation, and sample size.

\begin{table}[h]
\centering
\small
\caption{\textbf{Split-quality summary across $\alpha$ cuts on COCO val.} The quintile cut keeps the most anchors ($64/80$), about $1.5\times$ the images of the decile, and pooled NPMI means of opposite sign. The atypical side is preserved as-is; the typical side is down-sampled per-anchor to approximately match the atypical-side image count (see ``Image balancing''). NPMI means are pooled over all (anchor, partner) pairs of the $80$ anchors.}
\label{tab:alpha_split_quality}
\setlength{\tabcolsep}{4pt}
\begin{tabular}{l c cc cc c}
\toprule
& & \multicolumn{2}{c}{\textbf{NPMI ($\mu$)}} & \multicolumn{2}{c}{\textbf{Images (post-bal.)}} & \\
\cmidrule(lr){3-4} \cmidrule(lr){5-6}
Cut & $\alpha$ & typ. & atyp. & typ. & atyp. & Anchors \\
\midrule
$50/50$  & $0.5$ & $+0.09$ & $-0.15$ & 491 & 419 & $56/80$ \\
$60/40$  & $0.4$ & $+0.12$ & $-0.17$ & 614 & 536 & $63/80$ \\
$70/30$  & $0.3$ & $+0.16$ & $-0.20$ & 554 & 490 & $63/80$ \\
\textbf{quintile $80/20$} & $\boldsymbol{0.2}$ & $\boldsymbol{+0.21}$ & $\boldsymbol{-0.23}$ & $\boldsymbol{465}$ & $\boldsymbol{389}$ & $\boldsymbol{64/80}$ \\
decile $90/10$    & $0.1$ & $+0.29$ & $-0.26$ & 300 & 259 & $61/80$ \\
\bottomrule
\end{tabular}
\end{table}

\begin{figure}[t]
\centering
\includegraphics[width=\linewidth]{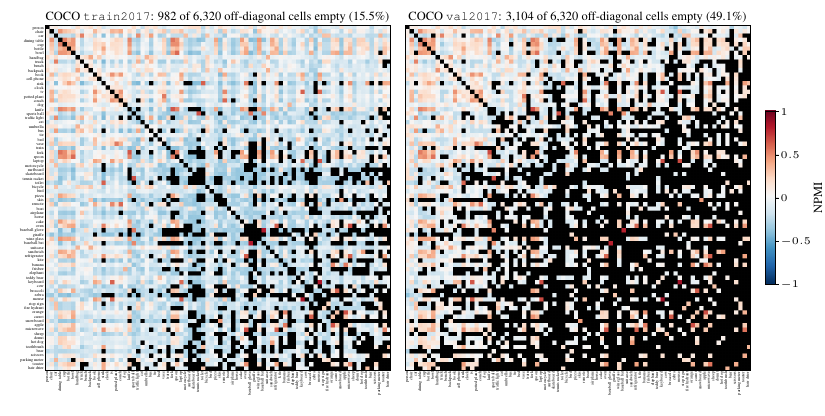}
\caption{\textbf{NPMI matrix on COCO train vs.\ COCO val (80$\times$80 categories, same train-frequency ordering, black = never co-occur; the diagonal is black by construction).} \emph{Left:} on \texttt{train2017}, $982$ of the $80 \cdot 79 = 6{,}320$ off-diagonal cells are zero ($15.5\%$): a small minority of category pairs never co-occur in training. \emph{Right:} on \texttt{val2017}, the empty-cell fraction jumps to $49.1\%$ ($3{,}104$ zero cells), more than $3\times$ the training fraction. The added zeros are structured: the empty-cell rate is highest in the lower-right long-tail $\times$ long-tail quadrant ($78.5\%$ vs.\ $39.5\%$ elsewhere), exactly the regime where typicality bias is expected to be strongest. For nearly half of all category pairs the natural val set provides \emph{no} signal at all, irrespective of how we slice it; this is the structural coverage hole that motivates the synthetic STRANGE-Bench evaluation (\S\ref{app:strange-pipeline}).}
\label{fig:npmi_train_vs_val}
\end{figure}

\noindent\textbf{Robustness of the gap across cuts.}
We additionally verify that the typicality gap is not an artifact of the quintile cut by sweeping $\alpha$ across the augmentation suite of Sec.~\ref{sec:experiments} on Co-DINO R-50 (12 epochs at the base LR and one fine-tuning epoch at $0.1\times$ LR). Tab.~\ref{tab:alpha_sweep_codino} reports the per-cut gap $\Delta^{\text{T-A}} = M^{\text{T}} - M^{\text{A}}$ on COCO val paired-real (full-annotation mAP, $\%$) for eight configurations covering no augmentation, standard augmentation, and six of the seven non-AAS augmentation baselines of the main paper (all but XPaste). Two patterns emerge. First, a non-trivial typicality gap is present under \emph{every} configuration at \emph{every} cut: the per-configuration mean across cuts $\overline{\Delta}$ ranges from $4.5$ (Mosaic, which lowers AP on both splits; see \S\ref{app:codino_full}) to $6.5$ (Std.). The gap is not a quirk of the quintile cut. Second, standard augmentation does \emph{not} reduce $\Delta^{\text{T-A}}$ relative to no-aug ($\overline{\Delta}_{\text{Std.}} = 6.5$ vs.\ $\overline{\Delta}_{\text{No aug}} = 6.0$), and at the quintile cut itself it slightly \emph{widens} the gap ($9.2$ vs.\ $8.7$). This is the diagnostic signature that motivates Sec.~\ref{sec:aas}: a typicality intervention has to be designed for the bias rather than inherited as a side effect of conventional augmentation.

\begin{table}[h]
\centering
\small
\caption{\textbf{$\alpha$-sweep on Co-DINO R-50 on COCO val
paired-real, full-annotation mAP gap (\%).} Per-cut typicality gap
$\Delta^{\text{T-A}}$ across five $\alpha$ cuts and the per-row mean
$\overline{\Delta}$ for each augmentation. The gap is non-zero
under every configuration at every cut; standard augmentation does
not reduce the gap relative to no-aug. Co-DINO R-50 trained for 12 epochs at the base LR and one fine-tuning epoch at $0.1\times$ LR. The SCP row is the submission's SCP run, not the SCP row of Tab.~\ref{tab:codino_main}; re-evaluating these checkpoints changes the quintile and decile gaps by at most 0.2.}
\label{tab:alpha_sweep_codino}
\setlength{\tabcolsep}{4pt}
\begin{tabular}{l ccccc c}
\toprule
Aug & $50/50$ & $60/40$ & $70/30$ & \textbf{quintile} & decile & $\overline{\Delta}$ \\
\midrule
No aug & 6.7 & 3.6 & 4.8 & 8.7 & 6.1 & 6.0 \\
Std. & 9.2 & 5.1 & 3.9 & 9.2 & 4.9 & 6.5 \\
Color jitter & 8.1 & 4.7 & 5.0 & 9.0 & 5.0 & 6.4 \\
MixUp & 7.6 & 5.0 & 4.2 & 8.0 & 5.9 & 6.1 \\
Mosaic & 6.5 & 4.0 & 3.4 & 6.4 & 2.4 & 4.5 \\
YOLOX & 5.7 & 3.7 & 3.7 & 6.8 & 3.6 & 4.7 \\
SCP & 6.7 & 4.7 & 3.7 & 8.7 & 4.7 & 5.7 \\
InstAug & 7.7 & 5.0 & 4.0 & 8.2 & 4.2 & 5.8 \\
\bottomrule
\end{tabular}

\end{table}

\FloatBarrier
\noindent\textbf{Why is natural data not enough?}
Even at the standard cuts we use, the COCO paired-real splits comprise only $465 / 389$ typical-matched / atypical images at the quintile cut ($\alpha = 0.2$) and $300 / 259$ at the decile ($\alpha = 0.1$) (Tab.~\ref{tab:alpha_split_quality} and Tab.~\ref{tab:app-eval-datasets}), spread across $64$ and $61$ anchor classes. Per-anchor coverage is therefore modest in absolute terms and very uneven across the long tail; anchors whose atypical partners never appear with them in val are dropped ($16$ of $80$ at the quintile cut), and $28$ of the remaining $64$ anchors have fewer than three atypical images. More fundamentally, no natural cut can reach the regime where partners \emph{never} co-occur with the anchor. As Fig.~\ref{fig:npmi_train_vs_val} shows, on \texttt{train2017} the never-co-occur cells form a small minority of the $80 \times 80$ NPMI matrix ($982$ of the $6{,}320$ off-diagonal cells, $15.5\%$), but on \texttt{val2017} the empty fraction jumps to $49.1\%$: nearly half of all category pairs that we might want to probe simply have no images in val. The empty-cell rate is highest in the long-tail $\times$ long-tail quadrant ($78.5\%$ vs.\ $39.5\%$ elsewhere), exactly where typicality bias should be most pronounced and stress-tests are most informative. By construction, no quantile cut on val can recover these cells, and the never-co-occur regime ($T_{\text{never}}$) is inaccessible from natural data altogether. This is the gap STRANGE-Bench (\S\ref{app:strange-pipeline}) is designed to close: rendering scenes from sampled category sets lets us produce paired typical / atypical scenes \emph{and} a clean $T_{\text{never}}$ tier with balanced per-anchor coverage that natural validation sets structurally cannot offer.

\clearpage

\section{STRANGE-Bench: Pipeline Details and Statistics}
\label{app:strange-pipeline}

STRANGE-Bench is our diffusion-generated tiered evaluation set: $5{,}523$ photorealistic scenes with $17{,}332$ instance annotations across $17$ NPMI-stratified sub-tiers, spanning typical, atypical, and never-co-occurring COCO object pairs. The remainder of this section describes the dataset release (\S\ref{app:strange:release}), shows qualitative samples (\S\ref{app:strange:samples}), reports the per-tier composition and sanity diagnostics (\S\ref{app:strange:tiers}--\S\ref{app:strange:diag}), describes the five-stage generation pipeline (\S\ref{app:strange:pipeline}), and lists the implementation hyperparameters
(\S\ref{app:strange:impl}).

\subsection{Dataset Release}
\label{app:strange:release}

\noindent\textbf{Hosting.}
The full dataset and its metadata are publicly hosted on Hugging Face, and the generation code on GitHub (Tab.~\ref{tab:strange_links}).\footnote{The dataset repository is public and requires no login.} A stratified inspection sample drawn with a fixed seed is shipped under \texttt{samples/} so readers can assess data quality without downloading the full release. 

\begin{table}[h]
\centering
\tiny
\setlength{\tabcolsep}{6pt}
\caption{\textbf{STRANGE-Bench artifact URLs}.}
\label{tab:strange_links}
\begin{tabular}{@{}ll@{}}
\toprule
Artifact & URL \\
\midrule
Dataset    & \url{https://huggingface.co/datasets/sidgairo18/strange-bench} \\
Code       & \url{https://github.com/sidgairo18/stranger-things} \\
Samples    & \url{https://huggingface.co/datasets/sidgairo18/strange-bench/tree/main/samples} \\
Croissant  & \url{https://huggingface.co/datasets/sidgairo18/strange-bench/blob/main/croissant.json} \\
\bottomrule
\end{tabular}
\end{table}

\noindent\textbf{Variants.}
We release two variants: the \emph{balanced} variant ($5{,}523$ images / $17{,}332$ instance annotations) is the configuration whose numbers we report throughout this paper; the pre-balancing \emph{clean} variant ($6{,}571$ / $21{,}349$) is included for transparency. We also release the IDs of the $3{,}234$ scenes in which no class appears in the generated training scenes of the AAS late stage (the held-out-class subset of App.~\ref{app:circularity}).

\noindent\textbf{Format.}
Images are $1024 \times 1024$ PNG; annotations follow the COCO instance format (bounding boxes + per-class segmentation masks).
A Croissant~1.1 file with both core and Responsible-AI~1.0 fields is shipped at the repository root.

\noindent\textbf{License.}
Data: \textbf{CC-BY 4.0}. Generation code: \textbf{MIT}. The dataset is intended for evaluation only; we ask users not to train on it.

\noindent\textbf{Reproducibility.}
The full pipeline, from the NPMI matrix through layout sampling, rendering, segmentation and the two audits to tier balancing, is reproducible from the included code; required external assets are public model weights and the COCO~\texttt{train2017} annotations.

\noindent\textbf{Generality and Extensibility.}
The pipeline above is deliberately recipe-driven and not COCO-specific. Swapping the input \emph{label vocabulary} (here, the $80$ COCO classes) and the source \emph{co-occurrence statistics} (here, NPMI on COCO~\texttt{train2017}) is sufficient to generate a STRANGE-Bench-style probe for any other dataset or task family. We release dataset, code, samples, and the Croissant manifest (Tab.~\ref{tab:strange_links}) under a permissive license so the community can:
\begin{itemize}

 \item regenerate STRANGE-Bench at higher resolution or larger scale, 
 \item target alternative vocabularies (LVIS~\cite{gupta2019lvis}, Objects365~\cite{shao2019objects365}, fine-grained categories, video frames), 
 \item substitute newer generative models (e.g., higher-resolution SD variants~\cite{stabilityai2024sd35}, Imagen~\cite{saharia2022photorealistic}, FLUX~\cite{flux2024}) or stronger auditors as they appear, and 
 \item compose with new typicality axes (action-object, scene-object, temporal context).
\end{itemize}
The pipeline's modular form (layout, render, segment, audit) means each stage can be replaced independently without rebuilding the rest. Our intent is to provide both a benchmark and a reusable construction recipe for stress-testing co-occurrence shortcuts in vision systems. 

\subsection{Qualitative Samples}
\label{app:strange:samples}

\begin{figure}[h!]
\centering
\includegraphics[width=\linewidth]{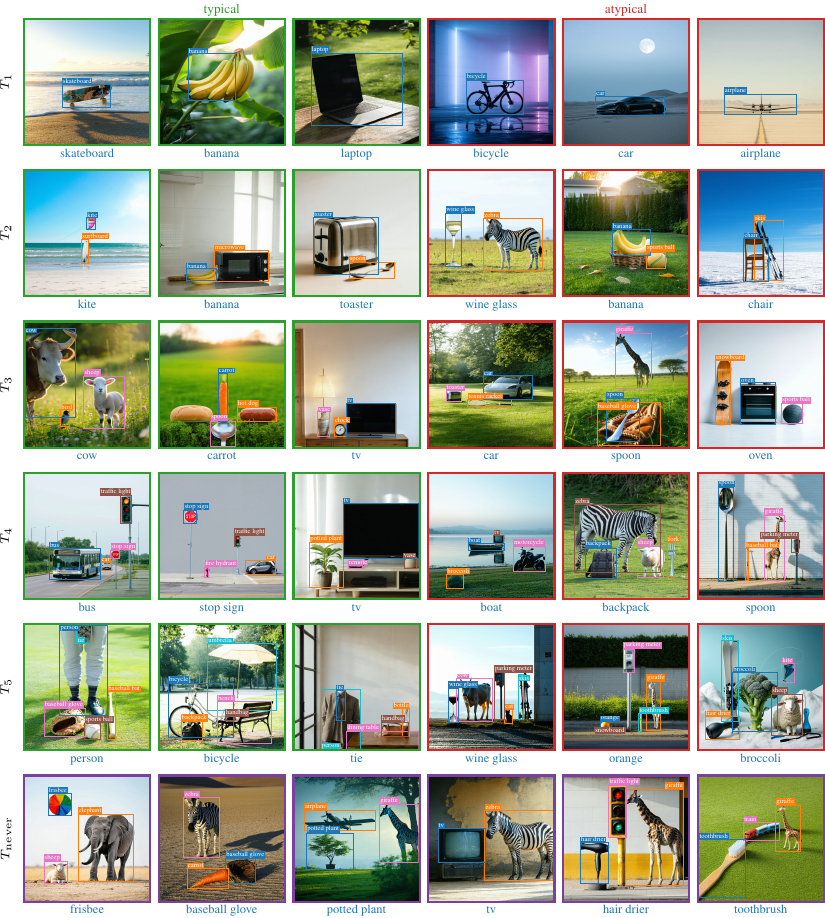}
\caption{\textbf{Random samples across the main tiers.} Three typical (green) and three atypical (red) scenes per tier, for $T_1$ with natural and unusual background, and six $T_\text{never}$ scenes (purple), drawn uniformly at random and not curated; the anchor class is below each panel. Across panels, partner-count $k\!\in\!\{0,1,2,3,4\}$ and partner typicality (typical / atypical / never) jointly drive the contextual variation.}
\label{fig:samples_random_1}
\end{figure}

\begin{figure}[h!]
\centering
\includegraphics{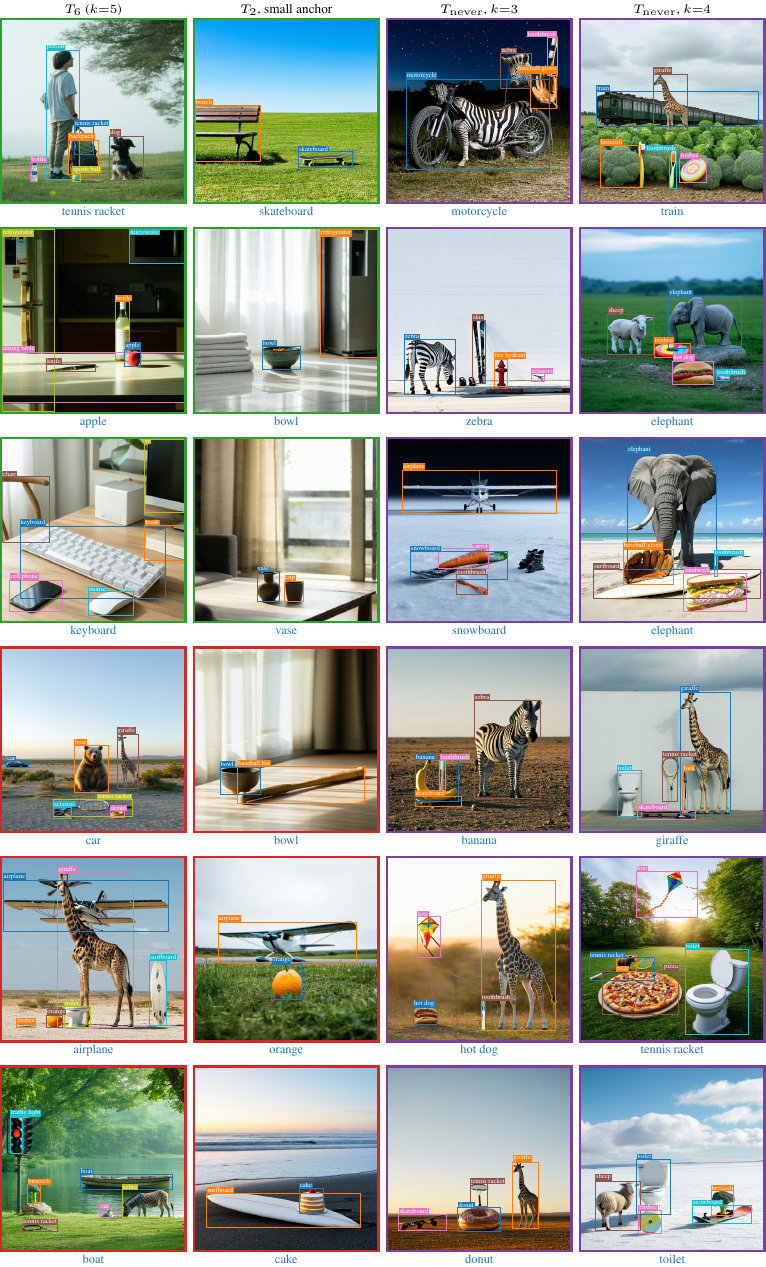}
\caption{\textbf{Random samples of the six sub-tiers outside the main grid.} One column per sub-tier, drawn at random as in Fig.~\ref{fig:samples_random_1}: $T_6$ ($k = 5$) and the $k = 1$ pair with a small anchor (three typical scenes above three atypical ones), and the never-co-occurring cliques with $k = 3$ and $k = 4$ (App.~\ref{app:additional_results}).}
\label{fig:samples_random_2}
\end{figure}

\begin{figure}[h!]
\centering
\includegraphics{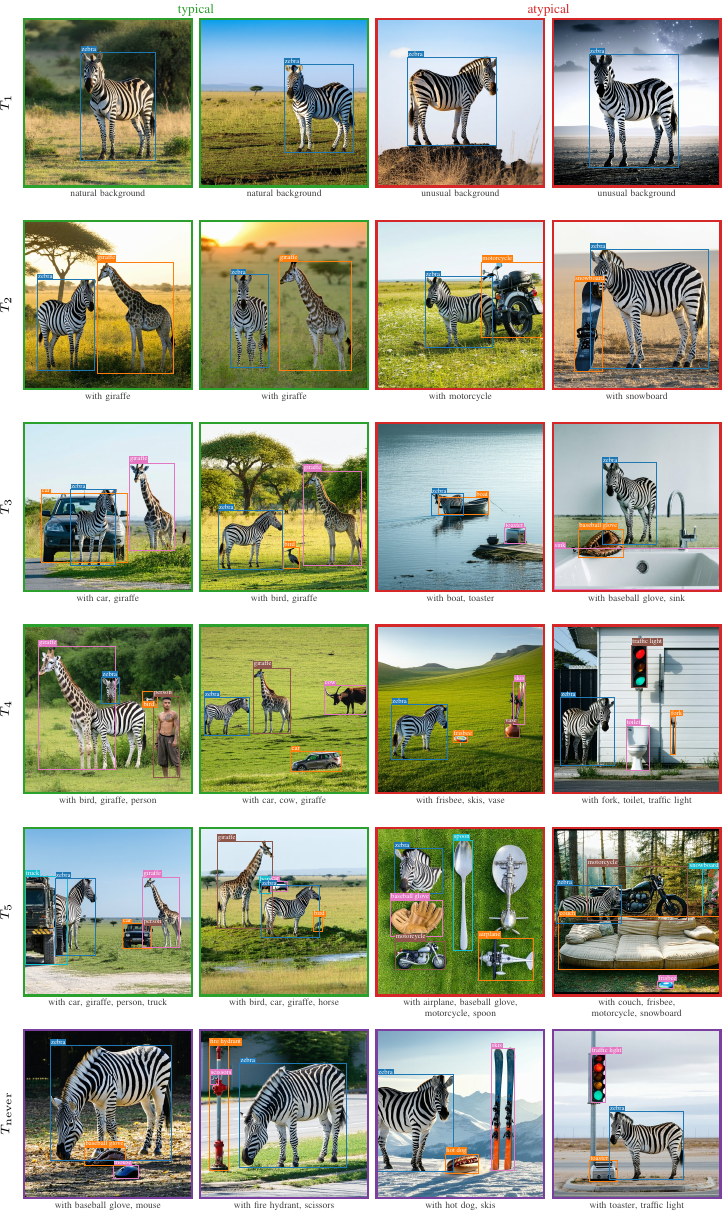}
\caption{\textbf{Anchor: \emph{zebra}.} Two typical (green) and two atypical (red) scenes per tier and four $T_\text{never}$ scenes (purple) with this anchor, drawn at random; the partner classes are listed below each panel. Top to bottom: the anchor alone with a natural or unusual background ($T_1$); $T_2$ to $T_5$ add $1$ to $4$ typical or atypical partners; $T_\text{never}$ adds one or two partners that never co-occur with the anchor. The zebra stays visually consistent; the surrounding scene shifts as we sweep partner-count and pair-typicality.}
\label{fig:samples_class_zebra} 
\end{figure}

\begin{figure}[h!]
\centering
\includegraphics{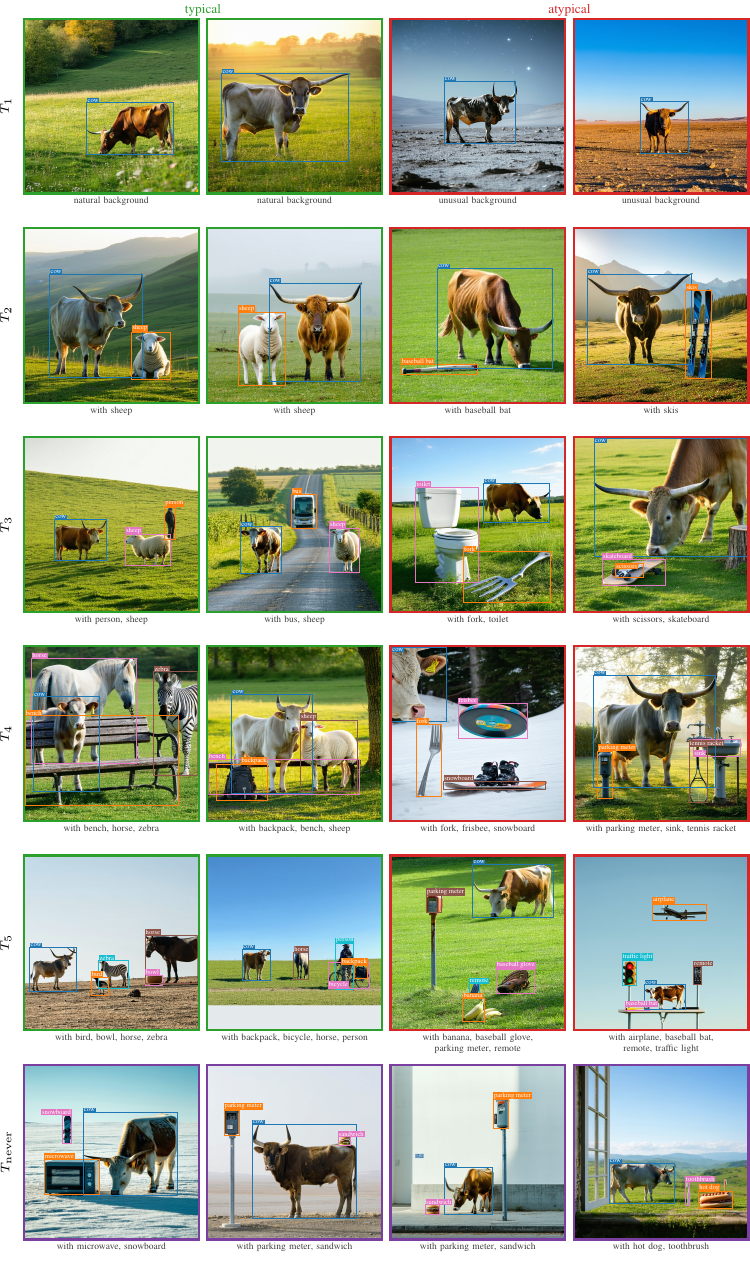}
\caption{\textbf{Anchor: \emph{cow}.} Same layout as
Fig.~\ref{fig:samples_class_zebra}. Notice how the \emph{cow} appears in pasture / farmland contexts on the typical side, while on the atypical / never side it shares the scene with out-of-place objects (toilet, snowboard, microwave) or, at $T_1$, an unusual background.}
\label{fig:samples_class_cow}
\end{figure}

\begin{figure}[h!]
\centering
\includegraphics{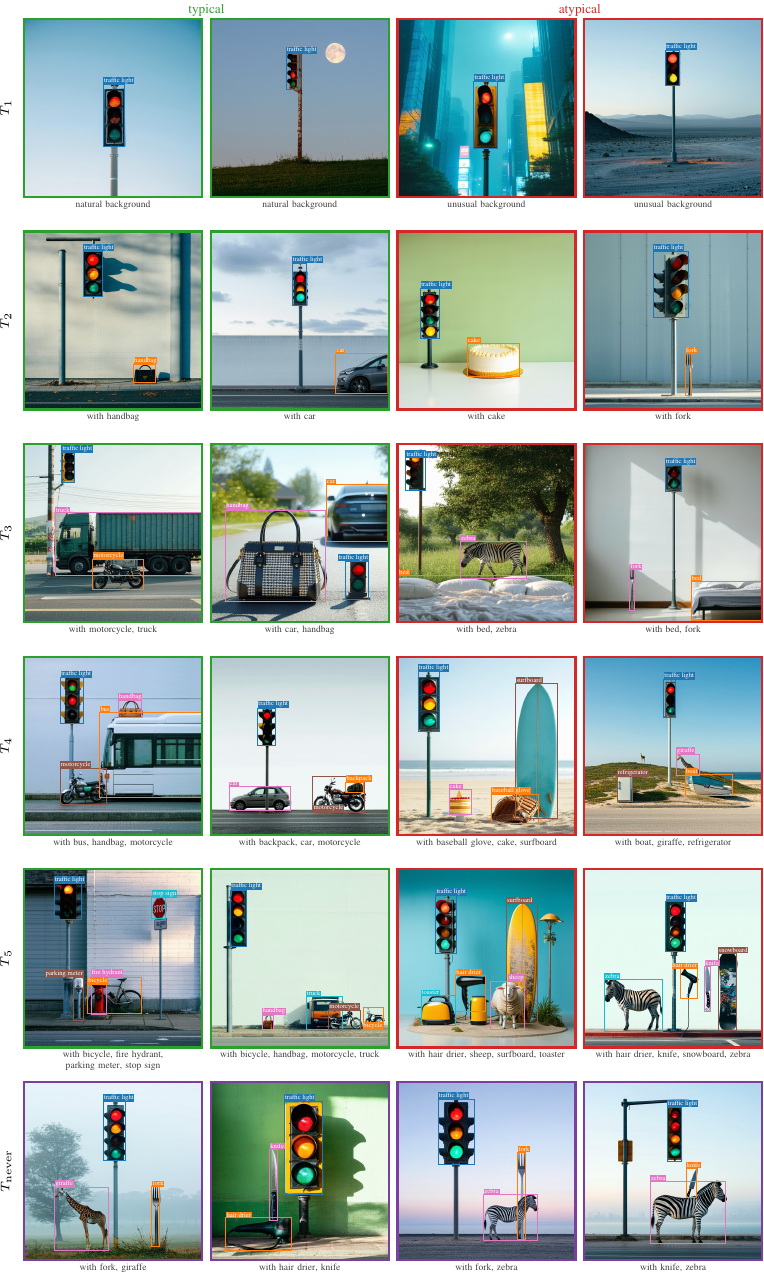}
\caption{\textbf{Anchor: \emph{traffic light}.} Same layout as Fig.~\ref{fig:samples_class_zebra}. Typical contexts pair it with street scenes and vehicles; atypical / never contexts displace the traffic light into indoor or natural scenes (bedroom, beach, grassland).}
\label{fig:samples_class_traffic_light}
\end{figure}

\begin{figure}[h!]
\centering
\includegraphics{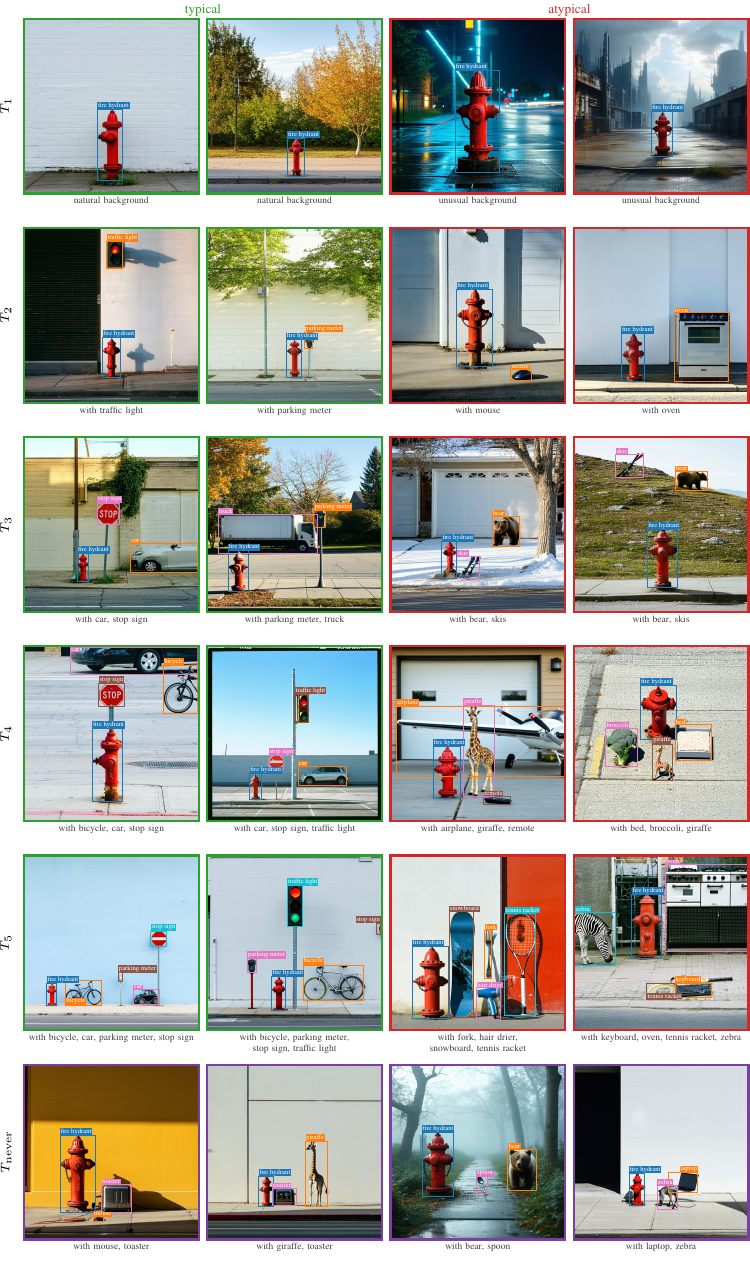}
\caption{\textbf{Anchor: \emph{fire hydrant}.} Same layout as Fig.~\ref{fig:samples_class_zebra}. Typical pairings include vehicles and other street objects (car, stop sign, parking meter); atypical / never pairings place the fire hydrant next to animals and household objects (bear, giraffe, oven, toaster).}
\label{fig:samples_class_fire_hydrant}
\end{figure}

\begin{figure}[h!]
\centering
\includegraphics{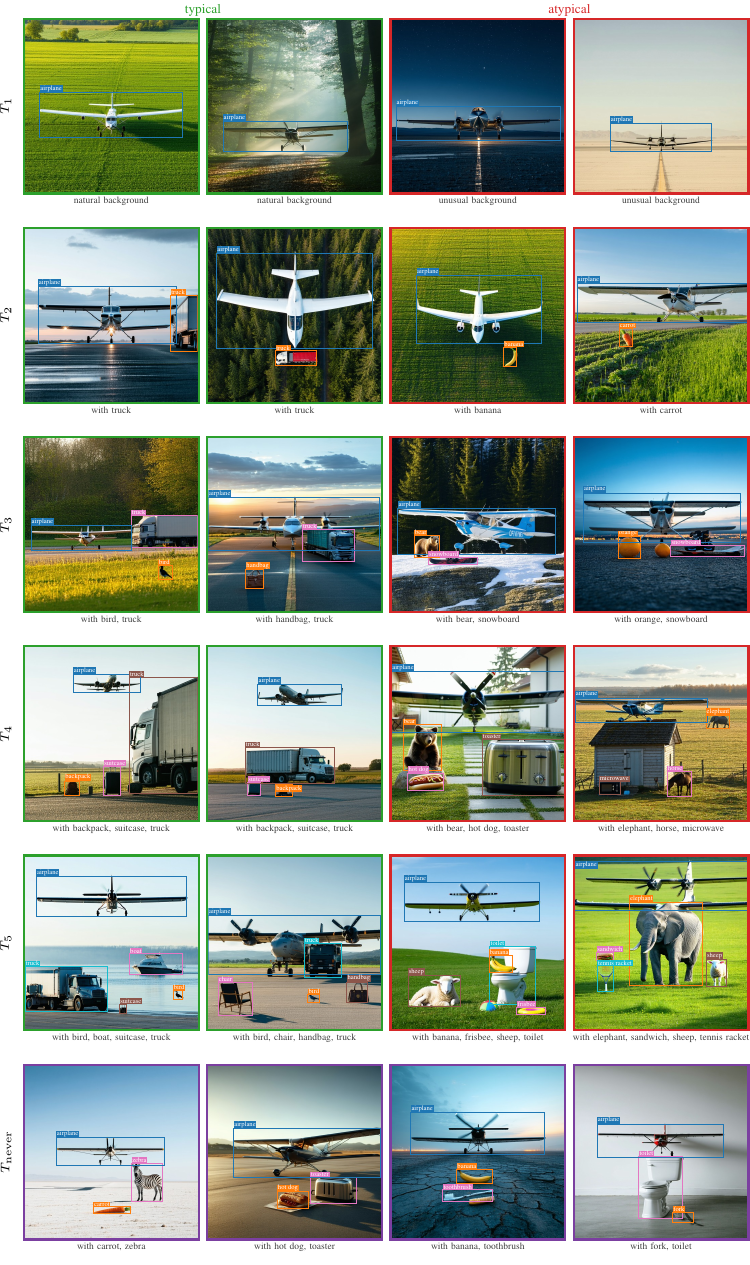}
\caption{\textbf{Anchor: \emph{airplane}.} Same layout as Fig.~\ref{fig:samples_class_zebra}. Typical pairings usually appear on runways or airfields with related vehicles and luggage (truck, suitcase); atypical / never pairings put the airplane next to food, animals, or household objects (banana, elephant, toilet).}
\label{fig:samples_class_airplane}
\end{figure}

\begin{figure}[h!]
\centering
\includegraphics{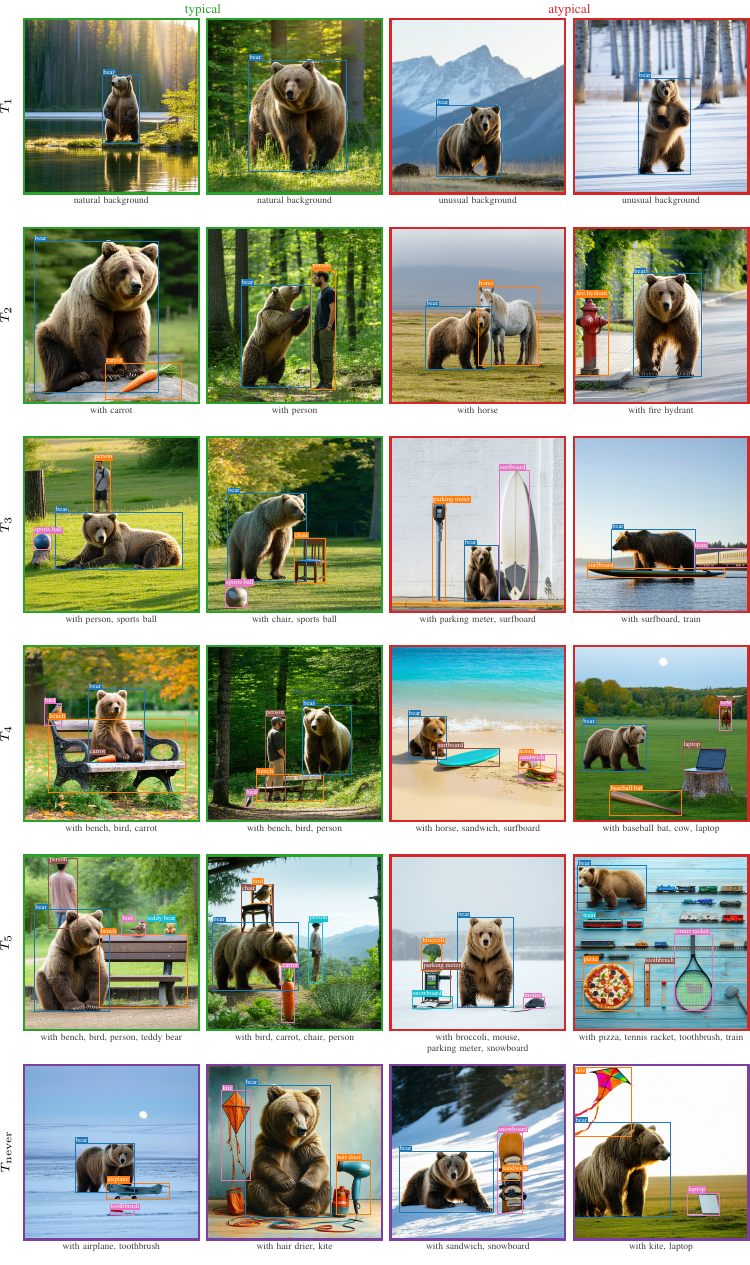}
\caption{\textbf{Anchor: \emph{bear}.} Same layout as
Fig.~\ref{fig:samples_class_zebra}. Typical pairings are forest / park scenes; atypical / never pairings move the bear to beaches, snow, or studio backdrops next to household objects (laptop, hair drier,~toothbrush).}
\label{fig:samples_class_bear}
\end{figure}

We first show what the dataset looks like. Each panel of every figure below is a single STRANGE-Bench scene (post-audit), with its ground-truth boxes overlaid (anchor in blue). Figs.~\ref{fig:samples_random_1}--\ref{fig:samples_random_2} show uncurated random samples of the eleven sub-tiers of the main grid and of the six others. Figs.~\ref{fig:samples_class_zebra}--\ref{fig:samples_class_bear} fix one anchor class and arrange two typical and two atypical scenes per tier (top to bottom: $T_1 \to T_5 \to T_\text{never}$), so the anchor is held visually constant while its surrounding context sweeps from typical to never-seen. 

\clearpage
\subsection{Sub-Tier Composition}
\label{app:strange:tiers}

\paragraph{Two axes.}
STRANGE-Bench varies two axes: \emph{partner-count} $k\!\in\!\{0,\dots,5\}$ and \emph{pair-typicality} (typical: high-NPMI partners; atypical: low-NPMI or never-co-occurring partners; never-co-occurring: $N_{ij}\!=\!0$ for every pair of the scene, with the fallback below). The anchor-only setting ($k\!=\!0$) splits into a natural-background and an atypical-background variant, giving two sub-tiers at $k\!=\!0$, two each at $k\!=\!1,\dots,4$, and one \texttt{tier\_never} sub-tier for a total of eleven, the grid of the main tables. Six more sub-tiers appear only in App.~\ref{app:additional_results}: the typical and atypical pair at $k\!=\!5$ ($T_6$), a $k\!=\!1$ pair whose anchor box covers at most $4\%$ of the image, and never-co-occurring cliques with $k\!=\!3$ and $k\!=\!4$.

\paragraph{Anchor sampling and partner pools.}
Anchors are allocated evenly over the $80$ COCO categories ($74$ to $80$ anchors per sub-tier of the main grid after filtering). Partners are added one at a time, and each new partner must meet the arm's rule with the anchor and with every partner already chosen. The rule uses $A_{ij} = (1 - \text{NPMI}_{ij})/2$ on COCO~\texttt{train2017}: typical pairs need $A_{ij} \leq 0.45$ (NPMI $\geq 0.1$) and at least $10$ co-occurrences with the anchor; atypical pairs need $A_{ij} \geq 0.55$ (NPMI $\leq -0.1$). Pairs that never co-occur get $A_{ij} = 0.8$, above every co-occurring pair. Each partner is drawn by a softmax over the $15$ highest-scoring candidates; if no candidate meets the rule, the sampler takes the highest-scoring one. 

\paragraph{The \texttt{tier\_never} construction.}
For \texttt{tier\_never} we attempt a strict $3$-clique (every pair among $\{\text{anchor}, p_1, p_2\}$ has $N_{ij}\!=\!0$), or a pair when the anchor has a single never-co-occurring partner; for ubiquitous anchors that admit no $3$-clique (\emph{person}, \emph{car}, \emph{chair}, \dots), we pad the largest strict sub-clique with atypical partners. Of the $466$ \texttt{tier\_never} scenes ($397$ with two partners, $69$ with one), $83.7\%$ are strict cliques; the rest are fallback scenes whose strict-clique status can be derived from \texttt{labels\_intended} together with the COCO-train co-occurrence matrix. The cliques with $k\!=\!3$ and $k\!=\!4$ are strict by construction and use only pairs whose expected count under independence, $N p_i p_j$, is at least $10$, so that a zero count reflects absence rather than two rare classes; anchors without such a clique are skipped ($46$ and $28$ anchors remain). Per-tier composition is in Table~\ref{tab:strange_tiers}.

\begin{table}[h!]
\centering\footnotesize
\caption{\textbf{The 17 STRANGE-Bench sub-tiers.} The main tables use the eleven sub-tiers above the middle rule; App.~\ref{app:additional_results} reports the six below it. Med $|$ Mean NPMI is over the anchor-partner pairs that co-occur in COCO~\texttt{train2017}; $N_{ij}{=}0$ is the share of anchor-partner pairs that never co-occur there. Anns/Img leaves out the $300$ ignore regions.}
\label{tab:strange_tiers}
\resizebox{\linewidth}{!}{%
\begin{tabular}{lclrrcr}
\toprule
Sub-tier & $k$ & Partner type & Scenes & Anns/Img & Med $|$ Mean NPMI & $N_{ij}{=}0$ (\%) \\
\midrule
\texttt{tier1\_natural\_bg} & 0 & none, natural background & $421$ & $1.00$ & -- & -- \\
\texttt{tier1\_atypical\_bg} & 0 & none, unusual background & $421$ & $1.00$ & -- & -- \\
\texttt{tier2\_1typical} & 1 & typical (high NPMI) & $400$ & $2.00$ & $+0.20$ $|$ $+0.24$ & 0.0 \\
\texttt{tier2\_1atypical} & 1 & atypical (low NPMI) & $400$ & $2.00$ & $-0.32$ $|$ $-0.31$ & 57.5 \\
\texttt{tier3\_2typical} & 2 & typical & $392$ & $3.00$ & $+0.18$ $|$ $+0.21$ & 0.0 \\
\texttt{tier3\_2atypical} & 2 & atypical & $392$ & $3.00$ & $-0.29$ $|$ $-0.29$ & 53.1 \\
\texttt{tier4\_3typical} & 3 & typical & $399$ & $4.00$ & $+0.18$ $|$ $+0.20$ & 0.0 \\
\texttt{tier4\_3atypical} & 3 & atypical & $399$ & $4.00$ & $-0.29$ $|$ $-0.28$ & 45.9 \\
\texttt{tier5\_4typical} & 4 & typical & $229$ & $5.00$ & $+0.17$ $|$ $+0.19$ & 0.0 \\
\texttt{tier5\_4atypical} & 4 & atypical & $229$ & $5.00$ & $-0.28$ $|$ $-0.27$ & 43.1 \\
\texttt{tier\_never} & 1--2 & never co-occurring & $466$ & $2.85$ & -- & 82.4 \\
\midrule
\texttt{tier2\_1typical\_\_small\_anchor} & 1 & typical, small anchor & $200$ & $2.00$ & $+0.19$ $|$ $+0.23$ & 0.0 \\
\texttt{tier2\_1atypical\_\_small\_anchor} & 1 & atypical, small anchor & $200$ & $2.00$ & $-0.30$ $|$ $-0.30$ & 64.0 \\
\texttt{tier6\_5typical} & 5 & typical & $193$ & $6.00$ & $+0.16$ $|$ $+0.18$ & 0.0 \\
\texttt{tier6\_5atypical} & 5 & atypical & $193$ & $6.00$ & $-0.27$ $|$ $-0.27$ & 39.2 \\
\texttt{tier\_never\_k3} & 3 & never co-occurring (clique) & $334$ & $4.00$ & -- & 100.0 \\
\texttt{tier\_never\_k4} & 4 & never co-occurring (clique) & $255$ & $5.00$ & -- & 100.0 \\
\midrule
\textbf{Total} & & & $\mathbf{5{,}523}$ & $\mathbf{3.14}$ & & \\
\bottomrule
\end{tabular}

}
\end{table}

\subsection{Sanity Diagnostics}
\label{app:strange:diag}

\begin{figure}[h!]
\centering
\includegraphics[width=\linewidth]{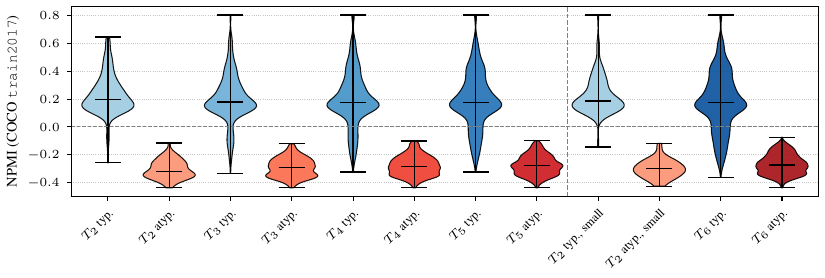}
\caption{\textbf{Tier semantics.} Per-sub-tier distribution of intended-pair NPMI on COCO-train. Typical and atypical sub-tiers have the expected sign and are separated at every quartile. Pairs that never co-occur have no NPMI and are not shown: the never-co-occurring sub-tiers, and in the atypical sub-tiers the share in the last column of Tab.~\ref{tab:strange_tiers}. Sub-tiers right of the dashed line are outside the main grid.}
\label{fig:npmi_per_tier}
\end{figure}

\begin{figure}[h!]
\centering
\includegraphics[width=\linewidth]{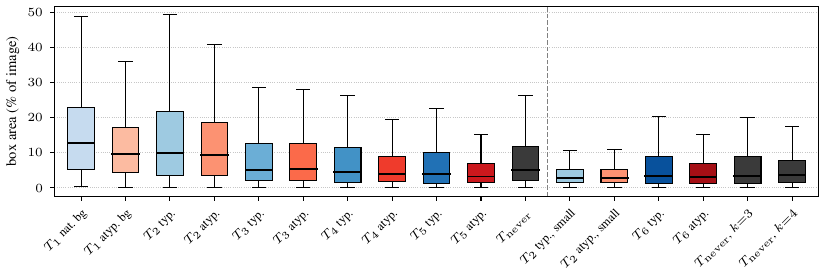}
\caption{\textbf{Object-scale balance.} Per-sub-tier boxplot of relative bbox area (\% of image). Median absolute difference between the typical and atypical sub-tiers at fixed $k$: $0.66$, $0.37$, $0.47$, $0.65$, $0.15$\,pt for $k\!=\!1,\dots,5$ ($T_2$ to $T_6$), $0.09$\,pt for the small-anchor pair, and $3.0$\,pt for $T_1$. Sub-tiers right of the dashed line are outside the main grid.}
\label{fig:bbox_area_per_tier}
\end{figure}

\begin{figure}[h!]
\centering
\includegraphics[width=\linewidth]{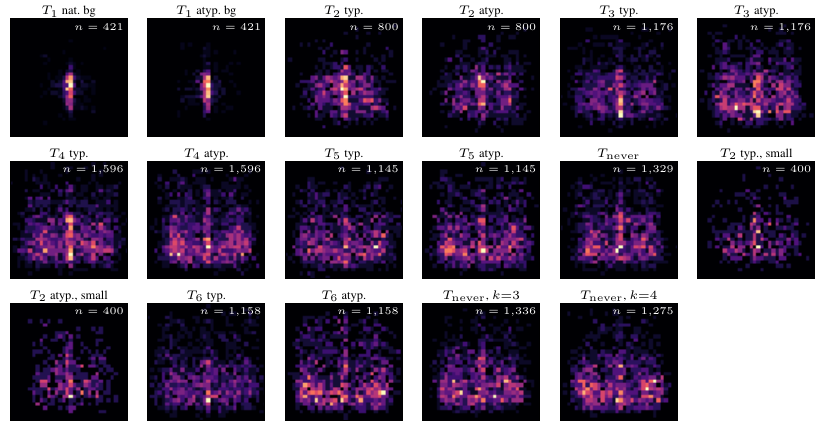}
\caption{\textbf{Spatial layout.} Box-center density
per sub-tier on a $32\!\times\!32$ grid (normalized to image
coordinates). Centers lie below the image center on average in every sub-tier, with the same distribution in the typical and atypical sub-tiers at each $k$; the single anchors of $T_1$ sit near the horizontal center.}
\label{fig:spatial_centers}
\end{figure}

Three checks test the typicality structure and two nuisance variables, object scale and position. \textit{(i) Tier semantics.} Per-tier intended-pair NPMI on COCO-train separates the typical and atypical sub-tiers at every quartile: the lower quartile of every typical sub-tier is at least $+0.05$ and the upper quartile of every atypical sub-tier at most $-0.20$ (typical means $+0.18$ to $+0.24$, atypical $-0.27$ to $-0.31$; pairs that never co-occur have no NPMI and are left out; Fig.~\ref{fig:npmi_per_tier}). \textit{(ii) Object-scale balance.} Median relative bbox area within typical/atypical pairs at fixed $k$ is matched to within $0.15$--$0.66$\,pt for $k\!=\!1,\dots,5$ (Fig.~\ref{fig:bbox_area_per_tier}); the largest gap is at $k\!=\!1$ ($0.66$\,pt), and the single anchor of $T_1$ is $3.0$\,pt larger on the natural than on the atypical background. \textit{(iii) Spatial layout.} Box centers sit below the image center on average in every sub-tier (mean vertical position $0.60$ to $0.67$ of the image height, $0$ at the top), and the typical and atypical sub-tiers at the same $k$ match within $0.013$ in mean position (Fig.~\ref{fig:spatial_centers}), so the two arms share the same center bias.

\FloatBarrier
\subsection{Generation Pipeline}
\label{app:strange:pipeline}

For each scene we sample $k{+}1$ COCO categories from a target sub-tier, ask \textit{Qwen2.5-3B-Instruct}~\citep{qwen2024qwen25} for a layout (JSON bounding boxes, LayoutGPT-style~\citep{feng2023layoutgpt,lian2024lmd,cho2023vpgen}) and a photographic caption from those boxes' positions \citep{betker2023dalle3,hao2023promptist}, render with \textit{Stable Diffusion 3.5 Large}~\citep{stabilityai2024sd35,esser2024scaling}, segment with \textit{SAM3}~\citep{carion2026sam3},
and discard the scene if a \textit{Grounding DINO}~\citep{liu2024grounding} auditor flags extras or duplicates, or if the second audit or the completeness check of Stage 5 fails. Numeric hyperparameters for every stage are deferred to \ref{app:strange:impl}.

\noindent\textbf{Stage 1: layout and caption (Qwen).}
We use Qwen2.5-3B-Instruct twice per scene with the same weights but two fixed system+user templates. The \emph{layout call} (LayoutGPT-style) emits pixel-coordinate bounding boxes; the \emph{caption call} writes the photographic prompt for SD~3.5L from the boxes' positions and per-class visual descriptions. 

\begin{quote}\small
\textbf{Layout system.} \emph{You are a layout designer. Output
ONLY a JSON array of bounding boxes.}

\medskip
\textbf{Layout user.} \emph{Generate a realistic bounding box layout for an image of size $W\!\times\!H$. Objects to place:
\texttt{\{labels\}}. Return ONLY a JSON list where each item has
``label'' and ``bbox'' $[x_1, y_1, x_2, y_2]$ in pixels. Ensure
boxes don't overlap excessively and are realistically sized and
positioned. Each box must be inside the image
($0 \leq x_1 < x_2 \leq W$, $0 \leq y_1 < y_2 \leq H$). Each box
should be at least $64$ pixels wide and tall. Return labels exactly
as given. Example: $[\{$``label'': ``cat'', ``bbox'': $[50, 200,
250, 450]\}, \dots]$.}
\end{quote}

The emitted boxes are the \emph{planner layout boxes}; they reach the renderer only through the caption; on persistent JSON failure we fall back to a rule-based layout solver that places the boxes from COCO size priors without overlap. $T_6$ and the never-co-occurring cliques use the solver only; $79.2\%$ of the scenes have Qwen layouts. For the caption call we
discretize each box center $(c_x, c_y)/W \in [0,1]^2$ at
$\{0.33, 0.67\}$ to one of nine position strings
(Table~\ref{tab:position_strings}).

\begin{table}[h!]
\centering\small
\caption{\textbf{Position strings emitted by the center-bin discretization}, indexed by column ($c_x$) and row ($c_y$).}
\label{tab:position_strings}
\begin{tabular}{lccc}
\toprule
 & \textbf{left} ($c_x{<}0.33$) & \textbf{center} ($0.33{\leq}c_x{<}0.67$) & \textbf{right} ($c_x{\geq}0.67$) \\
\midrule
\textbf{top} ($c_y{<}0.33$)        & \emph{top left}    & \emph{top center}    & \emph{top right} \\
\textbf{middle} ($0.33{\leq}c_y{<}0.67$) & \emph{left side}   & \emph{center side}   & \emph{right side} \\
\textbf{bottom} ($c_y{\geq}0.67$)  & \emph{bottom left} & \emph{bottom center} & \emph{bottom right} \\
\bottomrule
\end{tabular}
\end{table}

\begin{quote}\small
\textbf{Caption system.} \emph{You are an expert visual-art director writing prompts for a high-end photorealistic text-to-image model. Output ONLY the final prompt as a single sentence -- no preamble, no JSON.}

\medskip
\textbf{Caption user.} \emph{Write a single rich photographic prompt (one sentence, $30$--$70$ words) for a wide-shot photograph (full scene, not close-up) containing the following objects. The image is \texttt{\{context\_label\}}; the objects \texttt{\{tier\_phrase\}}.}

\smallskip
\emph{Objects (label $\rightarrow$ position $\rightarrow$ visual
description):}
\begin{itemize}\setlength\itemsep{0pt}
\item \texttt{\{label\}} $\rightarrow$ \texttt{\{position\}} within the image $\rightarrow$ \texttt{\{class\_visual\_description\}} \item \emph{...one entry per intended class}
\end{itemize}

\emph{The prompt MUST: be a wide / full shot; show every object completely (no cropping); mention each object by its exact label and spatial position; keep a minimalist scene (no extra people or unprompted clutter); specify environment, lighting, time of day, and mood; and use photorealistic-photography style words (wide angle, DSLR, 4K, sharp focus, natural light).}
\end{quote}

The \texttt{\{context\_label\}} and \texttt{\{tier\_phrase\}} fields are tier-conditioned (Table~\ref{tab:tier_phrases}). \texttt{\{class\_\allowbreak visual\_\allowbreak description\}} is a hand-written phrase per COCO class (e.g.\ ``a giraffe with very long neck and patterned coat''); the full $80$-class mapping is given in Table~\ref{tab:class_descriptions}. On total LLM failure we fall back to a deterministic template ``A photorealistic photo of a $\langle$label$_1\rangle$, \dots, and a $\langle$label$_n\rangle$, $\langle$bg-descriptor$\rangle$'' (loses spatial language but preserves the class set). The constraint list was hand-tuned across pilot runs to suppress two SD failure modes -- tight crops (headless figures, partial bodies) and unprompted clutter (extra people in the background).

\begin{table}[h!]
\centering\small
\caption{\textbf{Tier-conditioned scene descriptors.} $^\dagger$Defined in the released generator but not used for the benchmark scenes: its atypical sub-tiers $T_2$ to $T_6$ were captioned with the typical wording (construction note~(ii) below). Variant sub-tiers use the wording of their base sub-tier.}
\label{tab:tier_phrases}
\resizebox{\textwidth}{!}{
\begin{tabular}{lll}
\toprule
Sub-tier(s) & \texttt{\{context\_label\}} & \texttt{\{tier\_phrase\}} \\
\midrule
\texttt{tier1\_natural\_bg}        & set in a natural everyday environment       & fit naturally together \\
\texttt{tier1\_atypical\_bg}       & set in an unusual or unexpected environment & appear together in an unusual setting \\
\texttt{tier\{2--6\}\_*typical}    & in a natural everyday context               & are commonly seen together \\
\texttt{tier\{2--6\}\_*atypical}$^\dagger$   & in an unusual or unexpected context         & are unusually paired \\
\texttt{tier\_never}               & in a highly unusual scene                   & would normally never appear together \\
\bottomrule
\end{tabular}
}
\end{table}

\noindent\textbf{Stages 2-3: rendering and segmentation (SD~3.5L + SAM3).}
We render the caption with Stable Diffusion 3.5 Large up to three times under a fixed negative prompt covering crops, clutter, duplication, and quality artifacts. Per-class instance masks are extracted by SAM3 with text prompts of the form \texttt{``a \{label\}''}; a small set of overrides resolves ambiguous COCO names (\texttt{bed} $\to$ \texttt{``a picture of a bed''}, \texttt{keyboard} $\to$ \texttt{``a computer keyboard''}, \texttt{skis} $\to$ \texttt{``a pair of skis''}, \texttt{airplane} $\to$ \texttt{``an airplane''}, \texttt{hot\_dog} $\to$ \texttt{``a hot dog in a bun''}). SAM3 finds $89.5\%$ of the intended classes on one render and $95.9\%$ with up to three; we keep the first render in which it finds every intended class.

\noindent\textbf{Stage 4: consistency audit (Grounding DINO).}
Grounding DINO Swin-B runs over the COCO-$80$ vocabulary and applies two checks: (i)~\textit{extras}, any detected class above threshold not in the intended set;
(ii)~\textit{duplicates}, more than one detection of any intended class. Failing scenes are discarded.
The audit keeps $85.0\%$ of the renders at $k = 0$, $75.5$, $70.9$, $59.8$ and $51.0\%$ at $k = 1$ to $4$, and $46.5\%$ at $k = 5$; every rejected render has an extra or a duplicate object.

\noindent\textbf{Stage 5: second audit and balancing.} SAM3 and OWLv2~\cite{minderer2023scaling} run over the COCO-$80$ vocabulary at low thresholds. An unintended object on which both agree (SAM3 score $\geq 0.5$, OWLv2 score $\geq 0.2$, box IoU $\geq 0.3$) becomes an ignore region, on which predictions are not counted as false positives, if the scene is typical or the object's NPMI with every intended class is at most $0$; otherwise the scene is dropped, so that no unlabeled typical partner enters an atypical scene. Scenes that miss an intended class are dropped, and the typical and atypical sub-tiers at each $k$ are cut to the same size. Of the rendered scenes, $44.6\%$ enter the benchmark.

\begin{table}[p]
\centering\footnotesize\hbadness=10000
\caption{\textbf{Per-class visual descriptions for the \texttt{\{class\_visual\_description\}} field.} All $80$ COCO categories, in source order (vehicles $\to$ street furniture $\to$ animals $\to$ accessories $\to$ sports $\to$ kitchenware $\to$ food $\to$ furniture $\to$ electronics $\to$ appliances $\to$ indoor).}
\label{tab:class_descriptions}
\resizebox{\textwidth}{!}{
\begin{tabular}{@{}p{0.10\textwidth}p{0.36\textwidth}p{0.10\textwidth}p{0.36\textwidth}@{}}
\toprule
Class & Description & Class & Description \\
\midrule
person         & a person standing, full body, casual clothing             & wine glass     & a stemmed wine glass with bowl and base \\
bicycle        & a bicycle with two spoked wheels, frame and handlebars    & cup            & a ceramic cup with handle \\
car            & a four-wheeled passenger car, clearly side or front view  & fork           & a metal fork with four tines \\
motorcycle     & a motorcycle with two wheels, fuel tank and handlebars    & knife          & a sharp metal knife with blade and handle \\
airplane       & a fixed-wing airplane with wings and engines              & spoon          & a polished metal spoon \\
bus            & a long passenger bus with windows along its length        & bowl           & a round food bowl, ceramic or wooden \\
train          & a train with multiple connected carriages                 & banana         & a ripe yellow banana with curved shape \\
truck          & a heavy truck with a large cargo area or trailer          & apple          & a round red or green apple with stem \\
boat           & a boat floating on water, hull and deck visible           & sandwich       & a sandwich with bread slices and visible filling \\
traffic light  & a vertical traffic light with red, yellow, and green lamps & orange        & a round orange citrus fruit \\
fire hydrant   & a red fire hydrant on a sidewalk                          & broccoli       & a green broccoli floret with stem \\
stop sign      & a red octagonal stop sign on a metal pole                 & carrot         & an orange carrot with green tops \\
parking meter  & a single parking meter on a metal pole                    & hot dog        & a hot dog in a bun \\
bench          & a wooden or metal park bench with backrest                & pizza          & a round pizza with cheese and toppings \\
bird           & a small bird with feathers and beak                       & donut          & a round glazed or frosted donut with hole in the middle \\
cat            & a domestic cat with fur, four legs and tail               & cake           & a layered round cake with frosting \\
dog            & a dog with fur, four legs, ears and tail                  & chair          & a single chair with seat, backrest and four legs \\
horse          & a horse with mane and tail, four hooved legs              & couch          & a long upholstered couch with cushions \\
sheep          & a sheep with wooly white coat                             & potted plant   & a green leafy plant in a clay or ceramic pot \\
cow            & a cow with horns, large body and udder                    & bed            & a bed with mattress, pillows and bedding \\
elephant       & a gray elephant with long trunk and large ears            & dining table   & a wooden dining table with flat top \\
bear           & a brown or black bear with thick fur                      & toilet         & a white porcelain toilet with bowl and tank \\
zebra          & a zebra with bold black and white stripes                 & tv             & a flat-screen television on a stand or wall mount \\
giraffe        & a giraffe with very long neck and patterned coat          & laptop         & an open laptop computer with keyboard and screen \\
backpack       & a backpack with shoulder straps and zippered compartments & mouse          & a computer mouse peripheral device with scroll wheel and cable \\
umbrella       & an open umbrella with curved canopy and handle            & remote         & a TV remote control with buttons \\
handbag        & a stylish handbag with handle or shoulder strap           & keyboard       & a computer keyboard with QWERTY layout of keys for typing \\
tie            & a long necktie with patterned fabric                      & cell phone     & a smartphone with rectangular screen \\
suitcase       & a hard-sided rolling suitcase with handle                 & microwave      & a black or white microwave oven \\
frisbee        & a colorful round flying disc                              & oven           & a kitchen oven with door and control panel \\
skis           & a pair of long skis with bindings                         & toaster        & a metal toaster with two slots \\
snowboard      & a snowboard with bindings on its top surface              & sink           & a sink with faucet, basin and drain \\
sports ball    & a round sports ball with visible texture or panels        & refrigerator   & a tall refrigerator with door and handle \\
kite           & a colorful kite flying with a long tail                   & book           & a single closed hardcover or paperback book \\
baseball bat   & a baseball bat, smooth wooden cylinder with grip          & clock          & an analog wall clock with hour and minute hands \\
baseball glove & a leather baseball glove                                  & vase           & a ceramic or glass vase \\
skateboard     & a skateboard with four wheels and printed deck            & scissors       & a pair of scissors with two metal blades and grips \\
surfboard      & a long pointed surfboard with fin underneath              & teddy bear     & a soft fluffy teddy bear stuffed toy \\
tennis racket  & a tennis racket with strung head and grip                 & hair drier     & a handheld hair dryer with motor body and nozzle \\
bottle         & a single bottle, narrow neck and wider body               & toothbrush     & a single toothbrush with bristles and handle \\
\bottomrule
\end{tabular}
}
\end{table}

\noindent\textbf{Construction notes.} Three properties of the scenes as built qualify the design above. (i)~The atypical sub-tiers are not limited to low-NPMI pairs: pairs that never co-occur in COCO~\texttt{train2017} score above every co-occurring pair, so $39$ to $64\%$ of the anchor-partner pairs in the atypical sub-tiers have $N_{ij}{=}0$ (Tab.~\ref{tab:strange_tiers}); the NPMI medians cover the other pairs. \texttt{tier\_never} differs in asking for $N_{ij}{=}0$ for every pair, partner pairs included, with one exception: for $76$ of its $466$ scenes ($16.3\%$) the anchor is one of $11$ classes that co-occur with almost every other class (\eg, \emph{person}, \emph{car}, \emph{cup}), and the sampler falls back to the anchor's rarest partners (median $14$ co-occurring images, NPMI $-0.24$); $40$ of these scenes contain no pair with $N_{ij}{=}0$. Conversely, when no candidate meets the typical rule the sampler takes the highest-scoring one, so $3.5$, $10.3$, $16.5$ and $20.6\%$ of the typical anchor-partner pairs at $k = 1$ to $4$ have NPMI below $0.1$, most of them below $0$; the medians of Tab.~\ref{tab:strange_tiers} include them. (ii)~The caption request for the atypical sub-tiers $T_2$ to $T_6$ used the wording of the typical ones (Tab.~\ref{tab:tier_phrases}), so the two arms differ in the class set given to the caption model, not in the context wording; the final captions mention an unusual context in $6$ to $16\%$ of the atypical scenes and in none of the typical ones. (iii)~The benchmark boxes are the bounding boxes of the SAM3 masks; no audit compares them with the planner layout.

\FloatBarrier
\subsection{Implementation Details}
\label{app:strange:impl}

Table~\ref{tab:strange_impl} presents all numeric hyperparameters across the five pipeline stages. Defaults match the final configuration that produced the dataset of Table~\ref{tab:strange_tiers}.

\begin{table}[!t]
\centering\small
\caption{\textbf{Hyperparameters per pipeline stage for creating STRANGE-Bench.}}
\label{tab:strange_impl}
\begin{tabularx}{\linewidth}{@{}ll>{\raggedright\arraybackslash}X@{}}
\toprule
Stage & Knob & Value \\
\midrule
\multirow{4}{*}{1.1 Layout call (Qwen)}
 & canvas size $W\!\times\!H$           & $1024 \times 1024$ \\
 & sampling temperature (3 retries)     & $0.4 \to 0.7 \to 1.0$ \\
 & top-$p$ / max new tokens             & $0.95$ / $512$ \\
 & min box side / fallback          & $64$\,px / layout solver \\
\midrule
\multirow{2}{*}{1.2 Caption call (Qwen)}
 & sampling temperature                 & $0.7$ ($+0.2$ per retry, max $2$ retries) \\
 & top-$p$ / max new tokens             & $0.95$ / $256$ \\
\midrule
\multirow{5}{*}{2 Render (SD~3.5L)}
 & resolution               &  $1024 \times 1024$ \\
 & renders per scene                & up to $3$ \\
 & inference steps                      & $40$ \\
 & classifier-free guidance scale       & $3.5$ \\
 & negative prompt                      & fixed \\
\midrule
\multirow{1}{*}{3 Segment (SAM3)}
 & text prompt template                 & \texttt{``a \{class\_name\}''} \\

\midrule
\multirow{2}{*}{4 Audit (Grounding DINO)}
 & detection score threshold            & $0.50$ \\
 & checks        & extras, duplicates \\
\midrule
\multirow{2}{*}{5 Second audit}
 & SAM3 / OWLv2 score threshold   & $0.5$ / $0.2$ \\
 & agreement IoU                  & $0.3$ \\
\bottomrule
\end{tabularx}
\end{table}

\FloatBarrier
\subsection{User Study}
\label{app:strange:user-study}

\begin{figure}[!htbp]
\centering
\includegraphics[width=\linewidth]{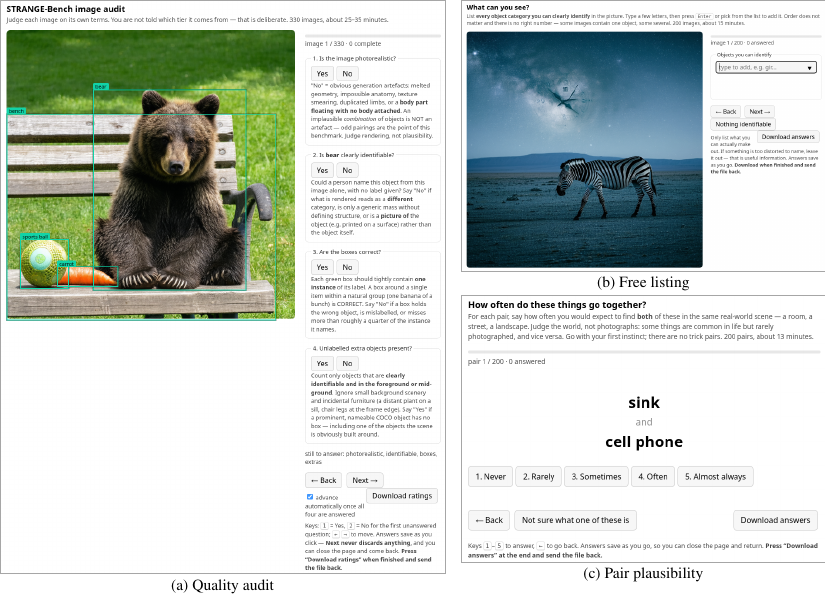}
\caption{\textbf{Rating interfaces of the user study.} (a)~Quality audit: the annotated boxes are drawn so that box correctness can be judged, and the instructions ask raters to judge the rendering, not the plausibility of the object combination. (b)~Free listing: no boxes are drawn and the anchor is never named. (c)~Pair plausibility: two class names, no image.}
\label{fig:user_study_ui}
\end{figure}

We ran three blinded studies on randomly drawn scenes, with the sub-tier hidden and the order shuffled (interfaces in Fig.~\ref{fig:user_study_ui}). The raters were unpaid volunteers, and no personal data was collected. They used an initial version of STRANGE-Bench ($4{,}003$ scenes) that shares the layout, rendering and segmentation stages of App.~\ref{app:strange:pipeline} but was filtered by the Grounding~DINO audit of Stage~4 alone: one SD~3.5L render per scene, no second audit, no missing-class rule, and no balancing of the typical and atypical sub-tiers. That audit kept about $76\%$ of the renders. The released benchmark renders up to three times per scene and adds Stage~5, after which $44.6\%$ of the rendered scenes remain.

\noindent\textbf{Quality audit.} Raters judged $330$ scenes ($30$ per sub-tier) on four binary criteria: the image is photorealistic, the anchor is identifiable, the boxes are correct, and no unlabeled object is present. Of four raters, one passed every scene on three of the four criteria and is excluded; agreement among the other three is high (Gwet's AC1 $0.86$ to $0.99$ per criterion). Typical and atypical scenes at the same $k$ are rendered equally well: their pass rates differ by at most $7.8$ points (unlabeled objects), and anchor identifiability by at most $1.1$. Tab.~\ref{tab:user_study_quality} compares the scenes with $330$ COCO photographs on the same instrument, rated by the human raters and by an LLM judge given their instructions.

\begin{table}[!b]
\centering\small
\caption{\textbf{Annotation quality of the initial STRANGE-Bench scenes and of COCO photographs.} Pass rates (\%) on the four criteria of the quality audit and on all four together, $330$ images per source (COCO images with their own annotations, stratified by object count). LLM judge: Claude Sonnet~5 given the raters' instructions; it is stricter on box tightness than the human raters. Humans: answers pooled over three raters for STRANGE-Bench and two for COCO; ``All four'' counts one rater's answers on one image.}
\label{tab:user_study_quality}
\begin{tabular}{@{}llccccc@{}}
\toprule
Rater & Source & Realism & Anchor & Boxes & No extras & All four \\
\midrule
LLM judge & STRANGE-Bench (initial) & $96.1$ & $99.4$ & $85.8$ & $86.7$ & $70.6$ \\
 & COCO photographs & $99.7$ & $96.1$ & $74.8$ & $93.0$ & $69.7$ \\
\midrule
Humans & STRANGE-Bench (initial) & $93.4$ & $99.5$ & $98.2$ & $94.5$ & $87.1$ \\
 & COCO photographs & $99.7$ & $98.5$ & $98.3$ & $98.0$ & $95.5$ \\
\bottomrule
\end{tabular}

\end{table}

\begin{figure}[!htbp]
\centering
\includegraphics[width=\linewidth]{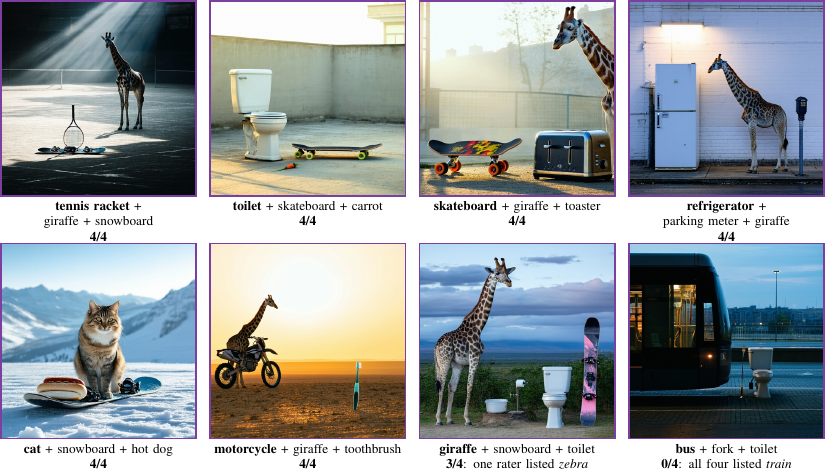}
\caption{\textbf{Free-listing answers on $T_{\text{never}}$ scenes.} Eight of the $100$ $T_{\text{never}}$ scenes of the free-listing task, as the raters saw them. Below each scene: the intended classes (anchor in bold) and how many of the four raters listed the anchor. In the two misses the raters named a similar class.}
\label{fig:user_study_examples}
\end{figure}

\noindent\textbf{Free listing.} Raters saw $200$ scenes ($100$ from $T_1$ with natural background, $100$ from $T_{\text{never}}$) without boxes and listed every category they could identify; we score whether the anchor is among them. Human anchor recall drops by $1.0$ point from $T_1$ to $T_{\text{never}}$, while a fine-tuned DINOv3 classifier loses $31.4$ points over all scenes of the two sub-tiers and $31.0$ on the $200$ rated ones, and the XPaste\,+\,AAS model of Tab.~\ref{tab:dinov3_main} $5.9$ and $3.0$ (Tab.~\ref{tab:user_study_listing}). A rendering defect would lower both; the drop is specific to the model. Fig.~\ref{fig:user_study_examples} shows eight $T_{\text{never}}$ scenes with the raters' answers.

\begin{table}[!htbp]
\centering\small
\caption{\textbf{Free listing: humans and a classifier.} Anchor recall (\%) on $T_1$ scenes (natural background) and $T_{\text{never}}$ scenes of the initial STRANGE-Bench. Rated: the $100 + 100$ scenes shown to the raters. All: every scene of the two sub-tiers ($609$ and $560$). Humans: the anchor is among the listed categories (no boxes drawn, anchor never named), mean over the three raters kept in the quality audit (the fourth rater: $99.0$ and $98.0$). DINOv3 ViT-B/16 fine-tuned: anchor score $\geq 0.5$; XPaste + AAS is the model of Tab.~\ref{tab:dinov3_main}.}
\label{tab:user_study_listing}
\begin{tabular}{@{}llccc@{}}
\toprule
Rater & Scenes & $T_1$ & $T_{\text{never}}$ & $\Delta$ \\
\midrule
Humans & rated & $98.3$ & $97.3$ & $-1.0$ \\
\midrule
DINOv3, Std.\ FT & all & $96.6$ & $65.2$ & $-31.4$ \\
 & rated & $98.0$ & $67.0$ & $-31.0$ \\
\midrule
DINOv3, XPaste + AAS & all & $99.5$ & $93.6$ & $-5.9$ \\
 & rated & $100.0$ & $97.0$ & $-3.0$ \\
\bottomrule
\end{tabular}

\end{table}

\noindent\textbf{Pair plausibility.} Five raters saw $200$ class pairs as two names without an image ($40$ pairs that never co-occur in COCO and $40$ per NPMI quartile) and rated how often both occur in the same real-world scene on a five-point scale. The consensus rating follows NPMI (Spearman $\rho = 0.81$; pairwise agreement between raters $0.65$ to $0.81$), so the NPMI ranking matches human judgments of real-world co-occurrence.

\FloatBarrier

\section{Limitations and Broader Impacts}
\label{app:discussion}

\noindent\textbf{Limitations.}
\emph{(i) Synthetic eval coverage.} STRANGE-Bench reflects SD~3.5L's rendering distribution and is best read as a paired comparison instrument alongside natural-image evaluation, not in place of it.
\emph{(ii) Vocabulary scope.} Both the diagnostic and STRANGE-Bench operate on COCO-$80$ as it is important to study on a manageable scale to exactly quantify typicality bias. Richer vocabularies (long-tail and fine-grained categories) require further regeneration, which our pipeline supports. 
Future work could be devoted to studying the impact on long-tail and finer-grained categories based on our protocol.
\emph{(iii) NPMI as a typicality proxy.} NPMI captures statistical co-occurrence regularities, not semantic or functional plausibility. Refining the typicality axis with semantic priors or human ratings is a natural follow-up. 
\emph{(iv) VLM / LMM mitigation.} CLIP, OpenCLIP, and \llava{}
appear in our diagnostic but not the mitigation study. Studying augmentations that lend themselves to VLM / LMM pre-training or fine-tuning is an obvious next step.

\noindent\textbf{Broader impacts.} Co-occurrence shortcuts inflate in-distribution metrics while silently degrading behavior on under-represented compositions. Our contribution is a model-agnostic diagnostic (NPMI-paired splits, STRANGE-Bench) and a data-side mitigation (AAS) that together let practitioners \emph{measure} this shortcut and, where it matters, push back. The construction recipe is general (\S\ref{app:strange:release}): swapping the vocabulary or co-occurrence source instantiates a fresh probe for any task family. Releasing dataset, code, and pipeline under permissive licenses (CC-BY 4.0 / MIT) is intended to lower the cost of auditing co-occurrence shortcuts in domains beyond COCO to LVIS, Objects365, fine-grained or medical vocabularies. We restrict the dataset's intended use to evaluation to keep the distinction between probe and training distribution intact.
As with any generative-model output, STRANGE-Bench may inherit cultural or demographic biases from SD~3.5L's training distribution, occasionally producing unintended or sensitive imagery. We restrict the vocabulary to common COCO object categories and audit each scene at the consistency-check stage to mitigate this risk.

\end{document}